\documentclass{article}

\makeatletter
\def\input@path{{template/}}
\makeatother

\PassOptionsToPackage{capitalize,noabbrev}{cleveref}   %
\let\OrigAddcontentsline\addcontentsline
\usepackage{labtemplate/aditi}
\let\addcontentsline\OrigAddcontentsline   %
\usepackage{labtemplate/utils}
\usepackage{xurl}
\usepackage{needspace}

\makeatletter\let\c@definition\relax\let\c@assumption\relax\let\c@remark\relax\makeatother
\renewcommand{\topfraction}{0.92}
\renewcommand{\textfraction}{0.06}
\usepackage{microtype}
\usepackage{graphicx}
\newcommand{\figwidth}{0.88\linewidth}   %
\usepackage{float}   %
\usepackage{booktabs}
\usepackage{multirow}
\usepackage{url}
\usepackage{caption}
\usepackage{subcaption}
\usepackage{placeins}
\usepackage{wrapfig}
\usepackage{needspace}
\usepackage{enumitem} %

\usepackage{titletoc}   %
\usepackage{hyperref}
\usepackage{amsmath}
\usepackage{amssymb}
\usepackage{mathtools}
\usepackage{amsthm}
\usepackage{cleveref}   %

\definecolor{darkblue}{rgb}{0, 0, 0.5}
\hypersetup{colorlinks=true, citecolor=darkblue, linkcolor=darkblue, urlcolor=darkblue}

\theoremstyle{definition}
\newtheorem{definition}[theorem]{Definition}
\newtheorem{assumption}[theorem]{Assumption}
\theoremstyle{remark}
\newtheorem{remark}[theorem]{Remark}
\newcommand{\thetapt}{\theta_{\mathrm{PT}}}
\newcommand{\thetaft}{\theta_{\mathrm{FT}}}
\newcommand{\losspt}{\mathcal{L}_{\mathrm{PT}}}
\newcommand{\lossft}{\mathcal{L}_{\mathrm{FT}}}

\definecolor{aditiViolet}{HTML}{7C3AED}  %

\definecolor{jacobOrange}{HTML}{EA580C}  %

\definecolor{jingchuPlum}{HTML}{C71585}   %

\usepackage[most]{tcolorbox}
\usepackage{colortbl}
\definecolor{light-blue}{HTML}{E6F0FF} 
\definecolor{statementBackground}{HTML}{F5F5FA}

\newtcolorbox{cleanbox}{
  colback=statementBackground,   %
  colframe=statementBackground,
  boxrule=0pt,
  boxsep=2pt,      %
  left=3pt,        %
  right=3pt,       %
  top=3pt,         %
  bottom=3pt,      %
  arc=3pt,         %
  enhanced,
  breakable
}
\definecolor{boxOrange}{RGB}{204,122,0}

\newtcolorbox{ttlbox}[2]{
  enhanced, breakable,     %
  colback=statementBackground, frame hidden, boxrule=0pt, arc=1mm,
  colupper=black, coltitle=black, fontupper=\normalfont, fonttitle=\normalfont\bfseries,
  title={#2}, attach title to upper={.\enspace},
  boxsep=0pt, left=3mm, right=3mm, top=2mm, bottom=2mm,
  before skip=8pt, after skip=8pt
}
\newcommand{\gdname}{\textcolor{boxTeal!85!black}{\textsf{GD}}}
\newcommand{\muname}{\textcolor{boxPink!88!black}{\textsf{Muon}}}
\newenvironment{defnbox}[2]
  {\crefalias{theorem}{definition}%
   \refstepcounter{theorem}\label{#2}%
   \ttlbox{boxOrange}{Definition \thetheorem\ (#1)}}
  {\endttlbox}

\definecolor{boxPurple}{RGB}{126,63,171}
\newenvironment{thmbox}[2]
  {\refstepcounter{theorem}\label{#2}%
   \ttlbox{boxPurple}{Theorem \thetheorem\ (#1)}\itshape}
  {\endttlbox}

\definecolor{boxPink}{RGB}{198,64,120}
\newenvironment{lmmbox}[2]
  {\crefalias{theorem}{lemma}%
   \refstepcounter{theorem}\label{#2}%
   \ttlbox{boxPink}{Lemma \thetheorem\ (#1)}\itshape}
  {\endttlbox}

\newenvironment{corbox}[2]
  {\crefalias{theorem}{corollary}%
   \refstepcounter{theorem}\label{#2}%
   \ttlbox{boxPurple}{Corollary \thetheorem\ (#1)}\itshape}
  {\endttlbox}

\definecolor{boxTeal}{RGB}{0,114,178}
\newtcolorbox{tblcard}{
  enhanced,   %
  colback=statementBackground, frame hidden, boxrule=0pt, arc=1mm,   %
  boxsep=1.2mm, top=1.4mm, bottom=1.8mm, left=2.5mm, right=3mm,
  before skip=8pt, after skip=8pt
}
\newenvironment{tblbox}[1]
  {\refstepcounter{table}\label{#1}%
   \tblcard\centering\small\setlength{\tabcolsep}{2pt}%
   \arrayrulecolor{black}\aboverulesep=0.45ex \belowrulesep=0.45ex}
  {\endtblcard}

\definecolor{boxQuestion}{RGB}{0,137,123}
\newtcolorbox{questionbox}{
  enhanced, width=\linewidth,
  colback=statementBackground, frame hidden, boxrule=0pt, arc=1.2mm,   %
  halign=center, fontupper=\small\itshape,
  left=4mm, right=4mm, top=1.8mm, bottom=1.8mm,
  before skip=7pt, after skip=9pt
}

\title{Pretraining Data Scale Reverses the Muon--AdamW Fine-Tuning Gap}

\setTitleruleGap{0.75pt}

\setauthors{Catherine Li$^{*}$ \authorsep Jingchu Gai$^{*}$ \authorsep Andrej Risteski \authorsep Jacob Mitchell Springer$^{\dagger}$ \authorsep Aditi Raghunathan$^{\dagger}$}
\setaffils{Carnegie Mellon University}
\setFirstPageFooter{* Equal contribution.\hspace{1em}$\dagger$ Equal advising.}

\setemail{\{catheri4, jgai\}@andrew.cmu.edu}

\begin{document}

\maketitle

\begin{abstract}
Muon has recently emerged as a promising optimizer for pretraining large language models, offering a speedup over AdamW in minimizing pretraining loss. However, it remains unclear whether Muon's advantage persists after fine-tuning. In this paper, we study the downstream behavior of Muon and AdamW and uncover a surprising trend: beyond a sufficiently large pretraining data budget, AdamW outperforms Muon after fine-tuning. Moreover, as model size increases, AdamW overtakes Muon at progressively smaller pretraining token--parameter ratios. To explain this reversal, we view fine-tuning as a perturbation of the outputs and introduce the margin-to-spread ratio, which measures how far the correct logit lies above the wrong ones relative to how widely the wrong logits are spread. In language models, Muon has the larger margin-to-spread ratio at small data budgets and AdamW at large ones, mirroring the fine-tuning trend. In a linear associative memory, we prove that after one update the margin-to-spread ratio of gradient descent gains on that of Muon as the data budget grows, and simulations show that this trend persists after extended training.
\end{abstract}

\section{Introduction}
\label{sec:intro}

The Muon optimizer has recently shown strong empirical performance for pretraining LLMs, challenging AdamW's status as the standard optimizer for minimizing pretraining loss \citep{jordan2024muon}. Prior work has shown that the pretraining optimization choices that minimize pretraining loss do not necessarily yield the best model after fine-tuning \citep{sam_forgetting}. What ultimately matters downstream is the \emph{learning--forgetting tradeoff}: how well a fine-tuned model learns the target task while retaining its pretrained knowledge. Motivated by this, we ask:

\begin{center}
\textit{Does Muon's pretraining advantage over AdamW extend to the learning--forgetting tradeoff?}
\end{center}

\noindent To answer this question, we measure the Muon--AdamW \emph{fine-tuning gap}, which compares the two optimizers on this tradeoff; a negative gap indicates that the Muon pretrained model dominates the AdamW pretrained model after fine-tuning, and a positive gap indicates the reverse. We show that the answer depends on the pretraining data budget, and find that \textbf{pretraining data scale reverses the Muon--AdamW fine-tuning gap}. At small budgets, Muon achieves lower fine-tuning loss with less forgetting than AdamW and the fine-tuning gap is negative. At larger budgets, this trend reverses: AdamW achieves a better learning--forgetting tradeoff than Muon, and the fine-tuning gap becomes positive.

In \cref{sec:ft}, we pretrain OLMo-style 30M--600M models \citep{groeneveld-etal-2024-olmo, olmo20242olmo2furious} on $150$M to $154$B DCLM tokens \citep{li2024datacomplm}, comparing AdamW and Muon, and fine-tune each base model on four datasets while sweeping the fine-tuning optimizer and learning rate, for nearly 600 pretraining runs and 29{,}000 fine-tuning runs in total. We show that when tuning pretraining hyperparameters for the lowest pretraining loss, Muon consistently achieves lower loss than AdamW. However, with sufficient pretraining data, AdamW outperforms Muon after fine-tuning, reversing the Muon–AdamW fine-tuning gap. This reversal also holds when we disregard pretraining loss, instead sweeping pretraining hyperparameters to obtain the best learning–forgetting tradeoff after fine-tuning. We additionally show that \textbf{as model size scales, AdamW surpasses Muon at progressively smaller data budgets}.

In \cref{sec:analysis}, we investigate why the Muon--AdamW fine-tuning gap reverses as the pretraining data budget grows. Fine-tuning changes the output distribution, so we generalize fine-tuning updates with perturbations to the output distribution and study which optimizer produces base models that are more robust to such perturbations. To quantify sensitivity to output perturbations, we introduce the \textbf{margin-to-spread ratio (MSR)}, a simple metric derived from the model's output logits that characterizes robustness. For each prediction, the MSR is the difference between the correct logit and the average of the incorrect logits, normalized by the spread of the incorrect logits (\cref{def:rho}): a prediction is more robust when the MSR is larger.

\begin{figure}[t]
\centering
\includegraphics[width=\figwidth]{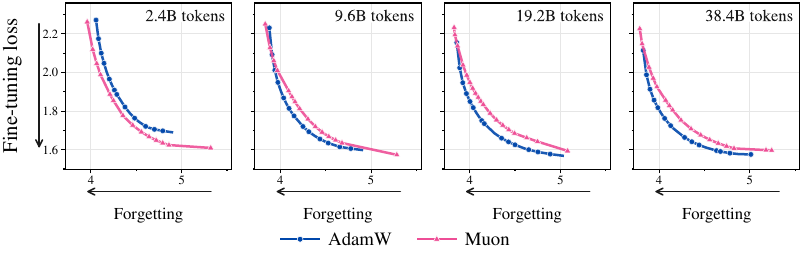}
\caption{\textbf{Beyond a sufficiently large data budget, AdamW outperforms Muon.} We pretrain OLMo-60M models with Muon and AdamW and fine-tune on StackMathQA. For each pretraining optimizer, we take the pretrained model with the lowest pretraining loss, sweep across all fine-tuning configurations, and plot the resulting learning--forgetting frontier, from $2.4$B to $38.4$B pretraining tokens. At $2.4$B tokens, Muon's frontier dominates AdamW's; at larger budgets the trend reverses and AdamW surpasses Muon. This reversal happens despite Muon achieving lower pretraining loss across all data budgets.}
\label{fig:combined}
\end{figure}

To understand where this trend comes from, we study a linear associative memory trained with gradient descent (GD) and Muon (\cref{sec:analysis-budget}). Analyzing a single update from zero initialization, we prove that \textbf{the ratio of GD's MSR to Muon's increases with the data budget} (\cref{thm:ratio}). Simulations of many training steps in the linear associative memory setting confirm the same trend: Muon has the larger MSR at small data budgets, while GD has the larger MSR at large data budgets (\cref{sec:budget-experiments}). Measuring the MSR of our pretrained language models, we find that Muon has the larger MSR at small data budgets, while AdamW has the larger MSR at large data budgets, mirroring the reversal of the fine-tuning gap (\cref{sec:rho-experiments}).

\section{Preliminaries}
\label{sec:prelim}

Our goal is to compare AdamW- and Muon-pretrained models \emph{after fine-tuning}. Since fine-tuning can induce forgetting of the pretraining distribution, performance on the fine-tuning task alone is not sufficient: one base model may learn more but also forget more. We therefore compare base models by the tradeoff between the two, and define metrics to track how this comparison changes with model size and pretraining data budget. Let $\thetapt$ denote a pretrained model.

\noindent\textbf{Learning--forgetting tradeoff.} Fine-tuning improves downstream performance but may degrade pretrained knowledge \citep{goodfellow2013empirical,kirkpatrick2017overcoming,overtrained}. When $\thetapt$ is fine-tuned to obtain $\thetaft$, we measure learning by the fine-tuning loss $\lossft(\thetaft)$ and forgetting by the pretraining loss $\losspt(\thetaft)$. Since both depend on the fine-tuning hyperparameters, a single fine-tuned checkpoint does not capture how well $\thetapt$ can be adapted. We therefore summarize each base model by the outcomes reachable by fine-tuning it,
\[
\mathcal{T}(\thetapt)
\coloneqq
\left\{
\bigl(\lossft(\thetaft),\losspt(\thetaft)\bigr)
\,\middle|\,
\thetaft\in\Theta_{\mathrm{FT}}(\thetapt)
\right\},
\]
where $\Theta_{\mathrm{FT}}(\thetapt)$ is the set of models obtained by fine-tuning $\thetapt$, and compare base models through the lower-left Pareto frontier of this set, which we call the \emph{learning--forgetting frontier}. For a collection of base models $\mathcal{B}$, we compare the corresponding sets
\[
\mathcal{T}(\mathcal{B})
\coloneqq
\{\mathcal{T}(\thetapt)\mid \thetapt\in\mathcal{B}\}.
\]

\noindent\textbf{Fine-tuning gap.} We compare the two optimizers on this tradeoff with a single metric, \emph{fine-tuning gap} $G_{\mathrm{FT}}$, which we define as the signed mean minimum Euclidean distance between the AdamW and Muon frontier. The fine-tuning gap is positive when the AdamW frontier dominates the Muon frontier, and the reverse else wise. Thus $G_{\mathrm{FT}}>0$ when AdamW dominates and $G_{\mathrm{FT}}<0$ when Muon dominates (\cref{app:evaluation} gives the details).

\noindent\textbf{Crossover token--parameter ratio (TPR).} We quantify pretraining data budget using \emph{token--parameter ratio} (TPR), the number of pretraining tokens per model parameter. For a fixed model size and fine-tuning dataset, the \emph{crossover TPR} is the minimum TPR required for the AdamW frontier to overtake the Muon frontier.

\section{Experiments}
\label{sec:ft}
\begin{figure}[t]
\centering
\includegraphics[width=\figwidth]{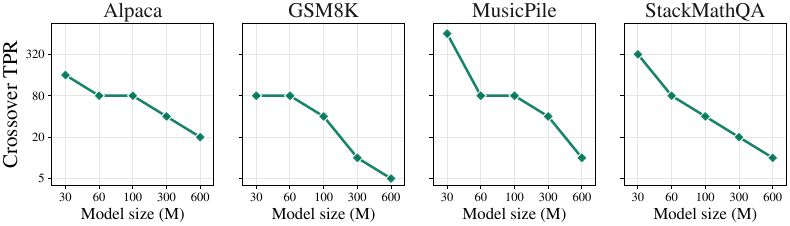}
\caption{\textbf{Crossover TPR decreases with model size.} In the lowest-pretraining-loss setting, where each pretraining optimizer's lowest-loss pretrained model is swept across all fine-tuning hyperparameters, Muon's advantage reverses at substantially smaller pretraining data budgets as model size increases. The exact crossover TPR varies depending on the fine-tuning dataset, but the high-level trend is consistent.}
\label{fig:crossover}
\end{figure}

We now compare how the learning--forgetting tradeoffs of Muon and AdamW scale with pretraining data budget and model size. In \cref{sec:main-result}, we show that as the pretraining data budget increases, Muon's advantage over AdamW shrinks and eventually reverses in favor of AdamW. This reversal happens at progressively smaller data budgets as model size grows.

\subsection{Experimental setup}
\label{sec:setup}
\noindent\textbf{Pretraining.} We pretrain OLMo-style 30M--600M models on $150$M to $154$B DCLM tokens, comparing AdamW and Muon under a warmup--stable--decay (WSD) schedule. We sweep hyperparameters broadly and report the complete pretraining methodology in \cref{app:experimental-details}.

\noindent\textbf{Fine-tuning.} For each base model, we evaluate the learning--forgetting tradeoff on four standard datasets: GSM8K~\citep{cobbe2021gsm8k}, StackMathQA~\citep{zhang2024stackmathqa}, Alpaca~\citep{taori2023alpaca}, and MusicPile~\citep{yuan2024chatmusician}, sweeping AdamW learning rates from $10^{-4}$ to $10^{-2}$ and Muon learning rates from $6\times10^{-4}$ to $2\times10^{-2}$ (\cref{app:ft-setup}). We fine-tune AdamW-pretrained models with AdamW and Muon-pretrained models with both AdamW and Muon: AdamW fine-tuning consistently outperforms Muon fine-tuning on AdamW-pretrained models, whereas neither optimizer is consistently better on Muon-pretrained models (\cref{app:ft-optimizer}). We then plot the learning--forgetting frontier over the resulting runs. We choose a broad learning-rate range to ensure we capture the complete learning--forgetting tradeoff spectrum: higher learning rates favor learning at the expense of forgetting, while lower rates preserve pretraining knowledge but learn less.

\noindent\textbf{Evaluating the learning--forgetting tradeoff.} For each fine-tuned checkpoint, we compute the fine-tuning loss and pretraining loss on held-out fine-tuning and pretraining (DCLM) data, and use these checkpoints to approximate the learning--forgetting tradeoff. Although this approximation is constructed from a finite set of fine-tuning runs, we conduct a comprehensive sweep over four datasets and across a wide range of fine-tuning learning rates to obtain a close estimate. We evaluate the learning--forgetting tradeoff in two settings: \emph{lowest-pretraining-loss setting}, and the \emph{best-frontier setting}.

\begin{enumerate}
    \item \textbf{Lowest-pretraining-loss setting}: Pretraining hyperparameters are typically selected to minimize pretraining loss. Therefore, for each optimizer, we tune the pretraining hyperparameters to minimize pretraining loss for a given pretraining data budget and model size. We then sweep across all fine-tuning configurations for each of the lowest-loss pretrained models, and evaluate the resulting frontier. 
    \item \textbf{Best-frontier setting}: However, prior studies find that the lowest-loss pretrained model is not always the best model after fine-tuning: models pretrained with other hyperparameters can offer a better learning--forgetting tradeoff~\citep{sam_forgetting}.  Therefore, for each pretraining optimizer, we additionally sweep across all pretraining and fine-tuning configurations to achieve the best possible learning--forgetting frontier for a given pretraining data budget and model size (\cref{app:best-frontier}).
    
\end{enumerate}

\subsection{Main result}
\label{sec:main-result}

We now investigate how the Muon--AdamW fine-tuning gap scales with pretraining data budget and model size.

\begin{figure}[t]
  \centering
  \includegraphics[width=\figwidth]{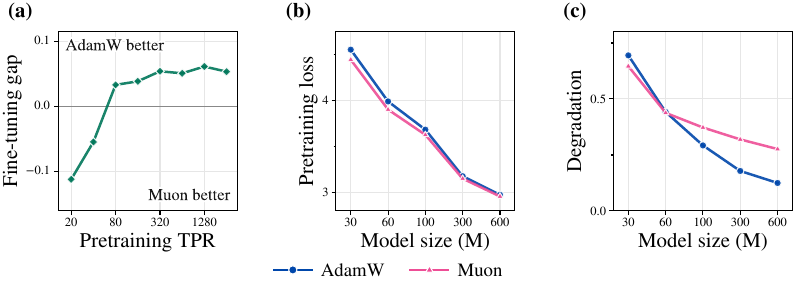}
  \caption{\textbf{The Muon--AdamW fine-tuning gap widens with increasing pretraining data budget, and Muon's advantage shrinks with model size.} \textbf{(a)} The fine-tuning gap after fine-tuning OLMo-60M models on StackMathQA, in the lowest-pretraining-loss setting (each pretraining optimizer's lowest-loss pretrained model swept across all fine-tuning configurations). As the pretraining data budget increases, the fine-tuning gap transitions from negative to positive, with AdamW outperforming Muon beyond the crossover TPR. \textbf{(b)} Pretraining loss at 40 TPR. Muon's advantage over AdamW shrinks as model size grows. \textbf{(c)} Degradation at 40 TPR, when we accept a StackMathQA loss $10\%$ above the best in exchange for less forgetting. Muon degrades increasingly more than AdamW as model size grows.}
  \label{fig:gap-row}
  \label{fig:size-forgetting}
\end{figure}

\textbf{Main result: Pretraining data scale reverses the Muon--AdamW fine-tuning gap.} We compare the fine-tuning gap for AdamW and Muon at different pretraining data budgets in the lowest-pretraining-loss setting, using StackMathQA as the fine-tuning dataset (\cref{fig:gap-row}a). At small pretraining data budgets ($2.4$B tokens), Muon Pareto-dominates AdamW (\cref{fig:combined}) and the fine-tuning gap is negative. At larger pretraining data budgets, this trend reverses: beyond $2.4$B tokens, AdamW achieves lower fine-tuning loss with less forgetting than Muon. The trend also holds in the best-frontier setting (\cref{fig:app-gap-all-best}), and for additional fine-tuning datasets in \cref{app:additional-results}.

We next investigate how the reversal of the Muon--AdamW fine-tuning gap scales with model size. Specifically, we ask how the amount of pretraining data needed for AdamW to overtake Muon changes as model size grows. 

\textbf{As model size scales, AdamW surpasses Muon at progressively smaller data budgets.} We find that the crossover TPR decreases with model size. On StackMathQA, the crossover is $320$ TPR for 30M models but only $10$ TPR for 600M models (\cref{fig:crossover}). The specific crossover TPR depends on the fine-tuning dataset, but the broader trend persists; as model size grows, the crossover TPR decreases, and Muon's advantage reverses with substantially smaller data budgets.

We hypothesize that in the lowest-pretraining-loss setting, this scaling trend is driven by two factors: Muon's shrinking pretraining loss advantage and increased degradation relative to AdamW (\cref{fig:gap-row}b,c). For each optimizer and model size, we compare its base pretraining loss and its degradation $\losspt(\thetaft) - \losspt(\thetapt)$, after fine-tuning on StackMathQA. First, for a fixed TPR, the gap in pretraining loss between Muon and AdamW shrinks as model size increases, reducing Muon's advantage as a lower-loss base model. Additionally, for a fixed TPR and fine-tuning loss, Muon forgets increasingly more than AdamW as model size grows. Consequently, as model size scales, Muon's pretraining advantage over AdamW narrows, while its forgetting relative to AdamW increases, enabling AdamW to overtake Muon at progressively smaller data budgets.

\WFclear   %
\section{Analyzing the Muon--AdamW Forgetting Gap}
\label{sec:analysis}
In \Cref{sec:ft}, we show that Muon offers a better learning--forgetting tradeoff at small token budgets, but AdamW overtakes it beyond a crossover budget. This reversal occurs even though Muon achieves lower pretraining loss at every budget we consider. In this section, we develop a theoretical explanation for this reversal. We compare Muon with exact orthogonalization to full-batch gradient descent (GD), a tractable comparison that captures the qualitative budget trend of Muon and AdamW in language models (\cref{sec:rho-experiments}).

To formalize the relationship between GD and Muon, we study a linear associative memory model of fact learning. In this setting, GD and Muon learn the facts to a similar degree, but their checkpoints can differ in their sensitivity to perturbations of the output distribution. Since fine-tuning changes the output distribution, we expect checkpoints that preserve factual accuracy under a wider range of perturbations to be more resistant to forgetting during fine-tuning. Our central goal is therefore to characterize this sensitivity across pretraining budgets and use it to explain why the forgetting gap reverses.

To quantify sensitivity to output perturbations, we introduce the \textbf{margin-to-spread ratio (MSR)}: the difference between the correct answer's logit and the average incorrect logit, normalized by the spread of the incorrect logits. The MSR captures a checkpoint's robustness to output perturbations and provides an analytically tractable basis for comparing GD and Muon.

\noindent\textbf{Roadmap.} We introduce the linear associative memory setting, define the MSR, and establish its connection to sensitivity (\cref{sec:analysis-rho}). We then prove that, after one update step, the ratio of Muon's MSR to GD's decreases as the data budget grows, indicating that Muon becomes more sensitive to output perturbations relative to GD (\cref{sec:budget-theory}). Finally, we show that the MSR tracks the robustness of AdamW and Muon in real language models (\cref{sec:rho-experiments}).

\subsection{Setup and the margin-to-spread ratio}
\label{sec:analysis-rho}
We begin by introducing our linear associative memory setup and defining the margin-to-spread ratio.

\noindent\textbf{Linear associative memory.}\label{sec:rho-setup} To capture the behavior of Muon and GD in a tractable setting, we consider a linear associative memory, in which a single weight matrix is trained to associate each of $N$ keys, represented by a vector, with its corresponding value, also represented by a vector. Formally, we represent each fact by a key--value pair $(u_i,v_i)\in\mathbb{R}^d\times\mathbb{R}^d$, with all keys and values drawn independently from $\mathcal{N}(0,I_d)$. The memory is parameterized by a matrix $W\in\mathbb{R}^{d\times d}$. Given a key $u_i$, it scores each candidate value $v_j$ as $s_{ij}(W):=v_j^\top Wu_i$, which we treat as logits, and retrieves the highest-scoring value. Fact $i$ is recovered when $s_{ii}(W)>s_{ij}(W)$ for all $j\neq i$.

We train with the softmax cross-entropy loss. Given a query distribution $q=(q_1,\ldots,q_N)$ over the facts, the predicted probabilities and cross-entropy loss are
\begin{equation}
  \begin{aligned}
    \hat{p}_W(\cdot \mid i)
      &:= \operatorname{softmax}\big(s_{i1}(W),\ldots,s_{iN}(W)\big), ~~~\mathcal{L}(W)
      := \textstyle -\sum_{i=1}^{N} q_i \log \hat{p}_W(i \mid i).
  \end{aligned}
\end{equation}

\noindent\textbf{Data budget.} We denote the memory dimension by $d$ and the number of facts by $N$, and grow the two together, keeping their ratio $\gamma=N/d>1$ fixed, so that $N=\gamma d$ as $d\to\infty$. This mirrors the typical scaling behavior of language models, which store many more facts than they have dimensions. We call $\gamma$ the \textbf{data budget}: it measures facts per dimension and plays the role of token-parameter ratio in \cref{sec:prelim}. We compare data budgets by taking this limit at different fixed values of $\gamma$. Since $\gamma>1$, we have $N-1\geq d$, and $\log N/d\to0$, written $\log N=o(d)$. For random variables $X$ and a deterministic $a>0$, $X=O_p(a)$ means that for every $\varepsilon>0$ there is a $C$ with $\Pr[|X|>Ca]\leq\varepsilon$ eventually, and $X=o_p(a)$ means that $\Pr[|X|>\varepsilon a]\to0$ for every $\varepsilon>0$.

\noindent\textbf{Optimization.} For simplicity, we consider the exact orthogonalization variant of Muon. For a matrix $A$ with singular value decomposition $A=P\Sigma Q^\top$, let $\operatorname{polar}(A):=PQ^\top$. Muon updates $W_{t+1}=W_t-\eta\operatorname{polar}(\nabla\mathcal{L}(W_t))$. We compare Muon to full-batch gradient descent (GD), whose updates descend along the gradient itself, $W_{t+1}=W_t-\eta\nabla\mathcal{L}(W_t)$.

\noindent\textbf{Margin-to-spread ratio.} In a binary classification problem, a prediction is robust if the correct logit is far from the incorrect logit: the larger the margin, the more robust the prediction. In a multiclass classification problem, such as our linear associative memory setting, there are many incorrect logits. The robustness of a correct output to perturbations of the logits thus depends on its separation from all of the incorrect logits. As it turns out, the degree of separation can be usefully summarized by a single statistic, which we call the \emph{margin-to-spread ratio (MSR)}. MSR captures the average separation between the correct logit and all other logits, normalized by the spread of the incorrect logits. A smaller spread, or a larger average separation, would both place the incorrect logits further below the correct logit relative to their spread, which can increase the robustness of the prediction to perturbations. To capture this intuition, we formally define the margin-to-spread ratio as follows:

\begin{defnbox}{Margin-to-spread ratio}{def:rho}
For fact $i$, define the pairwise margins $m_{ij}:=s_{ii}(W)-s_{ij}(W)$ for $j\neq i$, with mean and variance
\begin{equation}
  h_i := \operatorname*{mean}_{j\neq i} m_{ij},
  \qquad
  \kappa_i^2 :=
    \operatorname*{mean}_{j\neq i}(m_{ij}-h_i)^2.
\end{equation}
The \emph{margin-to-spread ratio} (MSR) for fact $i$ is $\rho_i:=h_i/\kappa_i$, for $\kappa_i>0$. At the model level, we average the mean margins and variances uniformly over facts before taking the ratio:
\begin{equation}
  \bar h := \operatorname*{mean}_i h_i,
  \qquad
  \bar\kappa^2 := \operatorname*{mean}_i \kappa_i^2,
  \qquad
  \rho := \frac{\bar h}{\bar\kappa},
\end{equation}
where $\rho$ is defined for $\bar\kappa>0$.
\end{defnbox}

Since $s_{ii}(W)$ is the same for every competitor $j$, $\kappa_i$ is also the standard deviation of the incorrect logits $s_{ij}(W)$, so the MSR measures the average margin in units of their spread. The definition applies to any model that assigns a logit to each candidate answer; for a language model, $s_{ij}(W)$ are the next-token logits (\cref{sec:rho-experiments}).

Under isotropic Gaussian noise added to the logits, the MSR gives a lower bound on the fraction of incorrect answers that stay below the correct one: with noise of standard deviation $\sigma$ on each logit, this fraction is at least $\rho_i^2/(\rho_i^2+1+2\sigma^2/\kappa_i^2)$ in expectation. For a fixed noise strength relative to the incorrect-logit spread, this bound improves as the MSR increases (\cref{cor:auc-logit}).

\subsection{One-step analysis of the margin-to-spread ratio}
\label{sec:analysis-budget}
\label{sec:budget-theory}
We have so far introduced our linear associative memory setup and built intuition for how the MSR captures the robustness of a model to perturbations of its logits. Here, we argue that the MSR of GD is greater than that of Muon for sufficiently large data budgets. We precisely track the one-step dynamics of the MSR ratio $\mathcal{R}:=\rho^{\mathrm{GD}}/\rho^{\mathrm{Muon}}$ between GD and Muon in the linear associative memory.

We analyze a single update from zero initialization, using uniform query weights $q_i=1/N$. The weights after many optimization steps depend on the whole training trajectory and are hard to characterize precisely, so theoretical studies of how an optimizer shapes the learned weights often analyze its first step~\citep{ba2022onestep,dandi2024giantstep,moniri2024onestep}. \citet{kim2026sharpcapacityscalingspectral} take the same approach to compare a single step of Muon, SGD, and Newton's method in a linear associative memory.

\begin{thmbox}{Informal}{thm:ratio}
Consider one GD step and one Muon step from zero in the Gaussian associative memory model, with data budget $\gamma=N/d$. For sufficiently large $\gamma$:
\begin{itemize}[leftmargin=*, itemsep=3pt, topsep=3pt]
  \item GD has a larger MSR than Muon.
  \item Increasing the data budget $\gamma$ by any fixed percentage increases the ratio $\mathcal{R}=\rho^{\mathrm{GD}}/\rho^{\mathrm{Muon}}$, with high probability as the dimension grows.
  \item The ratio approaches $3\pi/8\approx1.18$ from below at a rate of $1/\gamma$:
  \[
    \textstyle
    -\frac{C}{\gamma}\leq\mathcal{R}-\frac{3\pi}{8}\leq-\frac{c}{\gamma},
  \]
  where $0<c<C$ are absolute constants.
\end{itemize}
All claims hold with probability tending to one as $d\to\infty$, for sufficiently large fixed $\gamma$.
\end{thmbox}

\begin{figure}[H]
\centering
\vspace{-6pt}
\includegraphics[width=\figwidth]{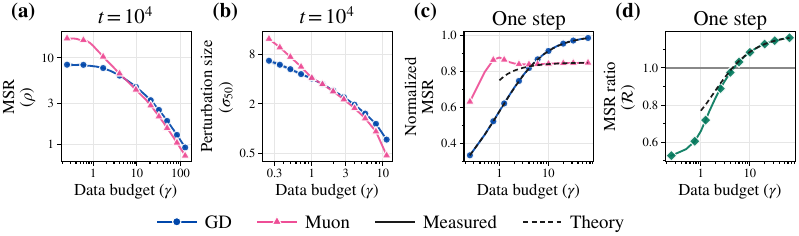}
\vspace{-2pt}
\caption{\textbf{The margin-to-spread ratio of GD gains on Muon's as the data budget grows.} \textbf{(a, b)} After $t = 10^4$ steps with model dimension $d = 256$, GD's MSR $\rho$ overtakes Muon's as $\gamma$ grows, and the ordering of the perturbed robustness $\sigma_{50}$ turns over with it. \textbf{(c, d)} After one step with $d = 512$, GD's normalized MSR $\rho/\sqrt{d/\gamma}$ overtakes Muon's (c), so the MSR ratio $\mathcal{R} = \rho^{\mathrm{GD}}/\rho^{\mathrm{Muon}}$ rises with $\gamma$ (d), following the theory of \cref{thm:ratio}.}
\label{fig:budget-main}
\vspace{-16pt}
\end{figure}

\noindent\textbf{Proof sketch.} Our goal is to compare one update step for Muon and for GD.
\begin{enumerate}[leftmargin=*, itemsep=5pt, topsep=4pt]
  \item Since the MSR does not change when $W$ is multiplied by a positive constant, the one-step weights can be written as
  \[
    \textstyle
    W^{\mathrm{GD}}=G,\qquad W^{\mathrm{Muon}}=O:=\operatorname{polar}(G)
  \]
  (Appendix, Lemma~\ref{lem:one-step}). We now want to compute $\rho=\bar h/\bar\kappa$ for these matrices. This requires the mean margins and their standard deviations; we begin with the means.

  \item \textit{Compute the mean margins.} Recall that $\bar h$ averages the difference between the correct and incorrect logits over facts and incorrect answers. For uniformly weighted facts,
  \[
    \textstyle
    G=\sum_{k=1}^{N}(v_k-\bar v)u_k^\top,\qquad \bar v=\frac1N\sum_{k=1}^{N}v_k.
  \]
  Substituting into this average gives $\bar h(W)=\langle W,G\rangle/(N-1)$, and therefore
  \[
    \textstyle
    \bar h^{\mathrm{GD}}=\frac{\lVert G\rVert_F^2}{N-1},\qquad \bar h^{\mathrm{Muon}}=\frac{\lVert G\rVert_*}{N-1},
  \]
  where $\lVert\cdot\rVert_F$ is the Frobenius norm and $\lVert\cdot\rVert_*$ is the sum of the singular values.

  \item \textit{Compute GD's variance.} For each query $i$, we take the variance across incorrect candidates $j$, holding the dataset and weights fixed:
  \[
    \textstyle
    \kappa_i^2=\operatorname{Var}_{j\neq i}(v_j^\top Gu_i).
  \]
  To calculate its typical size, expand the incorrect logit:
  \[
    \textstyle
    v_j^\top Gu_i\approx\underbrace{\textstyle\sum_{k\notin\{i,j\}}(v_j^\top v_k)(u_k^\top u_i)}_{\text{other facts}}+\underbrace{\lVert u_i\rVert^2(v_j^\top v_i)+\lVert v_j\rVert^2(u_j^\top u_i)}_{\text{query and candidate facts}}.
  \]
  Here $\approx$ omits the mean-subtraction term in $G$, which only affects lower-order terms in the variance. Each other fact contributes a product of two independent, mean-zero inner products, each with variance $d$, giving variance $d^2$. In the query and candidate terms, we approximate the squared norms by $d$, dropping their fluctuations. This gives variance $d^3$ each, with $O(d^2)$ corrections. Replacing $N-2$ by $N$ also discards only $O(d^2)$ terms.

  These contributions are uncorrelated over the Gaussian training data. Including the omitted corrections and concentration of the empirical average, we calculate GD's variance (Appendix, Lemma~\ref{lem:kbar-gd}) to be
  \[
    \textstyle
    (\bar\kappa^{\mathrm{GD}})^2=\big(\underbrace{Nd^2}_{\text{other facts}}+\underbrace{2d^3}_{\text{query and candidate}}\big)(1+o_p(1))=d^3(\gamma+2)(1+o_p(1)).
  \]

  \item \textit{Compute Muon's variance.} Again, the variance is over incorrect candidates for each fixed query. Gaussian isotropy and orthogonality give the leading calculation
  \[
    \textstyle
    (\bar\kappa^{\mathrm{Muon}})^2\approx\frac1{N(N-1)}\sum_i\sum_{j\neq i}(v_j^\top Ou_i)^2\approx\frac1N\sum_i\lVert Ou_i\rVert^2=\frac1N\sum_i\lVert u_i\rVert^2\approx d.
  \]
  The first approximation drops the squared mean of the incorrect logits for each query. The second neglects departures of the sampled incorrect values from perfect isotropy, and the last drops fluctuations in the average squared key length. We calculate the variance including the omitted corrections and the dependence of $O$ on the training data (Appendix, Lemma~\ref{lem:kbar-muon}) to be
  \[
    \textstyle
    (\bar\kappa^{\mathrm{Muon}})^2=d\left[1+O(\gamma^{-3/2})+o_p(1)\right].
  \]

  \item \textit{Put the means and standard deviations together.} We calculate the norms in the mean margins using the Gaussian singular-value distribution (Appendix, Lemma~\ref{lem:norms-wgd}). Dividing by the standard deviations gives, to first order in $1/\gamma$,
  \[
    \textstyle
    \rho^{\mathrm{GD}}\approx\sqrt{\frac d\gamma}\left(1-\frac1\gamma\right),\qquad \rho^{\mathrm{Muon}}\approx\frac{8}{3\pi}\sqrt{\frac d\gamma}\left(1-\frac1{10\gamma}\right).
  \]
  These approximations omit higher-order budget corrections and relative fluctuations that vanish as $d\to\infty$. Taking their ratio cancels the common factor $\sqrt{d/\gamma}$. Keeping track of the errors gives
  \[
    \textstyle
    \mathcal{R}=\frac{3\pi}{8}\left(1-\frac9{10\gamma}\right)+O(\gamma^{-3/2})+o_p(1).
  \]
\end{enumerate}
Since $3\pi/8>1$, GD has the larger MSR at sufficiently large budgets. The negative $1/\gamma$ term in $\mathcal{R}-3\pi/8$ dominates the smaller remainder, giving the claimed rate. A fixed percentage increase in $\gamma$ raises the leading expression by order $1/\gamma$, which also dominates the remainder, establishing the increase with high probability. See \cref{app:ratio-proof} for the quantitative bound and full proof, and \cref{fig:budget-main,app:budget-extra} for numerical checks.

This theorem demonstrates that in the linear associative memory, when the data budget is sufficiently large, checkpoints learned with GD have a larger MSR than checkpoints learned with Muon after a single step of training. Combined with the robustness bound of \cref{sec:analysis-rho}, GD's checkpoints therefore carry the stronger guarantee of robustness to perturbations of the output distribution in our theoretical setting.

\begin{cleanbox}
\noindent\textbf{Remark (Choice of data distribution).} Independent standard Gaussian keys and values provide a canonical isotropic model with no preferred direction. Other data distributions can change both the gradient spectrum and the competing-score variances, so the optimizer ordering may differ. The analogous budget trend in the language-model experiments (\cref{sec:rho-experiments}) suggests that the behavior captured by this simple model is relevant beyond the Gaussian setting.
\end{cleanbox}

\noindent\textbf{Beyond one step.}\label{sec:budget-experiments} We next check whether the trend persists after extended training. We train the linear memory for $T=10^4$ full-batch steps from $W_0=0$ at constant learning rates across three seeds. We summarize the saturated MSR by the per-seed median over $t\geq10^3$ and measure robustness by $\sigma_{50}$, the relative perturbation size at which half of the initially recovered facts are lost. We exclude data budgets past the memory's capacity where the two optimizers recover different item sets. The full setup and results for every dimension $d$ appear in \cref{app:budget-extra}.

The results in \cref{fig:budget-main} show that $\mathcal{R}$ increases with the data budget. Muon has the larger MSR at small $\gamma$, while GD has the larger MSR at large $\gamma$. The robustness comparison shifts in the same direction: Muon tolerates stronger perturbations at small $\gamma$, and GD does at large $\gamma$. Thus, the budget trend persists in the linear setting with GD as the baseline.

\subsection{Connection to practice}
\label{sec:rho-experiments}
We test whether the budget trend predicted by \cref{thm:ratio} also appears in real language models, beyond the assumptions used in our theoretical analysis.

\noindent\textbf{Setup.} We evaluate the AdamW and Muon models trained in \cref{sec:ft}, spanning five model sizes, on held-out DCLM text. We compute the MSR at each token position and report the median over correctly predicted positions. At each position, $h$ and $\kappa$ are the mean and standard deviation of the margins between the correct next-token logit and all other vocabulary logits. We use the median to reduce sensitivity to outliers; \cref{def:rho} instead uses the mean margin divided by the square root of the mean variance.

\begin{figure}[H]
\centering
\vspace{-6pt}
\includegraphics[width=\figwidth]{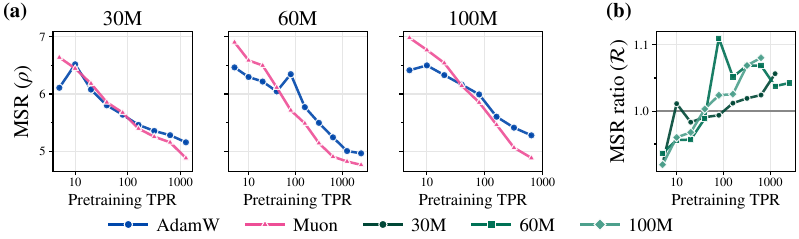}
\vspace{-2pt}
\caption{\textbf{In real language models, the margin-to-spread ratio of AdamW gains on Muon's as the token budget grows.} \textbf{(a)} MSR $\rho$ of AdamW and Muon against pretraining TPR for 30M, 60M, and 100M models: Muon has the larger MSR at small budgets and AdamW at large ones. \textbf{(b)} The MSR ratio $\rho^{\mathrm{AdamW}}/\rho^{\mathrm{Muon}}$ crosses one as the budget grows, at every model size.}
\label{fig:lm-budget}
\vspace{-16pt}
\end{figure}

\noindent\textbf{Experimental Result.} \Cref{fig:lm-budget} plots the MSR for the three model sizes with the widest budget ranges, from 30M to 100M. Muon has a larger MSR at small data budgets, while AdamW has a larger MSR at sufficiently large budgets. As the budget grows, the difference between AdamW's and Muon's MSRs increases, and the ratio of AdamW's MSR to Muon's crosses one. This shift toward AdamW is consistent with the theoretical budget trend and parallels the reversal in relative robustness to fine-tuning observed in \cref{sec:ft}.

The MSR and fine-tuning crossovers do not necessarily appear at exactly the same budget, since the learning--forgetting tradeoff also depends on pretraining loss. Nevertheless, the MSR captures the direction of the observed shift and connects our theoretical analysis to the behavior of trained language models.

\begin{cleanbox}
\noindent\textbf{Remark.} We summarize the analysis of this section and relate MSR to other metrics used to predict post-training.
\begin{itemize}[leftmargin=*, itemsep=3pt, topsep=2pt]
  \item \textbf{Takeaway.} In a linear associative memory, we prove that after one update the ratio $\rho^{\mathrm{GD}}/\rho^{\mathrm{Muon}}$ of the two optimizers' MSRs increases with the data budget (\cref{thm:ratio}). Simulations of many training steps show that the same trend persists after extended training (\cref{sec:budget-experiments}).
  \item \textbf{Extension: Relation of MSR to Other Metrics Predicting Post-Training.} Other work predicts forgetting after post-training with different metrics. For example, \citet{overtrained} show that the loss increase after adding Gaussian noise to the pretrained weights tracks how much a model degrades under fine-tuning. In \cref{app:other-metrics}, we show that MSR is connected to this robustness to Gaussian perturbations: in the linear memory, a larger MSR provably guarantees that more competitors stay ranked below the correct answer after the perturbation (\cref{thm:auc}), and in both language models and the linear memory, MSR is strongly correlated with perturbation robustness (\cref{app:noise-mnr-empirical}).
\end{itemize}
\end{cleanbox}

\section{Related Work}
\label{sec:related}

\noindent\textbf{Muon.} Muon, an optimizer introduced by \citet{jordan2024muon}, has recently shown strong empirical performance for pretraining LLMs, challenging AdamW's dominance as the status-quo optimizer. Recent work has sought to explain Muon's pretraining advantage through analyses of simplified theoretical models, associative memory, convergence behavior, and loss landscape geometry \citep{kim2026sharpcapacityscalingspectral, li2026muonassociativememorylearning, sato2026convergenceboundcriticalbatch, su2025isotropiccurvaturemodelunderstanding, shen2026convergenceanalysismuon, wang2026muonoutperformsadamcurvature}. Another line of work investigates the training dynamics of Muon, finding that it learns data components more uniformly than AdamW \citep{vasudeva2025muonsspectraldesignbenefits, wang2025muonoutperformsadamtailend, dragutinovic2026useusemuonsimplicity}, while also characterizing the implicit biases of Muon \citep{fan2025implicitbiasspectraldescent, dragutinovic2026useusemuonsimplicity}. Yet fewer works have examined the downstream consequences of Muon. \citet{muon_robust} compare Muon and AdamW after fine-tuning; they find that for small data regimes, Muon outperforms AdamW, consistent with our findings. We show that this advantage does not generalize across pretraining data scale; after a sufficiently large pretraining data budget, the trend reverses, and AdamW outperforms Muon.

\noindent\textbf{Muon versus AdamW on pretraining loss.} Muon has been reported to reach lower pretraining loss than AdamW for the same compute across a wide range of settings \citep{liu2025muonscalablellmtraining, shah2025practicalefficiencymuon, mehta2025muontradeoffs}, and frontier models have since been pretrained with Muon or its variants \citep{kimiteam2025kimik2, glm2025glm45}. The size of this advantage depends on scale: under controlled tuning, Muon's speedup over AdamW shrinks as model size and data grow, and Muon is overtaken by other matrix-based optimizers at large data-to-model ratios \citep{wen2025fantasticpretrainingoptimizers, wen2026hyperball}. All of these works measure pretraining loss. Optimizers can also reach qualitatively different solutions rather than faster or slower routes to the same minimum \citep{pascanu2025optimizersalter}; we ask whether these differences matter after fine-tuning.

\noindent\textbf{Catastrophic forgetting.} Catastrophic forgetting, the degradation of previously learned knowledge during subsequent training, remains a central challenge in deep learning \citep{goodfellow2013empirical}. Numerous approaches have been proposed to mitigate catastrophic forgetting, including regularization-based methods \citep{kirkpatrick2017overcoming, zenke2017continual, aljundi2018memory}, replay techniques \citep{shin2017continual}, gradient projection methods \citep{lopez2017gradient, chaudhry2018efficient}, and optimization-based approaches \citep{sam_forgetting}. Our learning--forgetting tradeoff is an instance of the stability--plasticity dilemma, with fine-tuning performance capturing plasticity and knowledge retention capturing stability \citep{lopez2017gradient}. We study forgetting as an intrinsic property of the pretrained checkpoint, rather than as a consequence of the fine-tuning algorithm. Specifically, we investigate how optimization choices during pretraining shape downstream forgetting, showing that Muon and AdamW induce different learning--forgetting tradeoffs.

\noindent\textbf{Pretraining and downstream adaptation.} Better pretraining loss does not always yield a better model after fine-tuning. \citet{overtrained} show that extending pretraining can make language models harder to fine-tune, as the pretrained parameters become increasingly sensitive to modifications, which they measure in part with Gaussian perturbations. \citet{sam_forgetting} show that sharpness-aware pretraining mitigates forgetting during fine-tuning, and \citet{liu2026optimizermodelconsistencyfinetuningoptimizer} find that fine-tuning with the same optimizer as pretraining forgets less. We study how the choice of pretraining optimizer interacts with the pretraining data budget, and connect the resulting fine-tuning gap to robustness to weight perturbations.

\section{Conclusion}
\label{sec:conclusion}

We asked whether the quality of Muon base models persists through fine-tuning, and found that the answer depends on the pretraining budget: beyond a crossover budget, AdamW outperforms Muon after fine-tuning, even though Muon reaches lower pretraining loss. We do not necessarily expect this for every fine-tuning distribution, since fine-tuning on data close to the pretraining data is simply continued pretraining. Still, the trend holds broadly on all datasets we study, including math, instruction following, and music. Furthermore, the crossover budget decreases with model size, which suggests that the problem persists as we scale.

We explain why AdamW outperforms Muon at large budgets by tracking how robust each model is to small isotropic perturbations of its output distribution, which we characterize with the margin-to-spread ratio (MSR). AdamW becomes increasingly more robust than Muon as the data budget grows, and this robustness is a property of the output distribution that the training loss does not capture. Since the MSR measures robustness to perturbations of the output distribution in general, our theory suggests that this robustness is fundamental to the checkpoint learned by each optimizer, rather than specific to fine-tuning, to the perturbations we study, or to a particular architecture or parameterization. It should therefore apply to any update that changes the output distribution.

Although outside the scope of this work, our analysis suggests that pretraining should explicitly optimize for properties beyond the pretraining loss. Optimizer selection requires more than comparing pretraining loss, and we can design objectives that explicitly target adaptability, for which the MSR is a potential candidate.

More broadly, we expect this lesson to extend beyond Muon. Many pretraining choices, such as hyperparameters, data, and architecture, are selected by pretraining loss. Yet, our results show that pretraining loss can be a poor proxy for adaptability, increasingly so as the data budget grows. This suggests that for any of these choices, a lower pretraining loss does not by itself yield a model that can better adapt.

\section*{Acknowledgements}
\label{sec:ack}
We gratefully acknowledge support from Apple, Google, Jane Street, the National Science Foundation and the FLAME cluster at Carnegie Mellon University.

This material is based upon work supported by the National Science Foundation Graduate Research Fellowship under Grant No. DGE2140739. Any opinions, findings, conclusions, or recommendations expressed in this material are those of the authors and do not necessarily reflect the views of the National Science Foundation.

The authors thank Kaiyue Wen, Alex Damian, Ishaan Watts, Lawrence Feng, and Gene Yang for helpful discussions.

\clearpage
\bibliographystyle{labtemplate/colm}
\bibliography{references}

\clearpage
\appendix

\startcontents[appendix]
\printcontents[appendix]{}{1}{\setcounter{tocdepth}{2}\section*{Appendix Contents}}
\vspace{1em}
\raggedbottom
\setcounter{topnumber}{4}\setcounter{bottomnumber}{4}\setcounter{totalnumber}{6}
\renewcommand{\topfraction}{0.95}\renewcommand{\bottomfraction}{0.95}
\renewcommand{\textfraction}{0.05}\renewcommand{\floatpagefraction}{0.8}
\makeatletter\setlength{\@fptop}{0pt}\setlength{\@fpsep}{12pt plus 2pt}\setlength{\@fpbot}{0pt plus 1fil}\makeatother
\newpage
\section{Definitions}
\label{app:definitions}
\label{app:optimizers}

\subsection{Optimizers} Optimizers update a model's parameters during training so as to decrease a loss. Given the gradient of the loss with respect to the parameters, an optimizer determines the direction and the magnitude of each update, and in doing so shapes training stability, convergence speed and the solution that is finally reached. This section recalls the two optimizers compared in the paper.

\subsubsection{AdamW}
\label{sec:adamw-def}

Adam \citep{kingma2014adam} combines momentum with a per-coordinate step size. With $g_t = \nabla_\theta \mathcal{L}(\theta_t)$ the gradient at step $t$, it maintains exponential moving averages of the first and second moments of the gradient,
\begin{equation*}
  \textstyle
  m_t = \beta_1 m_{t-1} + (1 - \beta_1) g_t , \qquad
  v_t = \beta_2 v_{t-1} + (1 - \beta_2) g_t^2 ,
\end{equation*}
with the square taken coordinatewise, corrects their bias towards the zero initialization,
\begin{equation*}
  \textstyle
  \hat m_t = \frac{m_t}{1 - \beta_1^t} , \qquad \hat v_t = \frac{v_t}{1 - \beta_2^t} ,
\end{equation*}
and updates
\begin{equation*}
  \textstyle
  \theta_{t+1} = \theta_t - \alpha \frac{\hat m_t}{\sqrt{\hat v_t} + \epsilon} ,
\end{equation*}
where $\alpha$ is the learning rate, $\beta_1, \beta_2 \in [0, 1)$ are the decay rates of the two averages, $\epsilon > 0$ guards the division, and the division and the square root are coordinatewise. In effect each coordinate is normalized by the scale of its own recent gradients.

Regularizing Adam with an $\ell_2$ penalty $\tfrac{\lambda}{2} \lVert \theta \rVert^2$ adds $\lambda \theta_t$ to $g_t$, so the penalty is itself divided by $\sqrt{\hat v_t} + \epsilon$ and coordinates with large past gradients are regularized less. AdamW \citep{loshchilov2019adamw} instead decouples the weight decay from the adaptive step,
\begin{equation*}
  \textstyle
  \theta_{t+1} = \theta_t - \alpha \frac{\hat m_t}{\sqrt{\hat v_t} + \epsilon} - \alpha \lambda \theta_t ,
\end{equation*}
with $g_t$ now the gradient of the unregularized loss and $\lambda$ the weight decay coefficient, so that every coordinate decays at the same rate. AdamW is the standard optimizer for language-model pretraining and the baseline throughout the paper.

\subsubsection{Muon}
\label{sec:muon-def}

Muon \citep{jordan2024muon} treats each weight matrix as a whole rather than coordinate by coordinate. For a matrix-valued parameter $\theta$ with gradient $G_t$, it accumulates momentum $M_t = \mu M_{t-1} + G_t$ and updates along the orthogonalized momentum,
\begin{equation}
  \textstyle
  \theta_{t+1} = \theta_t - \eta \, \mathrm{Orth}(M_t) ,
  \label{eq:muon-app}
\end{equation}
where $\eta$ is the learning rate and, for a matrix $X$ with singular value decomposition $X = U \Sigma V^\top$, $\mathrm{Orth}(X) = U V^\top$ is its polar factor: the matrix with the singular vectors of $X$ and all singular values set to one. The update therefore has the same size along every singular direction, and it can be read as the steepest descent direction when the size of a step is measured in the spectral norm \citep{jordan2024muon}. An exact singular value decomposition at every step is expensive, so $\mathrm{Orth}$ is approximated by a few steps of a Newton--Schulz iteration,
\begin{equation*}
  \textstyle
  X_0 = \frac{M_t}{\lVert M_t \rVert_F} , \qquad
  X_{k+1} = a X_k + b \, (X_k X_k^\top) X_k + c \, (X_k X_k^\top)^2 X_k ,
\end{equation*}
whose odd matrix polynomial pushes every singular value of $X_0$ towards one; the reference implementation uses $(a, b, c) = (3.4445, -4.7750, 2.0315)$ and five steps. Muon is meant for the two-dimensional weight matrices of the hidden layers; embeddings, the output layer and one-dimensional parameters such as normalization gains are updated by AdamW. Large-scale implementations further add decoupled weight decay and rescale the update to the root-mean-square size of AdamW's \citep{liu2025muonscalablellmtraining}. In our experiments the hidden matrices are updated by Muon and all other parameters by AdamW, each with its own learning rate and weight decay (\cref{sec:setup}).

\subsection{Learning-rate schedules}
\subsubsection{Warmup--stable--decay} All pretraining runs use a warmup--stable--decay (WSD) schedule: the learning rate increases linearly from zero to its peak value over the warmup steps, stays at the peak for most of training, and is annealed to zero over the final decay phase.

\subsubsection{Cosine} All fine-tuning runs use a cosine schedule: after a linear warmup of $20$ steps, the learning rate follows $\eta_t = \tfrac12\eta_{\max}\bigl(1+\cos(\pi t/T)\bigr)$ over the remaining $T$ steps.

\subsection{Token--parameter ratio}
\label{app:tpr}
We measure the pretraining data budget of a model with $N$ parameters trained on $D$ tokens by its token--parameter ratio, $\mathrm{TPR} = D/N$. The compute-optimal budget of \citet{hoffmann2022training} corresponds to $\mathrm{TPR} = 20$; our budgets range from $5$ to $2{,}560$ TPR.

\FloatBarrier
\section{Other Metrics for Analyzing Post-Training}
\label{app:other-metrics}

Besides the margin-to-spread ratio, the robustness of a base model to post-training can be probed by perturbing its weights directly. This section reports Gaussian perturbation results in language models and in the linear memory (\cref{app:gaussian-perturbation}), and then relates them to the MSR (\cref{app:noise-mnr}).

\subsection{Gaussian perturbation}
\label{app:gaussian-perturbation}
We first check whether Gaussian perturbations of the weights reproduce the budget trend of the learning--forgetting tradeoff, in language models and in the linear memory.

\subsubsection{Language models}
\label{app:perturb-lm}
We test whether the perturbed loss of the pretrained language models shifts from Muon toward AdamW as the token budget grows, as the learning--forgetting tradeoff does.

\noindent\textbf{Setup.} \citet{overtrained} show that adding isotropic Gaussian noise to pretrained weights reproduces key empirical trends in forgetting after fine-tuning. Motivated by this connection, we study Gaussian weight perturbations as a simpler setting in which to investigate why the learning--forgetting gap between Muon and AdamW changes with pretraining budget. At the same checkpoints as in \cref{sec:ft}, we add isotropic Gaussian noise to each weight matrix, scaled to a fraction $\delta$ of the matrix's norm, and measure the loss increase $\Delta_{\mathrm{noise}}(\thetapt;\delta):=\losspt(\widetilde\theta_\delta)-\losspt(\thetapt)$, where $\widetilde\theta_\delta$ denotes the perturbed weights; a smaller increase indicates greater perturbation robustness. See \Cref{app:perturb-gammas} for full experimental details.

\begin{figure}[H]
\centering
\vspace{-10pt}
\includegraphics[width=\figwidth]{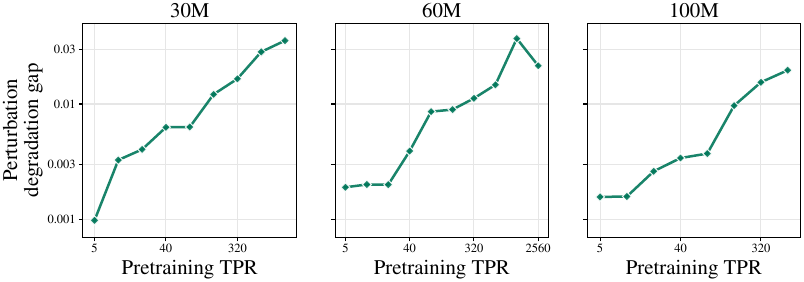}
\vspace{-2pt}
\caption{\textbf{The perturbation degradation gap grows with pretraining data budget.} Under a perturbation of magnitude $\delta = 0.02$, Muon degrades increasingly more than AdamW as the pretraining data budget grows. Points show the mean over three noise draws.}
\label{fig:perturb-lines}
\vspace{-10pt}
\end{figure}

\noindent\textbf{Result.} \Cref{fig:perturb-lines} plots the gap in loss degradation, $\Delta_{\mathrm{noise}}(\text{Muon}; \delta) - \Delta_{\mathrm{noise}}(\text{AdamW}; \delta)$. This gap is positive across the token budgets tested and grows with the pretraining budget: AdamW incurs a smaller loss increase throughout the sweep, and its advantage grows with budget.

To connect this gap to the learning--forgetting tradeoff, we also account for pretraining loss: loss after perturbation equals pretraining loss plus $\Delta_{\mathrm{noise}}$. Muon starts with lower pretraining loss, so it can retain lower loss after perturbation at small budgets despite its larger loss increase. As its pretraining-loss advantage shrinks and the degradation gap grows, AdamW achieves lower loss after perturbation beyond a crossover budget. For sufficiently strong perturbations, this crossover occurs at smaller token budgets as model size increases, paralleling the crossover observed after fine-tuning (\cref{tab:perturb-crossover} in \cref{app:perturb-gammas}).

\subsubsection{Linear associative memory}
\label{app:perturb-toy}
We repeat the perturbation experiment in the linear memory, where the effect of a perturbation can be measured fact by fact.

\noindent\textbf{Setup.} We train the memory of \cref{sec:rho-setup} with GD and Muon to minimize cross-entropy, varying the data budget $\gamma=N/d$ and the dimension $d$, as in \cref{sec:budget-experiments}. The trained weights are perturbed to $W' = W + \sigma \lVert W \rVert_F \xi$ with $\lVert \xi \rVert_F = 1$, over $16$ perturbation directions shared by the two optimizers. Since the memory retrieves facts, we measure degradation by the fraction of initially recovered facts that are lost. At a single perturbation magnitude, fact loss can be close to zero or one, making memories with different noise tolerances difficult to distinguish. We therefore compare the perturbation strengths needed to reach a common intermediate level of degradation: $\sigma_{50}$ is the relative perturbation strength at which half of the initially recovered facts are lost. A larger $\sigma_{50}$ means that more noise is needed to cause the same fraction of facts to be lost. Data budgets past the memory's capacity, where the two optimizers recover different sets of facts, are excluded. Full details appear in \cref{app:rho-extra,app:budget-extra}.

\noindent\textbf{Result.} The robustness ordering of the two optimizers turns over with the data budget: Muon tolerates stronger perturbations at small $\gamma$, and GD does at large $\gamma$ (\cref{fig:budget-main}b and the bottom row of \cref{fig:budget-hk-app}). The $\sigma_{50}$ ordering flips at $\gamma \approx 0.5$, $1.3$ and $11$ at $d = 128$, $256$ and $512$, and at $d = 64$ GD is already the more robust at the smallest data budget we ran.

\subsection{Gaussian perturbation and the margin-to-spread ratio}
\label{app:noise-mnr}
We next relate robustness to Gaussian perturbations to the MSR, first in theory and then in experiments.

\subsubsection{Theoretical analysis}
\label{app:noise-mnr-theory}

We use the scores $s_{ij}(W)=v_j^\top Wu_i$ of \cref{sec:rho-setup} and the margins $m_{ij}$, $h_i$, $\kappa_i$ and $\rho_i$ of \cref{def:rho}, for arbitrary keys, values and $W$. The weights are perturbed to $W+\sigma E$, where the entries of $E$ satisfy
\begin{equation}
  \mathbb{E}[E_{kl}]=0,\qquad \mathbb{E}[E_{kl}E_{k'l'}]=\mathbf{1}\{(k,l)=(k',l')\}.
  \label{eq:noise-moments}
\end{equation}
Scores are linear in $W$, so exactly
\[
  s_{ii}(W+\sigma E)-s_{ij}(W+\sigma E)=m_{ij}+\sigma\,(v_i-v_j)^\top E u_i .
\]
We measure robustness by the fraction of competitors still ranked below the correct value,
\[
  \operatorname{AUC}_i(\sigma)=\operatorname*{mean}_{j\ne i}\,\Pr_E\big[s_{ii}(W+\sigma E)>s_{ij}(W+\sigma E)\big].
\]
Let $J$ be uniform on $\{1,\dots,N\}\setminus\{i\}$, independent of $E$. Then
\[
  \operatorname{AUC}_i(\sigma)
  =\Pr\big[s_{ii}(W+\sigma E)-s_{iJ}(W+\sigma E)>0\big]
  =\Pr\big[m_{iJ}+\sigma(v_i-v_J)^\top Eu_i>0\big],
\]
so it suffices to bound the probability that one random variable is positive, from its mean and variance.

\begin{lmmbox}{Cantelli}{lem:cantelli}
If $\mathbb{E}X>0$ and $\operatorname{Var}X<\infty$, then
\[
  \Pr[X\le 0]\;\le\;\frac{\operatorname{Var}X}{\operatorname{Var}X+(\mathbb{E}X)^2}.
\]
\end{lmmbox}
\begin{proof}
For $t\ge0$, Markov's inequality gives $\Pr[X\le0]\le \mathbb{E}(\mathbb{E}X-X+t)^2/(\mathbb{E}X+t)^2=(\operatorname{Var}X+t^2)/(\mathbb{E}X+t)^2$; take $t=\operatorname{Var}X/\mathbb{E}X$.
\end{proof}

\begin{lmmbox}{Noise adds variance, not bias}{lem:noise-moments}
Let $E$ satisfy \cref{eq:noise-moments}. Then
\begin{align*}
  \mathbb{E}\big[m_{iJ}+\sigma(v_i-v_J)^\top Eu_i\big]&=h_i,\\
  \operatorname{Var}\big[m_{iJ}+\sigma(v_i-v_J)^\top Eu_i\big]&=\kappa_i^2+\sigma^2\lVert u_i\rVert^2\operatorname*{mean}_{j\ne i}\lVert v_i-v_j\rVert^2 .
\end{align*}
\end{lmmbox}
\begin{proof}
For fixed $j$, $(v_i-v_j)^\top Eu_i=\sum_{k,l}(v_i-v_j)_kE_{kl}(u_i)_l$ has, by \cref{eq:noise-moments}, mean $0$ and second moment $\lVert v_i-v_j\rVert^2\lVert u_i\rVert^2$. Average over $E$ first, then over $J$.
\end{proof}

\begin{thmbox}{Perturbed AUC and the margin-to-spread ratio}{thm:auc}
Let $E$ satisfy \cref{eq:noise-moments}. For every $W$, all data, every $\sigma\ge0$ and every $i$ with $h_i>0$,
\[
  \operatorname{AUC}_i(\sigma)\;\ge\;\frac{h_i^2}{h_i^2+\kappa_i^2+\sigma^2\lVert u_i\rVert^2\operatorname*{mean}_{j\ne i}\lVert v_i-v_j\rVert^2}.
\]
\end{thmbox}
\begin{proof}
$1-\operatorname{AUC}_i(\sigma)=\Pr\big[m_{iJ}+\sigma(v_i-v_J)^\top Eu_i\le 0\big]$. Apply \cref{lem:cantelli} with the mean and variance of \cref{lem:noise-moments}.
\end{proof}

No assumption is made on $W$, on the shape of the margin distribution, or on $d$ and $N$.

\begin{corbox}{Gaussian data}{cor:auc-gauss}
If moreover $u_1,\dots,u_N,v_1,\dots,v_N$ are i.i.d.\ $\mathcal{N}(0,I_d)$, then
\[
  \operatorname{AUC}_i(\sigma)\;\ge\;\frac{h_i^2}{h_i^2+\kappa_i^2+2\sigma^2d^2\big(1+O_p(d^{-1/2})\big)}
  \;=\;\frac{\rho_i^2}{\rho_i^2+1+2(\sigma d/\kappa_i)^2\big(1+O_p(d^{-1/2})\big)}.
\]
\end{corbox}
\begin{proof}
$\lVert u_i\rVert^2=d\big(1+O_p(d^{-1/2})\big)$ and $\operatorname*{mean}_{j\ne i}\lVert v_i-v_j\rVert^2=2d\big(1+O_p(d^{-1/2})\big)$.
\end{proof}

At a fixed noise level in units of $\kappa_i$, a larger $\rho_i$ therefore guarantees a larger AUC. For the relative perturbation $W+\delta\lVert W\rVert_F E/\lVert E\rVert_F$ used in the experiments, with i.i.d.\ standard Gaussian entries of $E$, the bounds hold with $\sigma$ replaced by $\delta\lVert W\rVert_F/d$.

\begin{corbox}{Noise on the logits}{cor:auc-logit}
Suppose instead that each logit is perturbed directly, to $s_{ij}(W)+\sigma\xi_{ij}$, where the $\xi_{ij}$ have mean $0$ and variance $1$ and are uncorrelated (for example, i.i.d.\ $\mathcal{N}(0,1)$). Then for every $W$, every $\sigma\geq0$ and every $i$ with $h_i>0$,
\[
  \operatorname*{mean}_{j\ne i}\Pr_\xi\big[s_{ii}(W)+\sigma\xi_{ii}>s_{ij}(W)+\sigma\xi_{ij}\big]
  \;\geq\;\frac{h_i^2}{h_i^2+\kappa_i^2+2\sigma^2}
  \;=\;\frac{\rho_i^2}{\rho_i^2+1+2\sigma^2/\kappa_i^2}.
\]
\end{corbox}
\begin{proof}
With $J$ uniform on $\{1,\dots,N\}\setminus\{i\}$ and independent of $\xi$, the variable $m_{iJ}+\sigma(\xi_{ii}-\xi_{iJ})$ has mean $h_i$ and variance $\kappa_i^2+2\sigma^2$, since $\xi_{ii}-\xi_{ij}$ has mean $0$ and variance $2$ for every $j\neq i$. Apply \cref{lem:cantelli}.
\end{proof}

A fact is mislabeled only if some incorrect answer overtakes the correct one, so by a union bound the probability of mislabeling fact $i$ is at most $N-1$ times one minus the left-hand side. This bound is informative only when $\rho_i^2$ is much larger than $N$.

\subsubsection{Empirical validation}
\label{app:noise-mnr-empirical}
We test whether the MSR predicts how much perturbation a checkpoint can tolerate, first in language models and then in the linear memory.

\begin{figure}[H]
\centering
\vspace{-10pt}
\includegraphics[width=\figwidth]{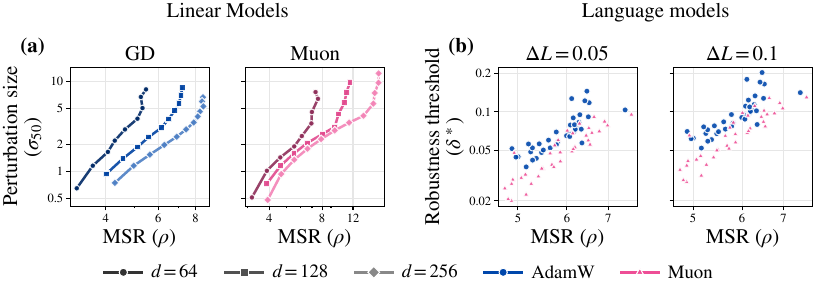}
\vspace{-2pt}
\caption{\textbf{The margin-to-spread ratio tracks robustness to weight perturbations.} \textbf{(a)} For the linear models trained under GD and Muon, we plot the perturbation $\sigma_{50}$ required to lose half of the recovered facts against its MSR, $\bar h/\bar\kappa$ (\cref{def:rho}). A higher MSR requires more noise to cause the same loss. \textbf{(b)} We measure the MSR of $72$ pretrained language models (30M--600M) as the median of $h/\kappa$ over correctly predicted tokens, and plot the perturbation $\delta^\ast$ at which held-out loss rises by $\Delta L$. The same relationship holds.}
\label{fig:rho-validation}
\vspace{-15pt}
\end{figure}

\noindent\textbf{Language models.} We measure MSR as in \cref{sec:rho-experiments} and degradation by the held-out loss increase of \cref{app:perturb-lm}. The robustness threshold $\delta^\ast$ is the perturbation magnitude at which this increase reaches $\Delta L$ nats. Across checkpoints, $\rho$ and $\delta^\ast$ are strongly correlated for both $\Delta L=0.05$ and $0.1$ (\cref{fig:rho-validation}b): the Spearman correlation over the $72$ checkpoints is $+0.83$ at both thresholds, and within each optimizer it is $+0.87$ to $+0.89$ for AdamW and $+0.92$ to $+0.93$ for Muon. Full experimental details appear in \cref{app:rho-extra}.

\noindent\textbf{Linear associative memory.} With $\sigma_{50}$ as in \cref{app:perturb-toy}, $\sigma_{50}$ is strongly correlated with MSR for both optimizers (\cref{fig:rho-validation}a). Within each optimizer, $\sigma_{50}$ and $\rho$ move together at every model dimension (\cref{fig:app-hk}).

\FloatBarrier
\section{Experimental Details}
\label{app:experimental-details}
\label{app:pretraining}

\subsection{Pretraining}
\label{app:pt-setup}

\paragraph{Models and data.} All language models are OLMo-style decoder-only transformers \citep{groeneveld-etal-2024-olmo, olmo20242olmo2furious} with 30M, 60M, 100M, 300M and 600M parameters, trained on DCLM \citep{li2024datacomplm} with the OLMo~2 tokenizer. Budgets span $5$--$1{,}280$ TPR at 30M, $5$--$2{,}560$ TPR at 60M, $5$--$640$ TPR at 100M, $5$--$160$ TPR at 300M and $5$--$40$ TPR at 600M.

\paragraph{Optimization.} Every model is pretrained with sequence length $4096$, a global batch of $2^{20}$ tokens, a warmup--stable--decay (WSD) schedule, weight decay $0.1$ and gradient clipping at $1.0$. AdamW uses $(\beta_1,\beta_2) = (0.9, 0.98)$. Muon updates the hidden weight matrices; the remaining parameters (embeddings, output layer and norms) are updated by AdamW.

\subsubsection{Lowest-pretraining-loss setting}
\label{app:lr-tuning}
\label{app:lowest-loss}
For each model size, budget and optimizer, we sweep the peak learning rate and call the run with the lowest held-out DCLM loss the \emph{loss-optimal} base model. For Muon we sweep the Muon learning rate and fix the learning rate of the AdamW-updated parameters to the tuned AdamW value of the same cell (\cref{app:muon-adamw-lr}). \Cref{tab:tuned-lr} lists the resulting learning rates. In the \emph{lowest-pretraining-loss setting}, used throughout the main text, each optimizer is represented by its loss-optimal base model alone: we fine-tune it over the full sweep of \cref{app:ft-setup} and compute the learning--forgetting frontier from its fine-tuning runs.

\begin{table}[htbp]
\centering
\caption{Tuned peak pretraining learning rates per model size and budget. For Muon, the Muon learning rate; the AdamW-parameter learning rate is fixed to the tuned AdamW value.}
\small
\begin{tabular}{@{}cc|cccccccccc@{}}
\toprule
\textbf{Size} & \textbf{Optimizer} & \multicolumn{10}{c}{\textbf{TPR}} \\
 & & $\mathbf{5}$ & $\mathbf{10}$ & $\mathbf{20}$ & $\mathbf{40}$ & $\mathbf{80}$ & $\mathbf{160}$ & $\mathbf{320}$ & $\mathbf{640}$ & $\mathbf{1{,}280}$ & $\mathbf{2{,}560}$ \\
\midrule
\multirow{2}{*}{30M} & AdamW & $0.08$ & $0.04$ & $0.04$ & $0.014$ & $0.02$ & $0.014$ & $0.02$ & $0.014$ & $0.01$ & -- \\
 & Muon & $0.01$ & $0.01$ & $0.01$ & $0.014$ & $0.02$ & $0.01$ & $0.014$ & $0.01$ & $0.014$ & -- \\
\midrule
\multirow{2}{*}{60M} & AdamW & $0.056$ & $0.04$ & $0.02$ & $0.02$ & $0.01$ & $0.014$ & $0.01$ & $0.01$ & $0.007$ & $0.01$ \\
 & Muon & $0.014$ & $0.01$ & $0.014$ & $0.01$ & $0.014$ & $0.01$ & $0.014$ & $0.007$ & $0.005$ & $0.005$ \\
\midrule
\multirow{2}{*}{100M} & AdamW & $0.04$ & $0.02$ & $0.01$ & $0.01$ & $0.01$ & $0.007$ & $0.007$ & $0.007$ & -- & -- \\
 & Muon & $0.014$ & $0.014$ & $0.014$ & $0.014$ & $0.01$ & $0.014$ & $0.01$ & $0.005$ & -- & -- \\
\midrule
\multirow{2}{*}{300M} & AdamW & $0.001$ & $0.0025$ & $0.0035$ & $0.0025$ & $0.0025$ & $0.0025$ & -- & -- & -- & -- \\
 & Muon & $0.02$ & $0.014$ & $0.02$ & $0.014$ & $0.014$ & $0.02$ & -- & -- & -- & -- \\
\midrule
\multirow{2}{*}{600M} & AdamW & $0.0025$ & $0.0025$ & $0.0025$ & $0.0025$ & -- & -- & -- & -- & -- & -- \\
 & Muon & $0.014$ & $0.014$ & $0.014$ & $0.014$ & -- & -- & -- & -- & -- & -- \\
\bottomrule
\end{tabular}
\label{tab:tuned-lr}
\end{table}

\subsubsection{Learning rate of Muon's AdamW-updated parameters}
\label{app:muon-adamw-lr}
\begin{figure}[htbp]
\centering
\includegraphics{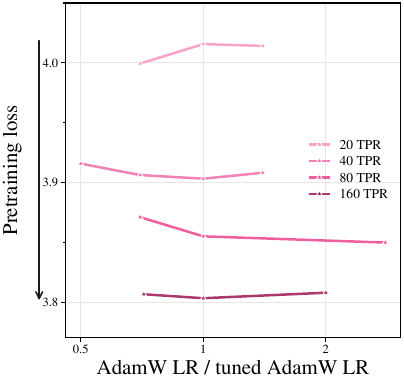}
\caption{\textbf{Muon is insensitive to the learning rate of its AdamW-updated parameters.} Best pretraining loss over Muon learning rates for each AdamW learning rate, relative to the tuned AdamW value (dashed), for 60M models.}
\label{fig:app-muon-adamw-lr}
\end{figure}

Muon has a second learning rate for the parameters it leaves to AdamW. Rather than tuning it in every cell, we set it to the tuned AdamW learning rate of the same cell. To check this choice, we pretrain 60M Muon models at $20$--$160$ TPR over a grid of Muon learning rates and three to four AdamW learning rates between $0.5\times$ and $2.8\times$ the tuned value. \Cref{fig:app-muon-adamw-lr} shows the best loss over Muon learning rates for each AdamW learning rate. It varies by at most $0.021$ nats, and the tuned value is within $0.016$ nats of the best, well below the $0.03$--$0.20$ nats by which Muon beats AdamW at these budgets.

\subsubsection{Best-frontier setting}
\label{app:best-frontier}
The loss-optimal base model is not always the best starting point for fine-tuning: models pretrained with a larger learning rate can offer a better learning--forgetting tradeoff \citep{sam_forgetting}. We therefore also compare the two optimizers in the \emph{best-frontier setting}, which sweeps over both pretraining and fine-tuning configurations. For each model size, budget and pretraining optimizer, we fine-tune the loss-optimal base model and the base models pretrained at the next five larger learning rates of the sweep grid $\{0.005, 0.007, 0.01, 0.014, 0.02, 0.028, 0.04, 0.056, 0.08, 0.12, 0.16\}$, pool all of their fine-tuning runs, and compute a single learning--forgetting frontier from the pooled runs. This frontier is at least as good as the one of the lowest-pretraining-loss setting, since it contains the loss-optimal base model's runs. \Cref{app:results-best-frontier} repeats the main results in this setting.

\subsection{Fine-tuning}
\label{app:ft-setup}
Each base model is fine-tuned for 20M tokens on each of Alpaca \citep{taori2023alpaca}, GSM8K \citep{cobbe2021gsm8k}, MusicPile \citep{yuan2024chatmusician} and StackMathQA \citep{zhang2024stackmathqa}, with sequence length $1024$, a global batch of $65{,}536$ tokens (about $305$ steps), a cosine schedule with $20$ warmup steps, weight decay $0.1$, $(\beta_1,\beta_2) = (0.9, 0.95)$ and gradient clipping at $1.0$. AdamW-pretrained models are fine-tuned with AdamW, and Muon-pretrained models with both AdamW and Muon (\cref{app:ft-optimizer} motivates this choice). We sweep AdamW learning rates from $10^{-4}$ to $10^{-2}$, and Muon learning rates from $6\times10^{-4}$ to $2\times10^{-2}$ with the learning rate of the AdamW-updated parameters set to $0.25\times$ the Muon learning rate. The sweep ranges from runs that barely move the base model to runs that reach the lowest fine-tuning loss at the cost of more than a nat of forgetting.

\subsection{Fine-tuning regularization}
\label{app:ft-reg}
A natural question is whether the reversal survives when fine-tuning itself is regularized to limit forgetting. We test two standard methods for mitigating catastrophic forgetting, data replay \citep{shin2017continual} and elastic weight consolidation \citep[EWC;][]{kirkpatrick2017overcoming}, on the loss-optimal base models of every size. Both keep the fine-tuning recipe and learning-rate sweep of \cref{app:ft-setup}, and each frontier pools all strengths of the regularizer. \Cref{app:ft-reg-results} reports the results.

\paragraph{Replay.} We replace a fraction $r \in \{0.1, 0.2, 0.3\}$ of the fine-tuning tokens with DCLM data drawn from a shard that is disjoint from both the pretraining data and the held-out evaluation shard.

\paragraph{EWC.} EWC penalizes movement away from the pretrained weights in the directions that matter for pretraining. It minimizes
\begin{equation}
  \lossft(\theta) + \frac{\lambda}{2} \sum_k F_k \bigl(\theta_k - \theta_{\mathrm{PT},k}\bigr)^2,
  \label{eq:ewc}
\end{equation}
where $F$ is the diagonal empirical Fisher information of the pretraining loss at $\thetapt$, estimated from squared gradients over $100$ batches of the disjoint DCLM shard used for replay. Following \citet{sam_forgetting}, $F$ is not normalized, so we sweep $\lambda \in \{10^3, 4\cdot10^3, 10^4, 4\cdot10^4\}$.

\subsection{Gaussian perturbations}
\label{app:perturb-setup}
We perturb each base model by adding independent isotropic Gaussian noise to every parameter tensor $W$, scaled so that $\lVert \Delta W \rVert_F = \delta \lVert W \rVert_F$, and measure the perturbation degradation $\Delta_{\mathrm{noise}}(\thetapt;\delta)$ of \cref{app:perturb-lm}. We average over three noise draws per checkpoint and magnitude, use the loss-optimal checkpoint of each cell, and consider magnitudes from $\delta = 0.002$ to $0.56$.

\subsection{Evaluation}
\label{app:evaluation}

\paragraph{Losses.} Forgetting is the loss on a held-out DCLM shard of about $4.19$M tokens that no model is trained on, and learning is the loss on the validation split of the fine-tuning dataset. Both are mean next-token cross-entropy in nats.

\paragraph{Frontiers.} Each fine-tuned checkpoint $\thetaft$ of a base model gives a point $\bigl(\losspt(\thetaft), \lossft(\thetaft)\bigr)$. We compute the learning--forgetting frontier as the lower-left arc of the convex hull of these points, treated as a polyline.

\paragraph{Fine-tuning gap and crossover.} We estimate the fine-tuning gap $G_{\mathrm{FT}}$ of \cref{sec:prelim} from $100$ points sampled uniformly by arc length along the AdamW frontier, each assigned its Euclidean distance to the nearest point of the Muon frontier, negative when that point lies closer to the origin. The crossover TPR is the smallest budget at which $G_{\mathrm{FT}} > 0$. In both settings, the gap and crossover compare the two optimizers' frontiers as defined above: one base model per optimizer in the lowest-pretraining-loss setting, pooled base models in the best-frontier setting.

\paragraph{Degradation.} For \cref{fig:gap-row}c, the degradation of a base model is the increase in held-out DCLM loss caused by fine-tuning, $\losspt(\thetaft) - \losspt(\thetapt)$. We read $\losspt(\thetaft)$ off the base model's frontier at a target fine-tuning loss of $1.1\times$ the lowest fine-tuning loss reached by either optimizer at that model size, so that both optimizers are compared at the same amount of learning.

\FloatBarrier
\section{Additional Results for Section~\ref{sec:ft}}
\label{app:additional-results}

This section extends the fine-tuning results of \cref{sec:ft}. Unless noted otherwise, figures show 60M models, and frontiers are restricted to the region of \cref{app:evaluation}.

\subsection{Lowest-pretraining-loss setting}
\label{app:results-loss-optimal}
\begin{figure}[!htbp]
\centering
\includegraphics[width=\figwidth]{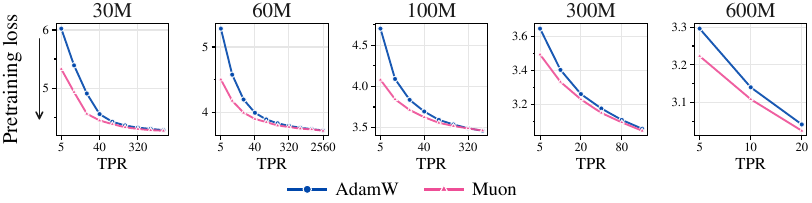}
\caption{\textbf{Muon reaches lower pretraining loss than AdamW.} DCLM held-out loss of the loss-optimal base models across pretraining data budgets and model sizes.}
\label{fig:app-pt-loss}
\end{figure}

\begin{figure}[!htbp]
\centering
\includegraphics[width=\figwidth]{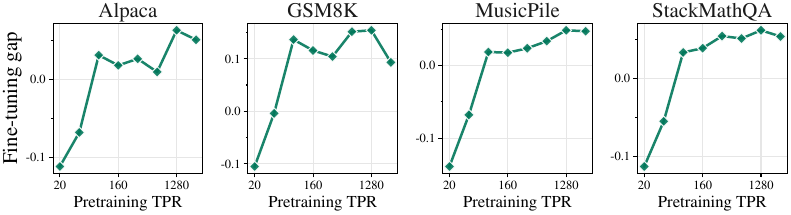}
\caption{\textbf{Fine-tuning gap on every fine-tuning dataset, loss-optimal base models} (60M). Negative values mean the Muon frontier dominates.}
\label{fig:app-gap-all-loss}
\end{figure}

\begin{figure}[!htbp]
\centering
\includegraphics[width=\figwidth]{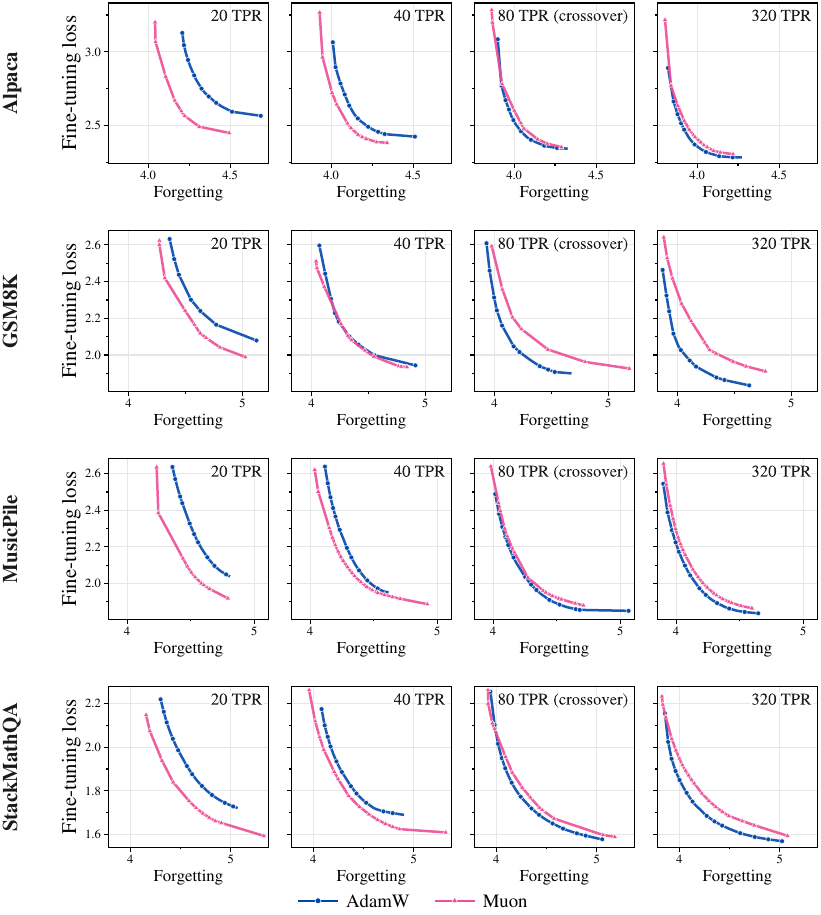}
\caption{\textbf{Learning--forgetting frontiers on every fine-tuning dataset, loss-optimal base models} (60M). Each row shows the budgets around that dataset's crossover TPR.}
\label{fig:app-frontiers-all-loss}
\end{figure}

\Cref{fig:app-pt-loss} shows that the loss-optimal Muon model reaches a lower pretraining loss than its AdamW counterpart at every budget and model size, so the reversal is not inherited from pretraining loss. \Cref{fig:app-frontiers-all-loss,fig:app-gap-all-loss} extend the frontiers and gaps of the lowest-pretraining-loss setting (\cref{fig:combined,fig:gap-row}) to all four datasets, at the budgets around each crossover. On every dataset Muon is better than or level with AdamW below the crossover and AdamW is better beyond it.

\FloatBarrier
\subsection{Best-frontier setting}
\label{app:all-datasets}
\label{app:results-best-frontier}
\begin{figure}[!htbp]
\centering
\includegraphics[width=\figwidth]{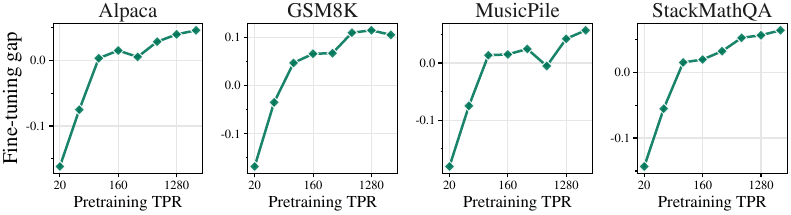}
\caption{\textbf{Fine-tuning gap on every fine-tuning dataset, best-frontier setting} (60M). Negative values mean the Muon frontier dominates.}
\label{fig:app-gap-all-best}
\end{figure}

\begin{figure}[!htbp]
\centering
\includegraphics[width=\figwidth]{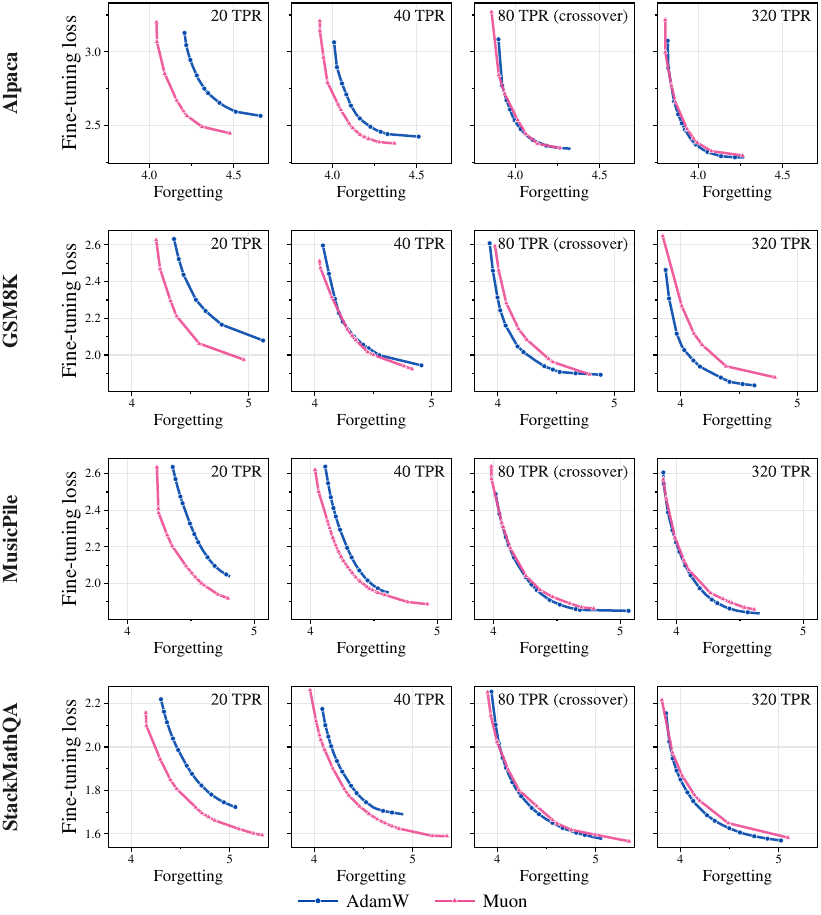}
\caption{\textbf{Learning--forgetting frontiers on every fine-tuning dataset, best-frontier setting} (60M). Each row shows the budgets around that dataset's crossover TPR.}
\label{fig:app-frontiers-all-best}
\end{figure}

\begin{figure}[!htbp]
\centering
\includegraphics[width=\figwidth]{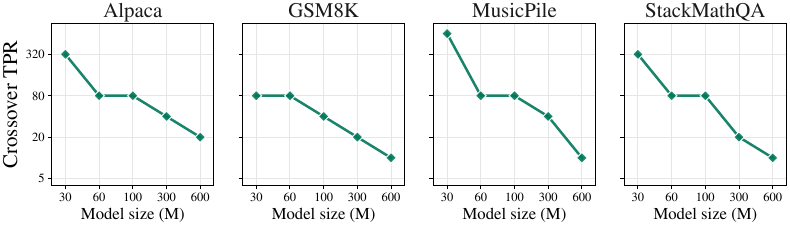}
\caption{\textbf{Crossover TPR decreases with model size, in the best-frontier setting.} As model size increases, Muon's advantage reverses with substantially smaller pretraining data budgets.}
\label{fig:hi5-crossover}
\end{figure}

We repeat the analyses of \cref{sec:ft} in the best-frontier setting (\cref{app:best-frontier}), where each pretraining optimizer is swept across all pretraining and fine-tuning configurations. \Cref{fig:app-frontiers-all-best,fig:app-gap-all-best} show the learning--forgetting frontiers and fine-tuning gaps of 60M models on all four fine-tuning datasets, at the budgets around each crossover. On every dataset, the Muon frontier is better below the crossover; at the crossover AdamW has caught up with or overtaken Muon, and beyond it the AdamW frontier dominates. The fine-tuning gap turns positive at the crossover and stays positive, broadly growing with the data budget. \Cref{fig:hi5-crossover} shows that, as in the lowest-pretraining-loss setting (\cref{fig:crossover}), the crossover TPR decreases with model size.

\FloatBarrier
\subsection{Choice of fine-tuning optimizer}
\label{app:ft-optimizer}
\begin{figure}[htbp]
\centering
\includegraphics[width=\figwidth]{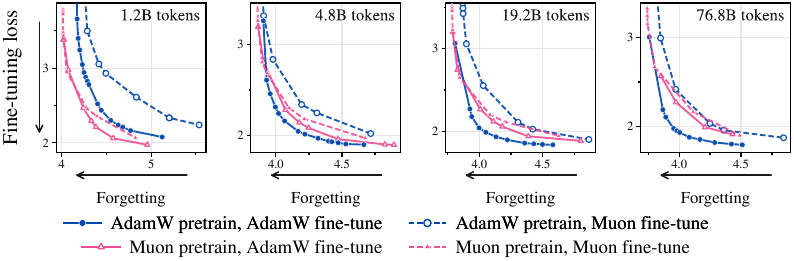}
\caption{\textbf{Muon's greater forgetting comes from the pretrained model, not the fine-tuning optimizer.} Learning--forgetting frontiers of the best 60M base models on GSM8K, split by fine-tuning optimizer (solid: AdamW; dashed: Muon). AdamW fine-tuning is always better for AdamW-pretrained models, while the two fine-tuning optimizers give nearly the same frontier for Muon-pretrained models. At large budgets, the AdamW-pretrained model dominates under either choice.}
\label{fig:app-ft-optimizer}
\end{figure}

Our frontiers fine-tune AdamW-pretrained models with AdamW and Muon-pretrained models with both AdamW and Muon. \Cref{fig:app-ft-optimizer} motivates this choice, and shows that Muon's greater forgetting at large budgets is a property of the pretrained model rather than of the fine-tuning recipe. For each pretraining optimizer and budget, we take the best 60M base model among those fine-tuned with both optimizers, and plot its AdamW- and Muon-fine-tuned frontiers separately. We use GSM8K, the dataset on which AdamW-pretrained models were fine-tuned with both optimizers across the full range of budgets.

Two observations follow. First, the fine-tuning optimizer matters for AdamW-pretrained models but not for Muon-pretrained ones: on AdamW-pretrained models, AdamW fine-tuning dominates Muon fine-tuning at every budget, whereas the two frontiers of a Muon-pretrained model lie close together with no consistent winner. Second, the reversal does not depend on how the Muon-pretrained model is fine-tuned: at $1.2$B tokens both of its frontiers lie below those of the AdamW-pretrained model, and from $19.2$B tokens onward the AdamW-pretrained, AdamW-fine-tuned frontier dominates both.

\FloatBarrier
\subsection{Fine-tuning regularization}
\label{app:ft-reg-results}
\begin{figure}[!htbp]
\centering
\includegraphics[width=\figwidth]{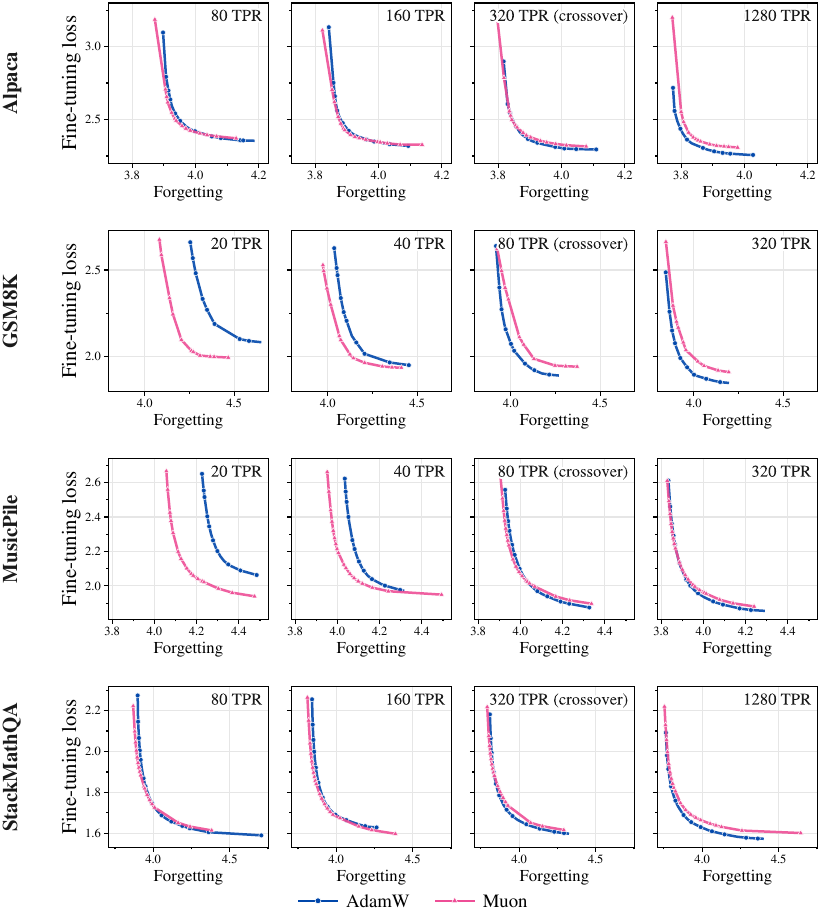}
\caption{\textbf{Learning--forgetting frontiers with replay} of $10$--$30\%$ DCLM data, loss-optimal 60M base models. Each row shows the budgets around that dataset's crossover under replay.}
\label{fig:app-frontiers-replay}
\end{figure}

\begin{figure}[!htbp]
\centering
\includegraphics[width=\figwidth]{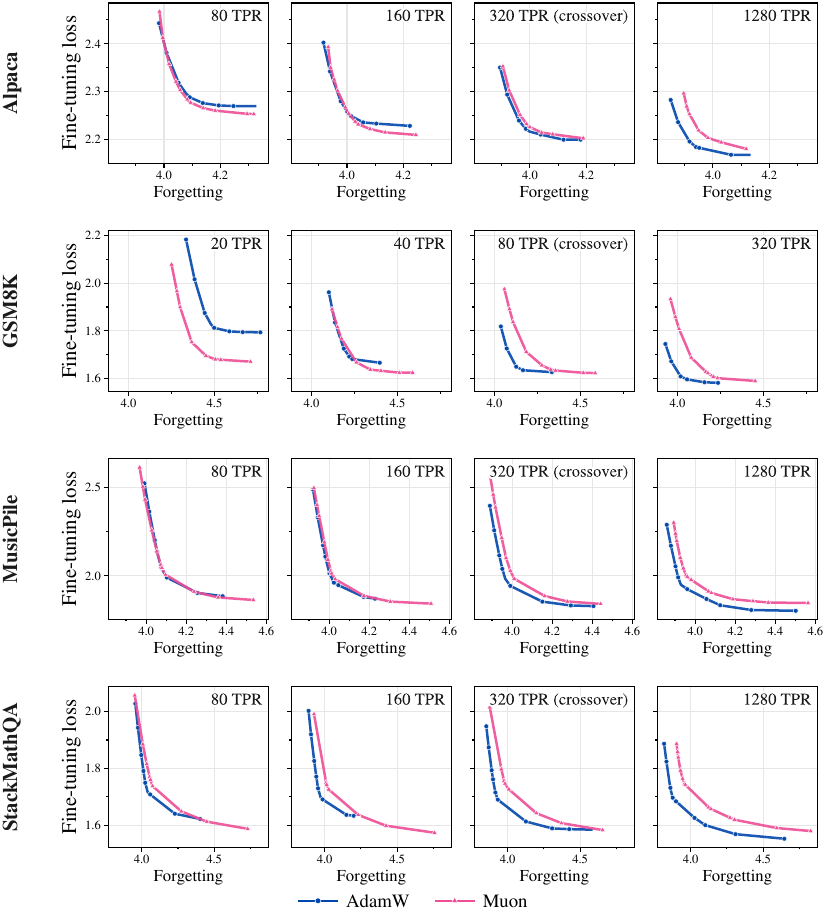}
\caption{\textbf{Learning--forgetting frontiers with EWC}, $\lambda \in \{10^3, 4\cdot10^3, 10^4, 4\cdot10^4\}$, loss-optimal 60M base models. Each row shows the budgets around that dataset's crossover under EWC.}
\label{fig:app-frontiers-ewc}
\end{figure}

\begin{figure}[!htbp]
\centering
\includegraphics[width=\figwidth]{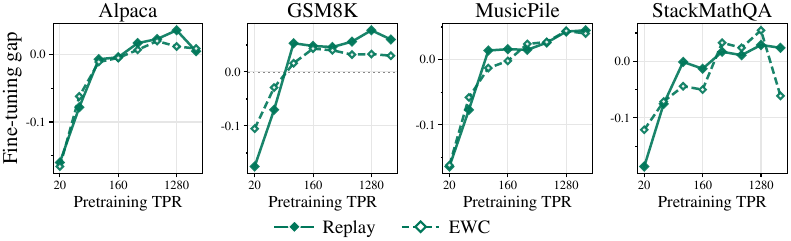}
\caption{\textbf{Fine-tuning gap under fine-tuning regularization}, with replay (solid) and EWC (dashed), loss-optimal 60M base models. Negative values mean the Muon frontier dominates.}
\label{fig:app-gap-reg}
\end{figure}

We repeat the comparison when fine-tuning is regularized with replay or EWC (setup in \cref{app:ft-reg}), using the loss-optimal 60M base models. \Cref{fig:app-frontiers-replay,fig:app-frontiers-ewc} show the frontiers on all four datasets; each row shows the budgets around that dataset's crossover under the regularizer, a quarter and half of the crossover TPR, the crossover itself, and four times it. The crossover is the first budget at which the fine-tuning gap turns positive (\cref{app:evaluation}). Both regularizers reduce forgetting and can move the crossover to larger budgets (from $80$ to $320$ TPR on Alpaca and StackMathQA), but AdamW's frontier is better at large budgets on every dataset under both (\cref{fig:app-gap-reg}), with one exception: under EWC on StackMathQA the gap turns negative again at the largest budget, $2{,}560$ TPR. With replay, Muon's frontier is better than or level with AdamW's below the crossover on every dataset. With EWC the picture below the crossover is mixed: Muon is clearly better at small budgets on GSM8K, while on the other datasets the two frontiers are close.

\FloatBarrier

\FloatBarrier
\section{Additional Results for Section~\ref{sec:analysis} and Appendix~\ref{app:other-metrics}}
\label{app:analysis-results}

This section collects details and additional results for the analysis of \cref{sec:analysis} and for the perturbation experiments of \cref{app:other-metrics}: the margin-to-spread ratio across data budgets in the linear memory (\cref{app:budget-extra}, for \cref{sec:analysis-budget}), perturbation robustness at every magnitude (\cref{app:perturb-gammas}, for \cref{app:gaussian-perturbation}), and the margin-to-spread ratio and robustness in the linear memory and in language models (\cref{app:rho-extra}, for \cref{sec:rho-experiments,app:noise-mnr}).

\subsection{Margin-to-spread ratio across data budgets}
\label{app:budget-extra}

\subsubsection{One step}

This is the regime \cref{thm:ratio} is about: the two solutions are the exact first step from $W_0 = 0$, so every quantity below can be compared against a closed form.

\paragraph{Setup}

\noindent\textbf{Data.} Keys and values are drawn as in \cref{sec:rho-setup} at $d = 512$, with $N = \gamma d$ and $\gamma \in \{\tfrac14, \tfrac12, \tfrac34, 1, \tfrac54, \tfrac32, 2, \tfrac52, 3, 4, 5, 6, 8, 10, 12, 16, 20, 24, 32, 48, 64\}$, five random seeds per data budget. \Cref{sec:rho-setup} requires $N - 1 \geq d$, so the points with $\gamma \leq 1$ lie outside the range in which \cref{thm:ratio} is stated and are drawn only for context.

\noindent\textbf{One step.} The gradient of the loss of \cref{sec:rho-setup} at $W_0 = 0$ is computed exactly, with equal item weights. GD takes it as $W^{\mathrm{GD}}$; Muon takes its polar factor $W^{\mathrm{Muon}} = \operatorname{Orth}(W^{\mathrm{GD}})$, computed exactly by singular value decomposition rather than by the Newton--Schulz iteration of \cref{sec:muon-def}. As a baseline we also record the first AdamW step, which with zero initialization and the usual $\epsilon$ is the entrywise $\operatorname{sign}(W^{\mathrm{GD}})$; \cref{thm:ratio} says nothing about it.

\noindent\textbf{Observables.} Every $h_i$ and $\kappa_i$ is evaluated in closed form for all $i$ as in \cref{def:rho}, giving $\bar h$, $\bar\kappa$, $\rho$ and $\mathcal{R}$.

\noindent\textbf{Plotting.} Solid lines are means over the five seeds and bands are $\pm$ one standard deviation; dashed lines are the closed forms behind \cref{thm:ratio}, with $\Phi(\gamma)$ replaced by its limit $\tfrac{8}{3\pi}$. In the $\bar h$ and $\bar\kappa$ panels each solution is normalized to $\lVert W \rVert_F = 1$, which is needed because the two steps have different scales while $\rho$ does not see them; the $\bar h$ and $\rho$ panels are divided by $\sqrt{d/\gamma}$, the unit in which \cref{thm:ratio} is stated.

\paragraph{Result}

\begin{figure}[htbp]
\centering
\vspace{-4pt}
\includegraphics[width=\figwidth]{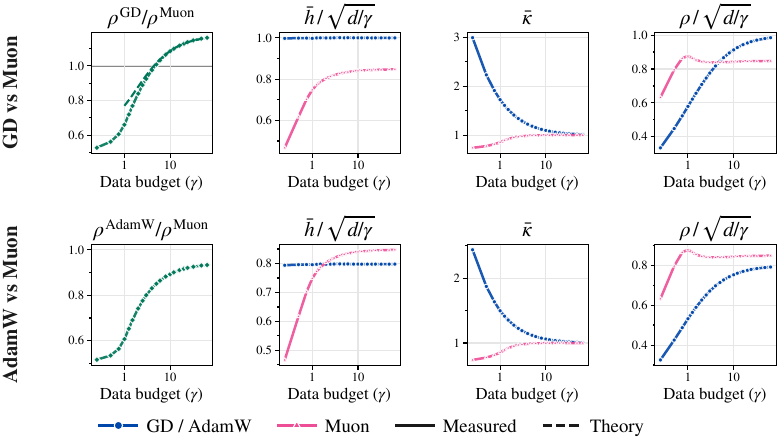}
\vspace{-8pt}
\caption{\textbf{One step at $d = 512$}, GD against Muon (top) and the AdamW surrogate against Muon (bottom); solid is experiment, dashed the closed form of \cref{thm:ratio}.}
\label{fig:budget-onestep-app}
\vspace{-6pt}
\end{figure}

\Cref{fig:budget-onestep-app} (top) shows $\mathcal{R}$ rising monotonically from $0.53$ at $\gamma = \tfrac14$ to $1.16$ at $\gamma = 64$ and crossing $1$ near $\gamma = 5$, with the closed form tracking it once $\gamma \gtrsim 2$. The remaining panels separate the two factors. Normalized, GD's margin is constant in $\gamma$, which is the identity $\bar h^{\mathrm{GD}} = \lVert W^{\mathrm{GD}} \rVert_F^2/(N-1)$ read at $\lVert W \rVert_F = 1$, while Muon's climbs to about $\tfrac{8}{3\pi}$; both variances approach the same limit, GD's from above and Muon's from below. The limiting ratio $\tfrac{3\pi}{8}$ is therefore carried by the margin. Its approach to that limit comes from both factors: Muon's margin contributes the $+\tfrac{1}{10\gamma}$ correction, while the $-\tfrac1\gamma$ part of the $\tfrac{9}{10\gamma}$ term comes from GD's variance, the factor $\gamma+2$ in $(\bar\kappa^{\mathrm{GD}})^2$ (\cref{sec:budget-theory}). The bottom row runs the same comparison against the AdamW surrogate: its $\mathcal{R}$ has the same shape but reaches only $0.93$ at $\gamma = 64$, so over the data budgets we ran the one-step reversal is a statement about GD, not about AdamW.

\subsubsection{Many steps}

The theory covers one step. This experiment runs the same memory to convergence and asks whether the ordering it predicts survives training.

\paragraph{Setup}

\noindent\textbf{Data.} Keys and values are drawn as in \cref{sec:rho-setup}, with item $i$ queried with probability $q_i \propto 1/i$ (Zipf). We use $d \in \{64, 128, 256, 512\}$, $\gamma$ log-spaced in $[\tfrac14, 128]$, and three seeds per $(d, \gamma)$ shared by the two optimizers.

\noindent\textbf{Training.} From $W_0 = 0$ we take $T = 10^4$ full-batch steps of $W_{t+1} = W_t + 0.02\, G_t$ for GD and $W_{t+1} = W_t + 0.005 \operatorname{Orth}(G_t)$ for Muon, where $G_t = \sum_i q_i \big( 1 - \hat p_{W_t}(i \mid i) \big) u_i v_i^\top$ is the Hebbian part of the gradient of the loss of \cref{sec:rho-setup}, at constant learning rate, without momentum or weight decay, and with the polar factor again taken by singular value decomposition.

\noindent\textbf{Observables.} $h_i$ and $\kappa_i$ are as in \cref{def:rho}, with equal item weights rather than the $q_i$ used in training. The saturated value of an observable is the per-seed median over $t \geq 10^3$; $\mathcal{R}$ is formed per seed and only then averaged. Curves are means over seeds and bands are $\pm$ one standard deviation.

\noindent\textbf{Perturbed robustness.} We perturb the trained weights as in \cref{app:perturb-toy}, $W' = W_T + \sigma \lVert W_T \rVert_F \xi$ with $\lVert \xi \rVert_F = 1$, over $16$ perturbation directions shared by the two optimizers, and read off $\sigma_{50}$, the $\sigma$ at which half of the items correct at $\sigma = 0$ are lost. For $N < d$ the Muon solution is rescaled to its component in the span of the keys, since outside that span it is an arbitrary orthogonal completion that carries norm but no information. Data budgets at which either optimizer is already past the capacity of the memory, meaning accuracy below $0.99$ at $\sigma = 0$, are dropped from the $\sigma_{50}$ panels: there the two optimizers recover different sets of items and the two $\sigma_{50}$ are not comparable.
\begin{figure}[htbp]
\centering
\vspace{-4pt}
\includegraphics[width=\figwidth]{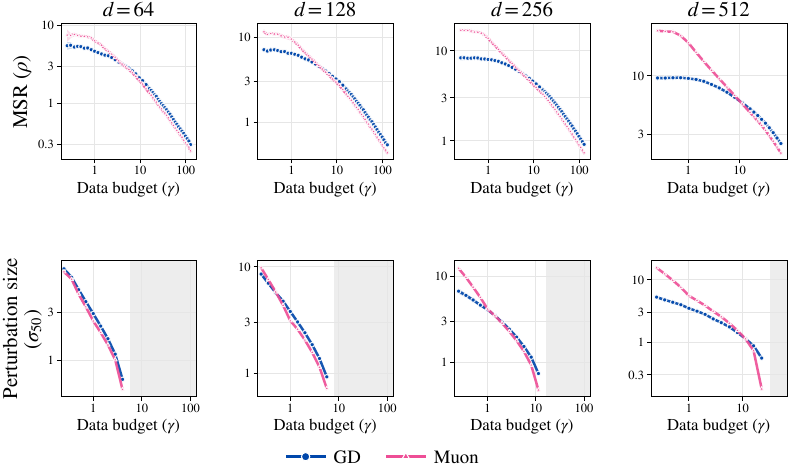}
\vspace{-8pt}
\caption{\textbf{Top: saturated $\rho$ against the data budget}, the two curves behind \cref{fig:budget-R-app}. \textbf{Bottom: perturbed robustness $\sigma_{50}$ against the data budget}, below capacity only. One column per $d$.}
\label{fig:budget-hk-app}
\vspace{-6pt}
\end{figure}

\paragraph{Result}

\begin{figure}[htbp]
\centering
\vspace{-4pt}
\includegraphics[width=\figwidth]{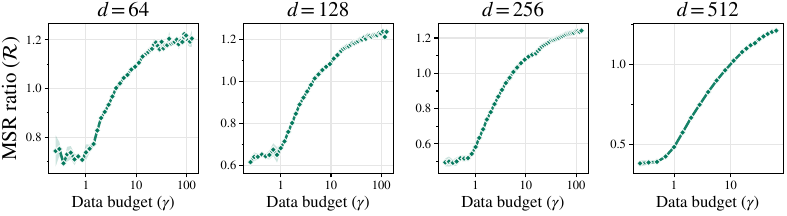}
\vspace{-8pt}
\caption{\textbf{Saturated $\mathcal{R}$ against the data budget}, one panel per $d$.}
\label{fig:budget-R-app}
\vspace{-6pt}
\end{figure}

Training does not undo the one-step picture. \Cref{fig:budget-R-app} shows $\mathcal{R}$ increasing with $\gamma$ at every $d$, over a range from about $0.4$ to about $1.2$, and the crossing $\mathcal{R} = 1$ moving to larger data budgets as $d$ grows: $\gamma^\ast \approx 4.0$, $4.4$, $5.4$ and $9.9$ at $d = 64$, $128$, $256$ and $512$. \Cref{fig:budget-hk-app} shows the two $\rho$ separately: both fall with the data budget, Muon's from roughly twice GD's at $\gamma = \tfrac14$ to below it past $\gamma^\ast$. The bottom row shows that the robustness ordering turns over as well, GD's $\sigma_{50}$ gaining on Muon's at every $d$. The two turnovers need not coincide, and they do not: the $\sigma_{50}$ ordering flips at $\gamma \approx 0.5$, $1.3$ and $11$ at $d = 128$, $256$ and $512$, and at $d = 64$ GD is already the more robust at the smallest data budget we ran.

\subsection{Perturbation robustness}
\label{app:perturb-gammas}

\subsubsection{Setup}

For every (model size, data budget) cell and each optimizer we take the checkpoint attaining the lowest held-out DCLM loss over the full pretraining learning-rate grid. A Gaussian perturbation is added independently to each parameter tensor of that checkpoint, scaled so that $\lVert \Delta W \rVert_F \approx \delta \lVert W \rVert_F$, and we report the increase in held-out DCLM loss over the checkpoint without perturbation, averaged over three perturbation draws. The grid spans three model sizes, 30M to 100M, token--parameter ratios from $5$ to $2{,}560$ TPR, and perturbation magnitudes $\delta$ from $0.005$ to $0.28$. Magnitudes below $\delta \approx 0.02$ are omitted from the figures because the increase in loss is then at the level of the seed-to-seed variation for both optimizers.

\Cref{tab:perturb-crossover} and the robustness counts below use the same protocol on the loss-optimal checkpoints of all five model sizes, 30M to 600M: $36$ (model size, data budget) cells and perturbation magnitudes $\delta$ from $0.002$ to $0.56$.

\subsubsection{Result}

\begin{figure}[H]
\centering
\vspace{-15pt}
\includegraphics[width=\figwidth]{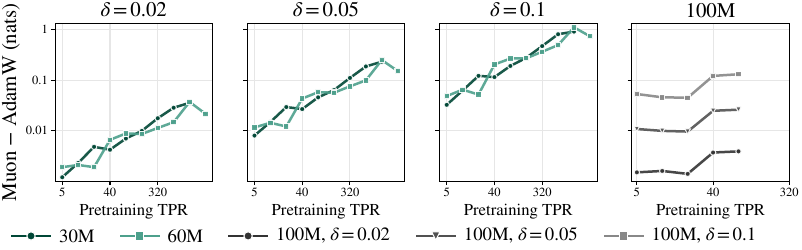}
\caption{\textbf{The perturbation degradation gap is positive at every perturbation magnitude.} Muon minus AdamW increase in held-out DCLM loss. Left three panels: one $\delta$ each; right: 100M across $\delta$.}
\label{fig:app-perturb-gap}
\end{figure}

\Cref{fig:perturb-lines} fixes the perturbation magnitude; \cref{fig:app-perturb-gap} reports the same difference of degradations at every magnitude we sweep. The left three panels fix $\delta$ and vary the model size, the right panel fixes the model size at 100M and varies $\delta$. The difference is positive at every budget, model size and magnitude we tested, and grows with the budget at every $\delta$. Magnitudes are not comparable across panels, since the degradation of both optimizers grows steeply with $\delta$; what the panels share is the shape of the curve.

Over the $36$ cells, the perturbation magnitude $\delta^\ast$ at which the loss has risen by $0.05$, $0.1$ or $0.2$ nats is larger for AdamW in every cell and at every threshold, by a median factor of $1.4$, and Muon's increase in loss exceeds AdamW's in every cell at every $\delta$ from $0.005$ to $0.2$. The ordering does not depend on which checkpoint of a cell is used: giving each optimizer its most perturbation-robust checkpoint within $0.02$ nats of the cell's best base loss still leaves Muon degrading $20$--$400\%$ more.

\noindent\textbf{Loss under perturbation.} \Cref{tab:perturb-crossover} compares the loss under perturbation itself, rather than its increase, and places the result next to the fine-tuning crossover of \cref{fig:crossover}. At every model size, Muon's loss under perturbation is the lower one at small budgets and AdamW's at large ones, apart from the three marked entries. The crossover moves to smaller budgets as $\delta$ grows, so its value is not comparable to the fine-tuning crossover; its trend with model size is. From $\delta = 0.1$ on, the crossover budget does not increase with model size, as for every fine-tuning dataset; at $\delta = 0.05$ it is not monotone.

\begin{table}[htbp]
\centering
\caption{Crossover budgets, in TPR (tokens per parameter). Left: the smallest budget at which AdamW's loss under a perturbation of magnitude $\delta$ is below Muon's (loss-optimal checkpoints, mean over three perturbation draws); $>20$: Muon's is lower up to the largest 600M budget, $20$ TPR; $\leq 5$: AdamW's is lower already at the smallest budget; $^{\dagger}$: Muon's is lower again at $160$ TPR, by $0.003$ nats. Right: the fine-tuning crossover of \cref{fig:crossover}.}
\small
\begin{tabular}{@{}c|cccc|cccc@{}}
\toprule
 & \multicolumn{4}{c|}{\textbf{Perturbation, $\boldsymbol{\delta}$}} & \multicolumn{4}{c}{\textbf{Fine-tuning}} \\
\textbf{Size} & $\mathbf{0.05}$ & $\mathbf{0.1}$ & $\mathbf{0.13}$ & $\mathbf{0.2}$ & \textbf{Alpaca} & \textbf{GSM8K} & \textbf{MusicPile} & \textbf{StackMathQA} \\
\midrule
30M  & $160$           & $40$ & $40$ & $20$      & $160$ & $80$ & $640$ & $320$ \\
60M  & $80$            & $40$ & $40$ & $20$      & $80$  & $80$ & $80$  & $80$  \\
100M & $160$           & $40$ & $20$ & $10$      & $80$  & $40$ & $80$  & $40$  \\
300M & $80^{\dagger}$  & $40$ & $20$ & $10$      & $40$  & $10$ & $40$  & $20$  \\
600M & $>20$           & $20$ & $10$ & $\leq 5$  & $20$  & $5$  & $10$  & $10$  \\
\bottomrule
\end{tabular}
\label{tab:perturb-crossover}
\end{table}

\FloatBarrier

\subsection{Margin-to-spread ratio and robustness}
\label{app:rho-extra}

\subsubsection{Setup}

\noindent\textbf{Linear model.} Each cell is a data budget $\gamma = N/d$ and a model dimension $d$, run with three seeds. We train $W$ from $W_0 = 0$ for $T = 10^4$ full-batch steps on Zipf-distributed item frequencies, with GD at a learning rate of $0.02$ and Muon at $0.005$, and read $\rho$ off the trained $W$. The same $W$ is then perturbed, $W' = W + \sigma \lVert W \rVert_F \xi$ with $\lVert \xi \rVert_F = 1$, over $16$ perturbation directions shared by the two optimizers, and $\sigma_{50}$ is the perturbation level at which half of the items recovered at $\sigma = 0$ are lost, interpolated between grid points on a log scale.

\noindent\textbf{Real language models.} The analogue of a stored fact is a next-token prediction, and the analogue of the competitors of \cref{def:rho} is the rest of the vocabulary. We evaluate the checkpoint on held-out DCLM text ($128$ sequences of $4096$ tokens, never trained on). At a position with gold token $t$ and logit vector $z$, the margins are
\begin{equation}
  \textstyle
  m_j \;=\; z_t - z_j , \qquad j \neq t ,
\end{equation}
over the whole vocabulary, and $h$ and $\kappa^2$ are their mean and variance over those $\lvert V \rvert - 1$ competitors, exactly as in \cref{def:rho}. Two choices make this well defined. First, only the positions at which the gold token is already the argmax, $\min_{j \neq t} m_j > 0$, are used: elsewhere the prediction is wrong before any perturbation is added and the margin is not a robustness margin. Second, the position-level values $h/\kappa$ are summarised by their median over positions rather than their mean, since the distribution over positions is heavy-tailed. The reduction over the vocabulary axis is carried out in double precision and chunked over positions, so the margin vector is never materialised. Both quantities are measured on the same checkpoints: for every (model size, token budget) cell, the checkpoint with the lowest held-out DCLM loss over the pretraining learning-rate grid, which gives $72$ checkpoints over five model sizes. For robustness we take $\delta^\ast$, the relative weight-perturbation level at which the held-out loss rises by $\Delta L$ nats, interpolated on a log--log scale and averaged over three perturbation draws.

\noindent\textbf{Learning-rate ablation.} At two data budgets, each optimizer is run at its learning rate and at that rate scaled by $\tfrac14$ and by $4$, all other settings unchanged, at $d = 128$ and $T = 10^4$ over several seeds.

\begin{table}[htbp]
\begin{tblbox}{tab:lr-ablation}
\begin{tabular*}{\linewidth}{@{\extracolsep{\fill}}clccccc@{}}
$\gamma$ & optimizer & lr & accuracy & final $\rho$ & saturated $\rho$ & $\mathcal{R}$ \\
\midrule
\multirow{6}{*}{$16$}
 & \gdname & $0.005$ $(\times \tfrac{1}{4})$   & $0.2057 \pm 0.0031$ & $2.4610 \pm 0.0142$ & $2.4550 \pm 0.0136$ & \multirow{2}{*}{$1.1354$} \\
 & \muname & $0.00125$ $(\times \tfrac{1}{4})$ & $0.1580 \pm 0.0031$ & $2.1470 \pm 0.0125$ & $2.1624 \pm 0.0133$ & \\
\addlinespace[2pt]
 & \gdname & $0.02$                            & $0.1978 \pm 0.0014$ & $2.4609 \pm 0.0145$ & $2.4610 \pm 0.0144$ & \multirow{2}{*}{$1.1468$} \\
 & \muname & $0.005$                           & $0.1566 \pm 0.0024$ & $2.1418 \pm 0.0121$ & $2.1461 \pm 0.0123$ & \\
\addlinespace[2pt]
 & \gdname & $0.08$ $(\times 4)$               & $0.1969 \pm 0.0017$ & $2.4607 \pm 0.0146$ & $2.4608 \pm 0.0145$ & \multirow{2}{*}{$1.1445$} \\
 & \muname & $0.02$ $(\times 4)$               & $0.1577 \pm 0.0042$ & $2.1492 \pm 0.0111$ & $2.1502 \pm 0.0112$ & \\
\midrule
\multirow{6}{*}{$64$}
 & \gdname & $0.005$ $(\times \tfrac{1}{4})$   & $0.0356 \pm 0.0013$ & $0.9413 \pm 0.0129$ & $0.9384 \pm 0.0128$ & \multirow{2}{*}{$1.1992$} \\
 & \muname & $0.00125$ $(\times \tfrac{1}{4})$ & $0.0272 \pm 0.0013$ & $0.7755 \pm 0.0070$ & $0.7825 \pm 0.0073$ & \\
\addlinespace[2pt]
 & \gdname & $0.02$                            & $0.0344 \pm 0.0012$ & $0.9411 \pm 0.0128$ & $0.9412 \pm 0.0129$ & \multirow{2}{*}{$1.2145$} \\
 & \muname & $0.005$                           & $0.0269 \pm 0.0015$ & $0.7731 \pm 0.0070$ & $0.7750 \pm 0.0070$ & \\
\addlinespace[2pt]
 & \gdname & $0.08$ $(\times 4)$               & $0.0341 \pm 0.0011$ & $0.9410 \pm 0.0128$ & $0.9410 \pm 0.0128$ & \multirow{2}{*}{$1.2138$} \\
 & \muname & $0.02$ $(\times 4)$               & $0.0271 \pm 0.0014$ & $0.7748 \pm 0.0074$ & $0.7753 \pm 0.0073$ & \\
\bottomrule
\end{tabular*}
\par\smallskip {\footnotesize\raggedright \textbf{Table~\thetable.}\ \textit{Learning-rate ablation in the linear model at $d = 128$, $T = 10^4$, mean $\pm$ sd over seeds; ``final'' is $\rho$ at the last step and ``saturated'' its median over $t \geq 10^3$. Each $\mathcal{R}$ pairs the two rows beside it. Accuracy and $\rho$ move by less than $1\%$ over the $16\times$ range, and so does $\mathcal{R}$.}\par}
\end{tblbox}
\end{table}

\subsubsection{Result}

\begin{figure}[htbp]
\centering
\vspace{-4pt}
\includegraphics[width=\figwidth]{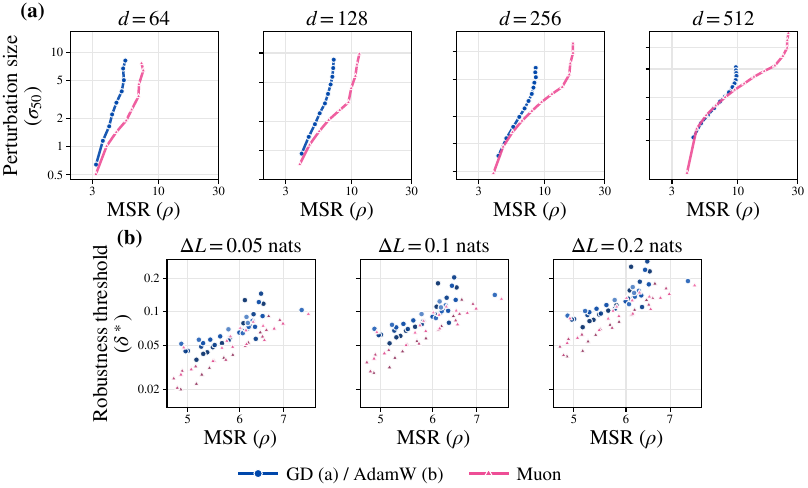}
\vspace{-8pt}
\caption{\textbf{Top: $\sigma_{50}$ against $\rho$ in the linear model}, one panel per model dimension, one point per $\gamma$ cell, below capacity only. \textbf{Bottom: real language models, $\delta^\ast$ against $\rho$ at every threshold}; marker shape is the model size.}
\label{fig:app-hk}
\vspace{-6pt}
\end{figure}

Within each optimizer $\sigma_{50}$ and $\rho$ move together at every model dimension, over more than an order of magnitude in $\sigma_{50}$, and the association is monotone in $\rho$ rather than in $d$ or $\gamma$ separately (\cref{fig:app-hk}). In the language models, the Spearman correlation between $\rho$ and $\delta^\ast$ over the $72$ checkpoints is $+0.83$ at $\Delta L = 0.05$ and $0.1$ nats and $+0.84$ at $0.2$ nats; within each optimizer it is $+0.87$ to $+0.89$ for AdamW and $+0.92$ to $+0.93$ for Muon (\cref{fig:app-hk}). Across the $16\times$ learning-rate range, accuracy moves by less than one percentage point and $\rho$ by less than $1\%$, far below the gap between the two optimizers, and the ratio $\mathcal{R}$ between them by at most $1.3\%$ (\cref{tab:lr-ablation}); the comparison in \cref{sec:budget-experiments} is therefore not an artefact of the learning rates.

\FloatBarrier

\FloatBarrier
\section{Proof of \texorpdfstring{\cref{thm:ratio}}{Theorem}}
\label{app:ratio-proof}

This section proves the quantitative bound underlying the informal \cref{thm:ratio}. For every fixed data budget $\gamma=N/d\geq16$, we establish
\begin{equation}
  \Big|\mathcal{R}-\frac{3\pi}{8}\Big(1-\frac9{10\gamma}\Big)\Big|
  \leq C_1\gamma^{-3/2}+o_p(1)
  \label{eq:ratio}
\end{equation}
as $d\to\infty$, where $C_1$ is an absolute constant. The fixed-percentage increase is established in Lemma~\ref{lem:fixed-ratio}; the other conclusions of \cref{thm:ratio} follow from the negative $1/\gamma$ term in $\mathcal{R}-3\pi/8$, as detailed below.

We first compute the one-step GD and Muon iterates and collect concentration estimates for the Gaussian data. We then evaluate the mean margin $\bar h$ and the margin spread $\bar\kappa$ of \cref{def:rho} separately for the two iterates, and finally take the ratio of the two margin-to-spread ratios. The GD estimates hold for every $\gamma > 1$; the Muon estimates, and hence the theorem, require $\gamma \geq 16$.

\noindent\textbf{Setup.} We work in the setting of \cref{sec:rho-setup} with \emph{uniform query weights}, $q_i = 1/N$ for every fact $i$; this is the setting of \cref{thm:ratio} and is assumed throughout. The keys and values $u_1, \dots, u_N, v_1, \dots, v_N$ are independent $\mathcal{N}(0, I_d)$ vectors, $N = \gamma d$ with the data budget $\gamma > 1$ held fixed, $N - 1 \geq d$, and $d \to \infty$; $O_p$ and $o_p$ are as in \cref{sec:rho-setup}.

\noindent\textbf{Notation.} We collect the data in $U = [u_1 \cdots u_N] \in \mathbb{R}^{d \times N}$ and $V = [v_1 \cdots v_N] \in \mathbb{R}^{d \times N}$, and write $\mathbf{1} = (1, \dots, 1)^\top$, $e_i$ for the $i$-th standard basis vector and $\langle X, Y \rangle = \operatorname{tr}(X^\top Y)$. The operator, Frobenius and nuclear norms $\lVert \cdot \rVert_{\mathrm{op}}$, $\lVert \cdot \rVert_F$ and $\lVert \cdot \rVert_*$ satisfy
\begin{equation}
\begin{aligned}
  \lVert X \rVert_{\mathrm{op}} &\leq \lVert X \rVert_F \leq \lVert X \rVert_*
  \leq \operatorname{rank}(X) \lVert X \rVert_{\mathrm{op}} ,
  \qquad
  \lVert X \rVert_F \leq \sqrt{\operatorname{rank}(X)} \, \lVert X \rVert_{\mathrm{op}} , \\
  \lvert \langle X, Y \rangle \rvert &\leq \lVert X \rVert_* \lVert Y \rVert_{\mathrm{op}} .
\end{aligned}
  \label{eq:norms}
\end{equation}
For $X \in \mathbb{R}^{d \times d}$ with compact SVD $X = P \Sigma Q^\top$ we write $\operatorname{Orth}(X) = P Q^\top$ for its polar factor, so that $\operatorname{Orth}(cX) = \operatorname{Orth}(X)$ for $c > 0$ and $\lVert \operatorname{Orth}(X) \rVert_{\mathrm{op}} \leq 1$.

\noindent\textbf{Scores and observables.} By \cref{def:rho}, for any $W$,
\begin{equation}
\begin{aligned}
  \bar h &= \operatorname*{mean}_i u_i^\top W v_i - \operatorname*{mean}_i \operatorname*{mean}_{j \neq i} u_j^\top W v_i , \\
  \bar\kappa^2 &= \frac{1}{N(N-1)} \sum_{i=1}^N \sum_{j \neq i} \big( u_j^\top W v_i \big)^2
  - \operatorname*{mean}_i \Big( \operatorname*{mean}_{j \neq i} u_j^\top W v_i \Big)^2 ,
\end{aligned}
  \label{eq:general-hk}
\end{equation}
where the scores are written as $u_j^\top W v_i$ rather than $v_j^\top W u_i$ as in \cref{sec:rho-setup}. The two conventions are exchanged by swapping the roles of $U$ and $V$: reading \cref{sec:rho-setup} with $(U, V)$ replaced by $(V, U)$ turns every score $v_j^\top W u_i$ into $u_j^\top W v_i$ and its one-step iterate into \cref{eq:one-step}, leaving the loss and every $h_i$ and $\kappa_i$ unchanged. Since $U$ and $V$ are i.i.d., the law of $\rho$ is the same under either convention.

\subsection{The two one-step iterates}

\begin{lmmbox}{One step}{lem:one-step}
Up to a positive multiple, one step of GD and one step of Muon from $W_0 = 0$ give
\begin{equation}
  W^{\mathrm{GD}} = U V^\top - \frac{1}{N} (U \mathbf{1})(V \mathbf{1})^\top
  = U \Big( I_N - \frac{\mathbf{1}\mathbf{1}^\top}{N} \Big) V^\top ,
  \qquad
  W^{\mathrm{Muon}} = \operatorname{Orth}\big( W^{\mathrm{GD}} \big) .
  \label{eq:one-step}
\end{equation}
The positive multiple is immaterial for $\rho$.
\end{lmmbox}

\begin{proof}
\emph{Gradient.} With $\partial_W (u_j^\top W v_i) = u_j v_i^\top$, the gradient of the loss of \cref{sec:rho-setup} with uniform weights $q_i = 1/N$ is
\begin{equation}
  - \nabla_W \mathcal{L}(W) = \frac{1}{N} \sum_{i=1}^N \Big( u_i - \sum_{j=1}^N \hat p_W(j \mid i) u_j \Big) v_i^\top .
\end{equation}
At $W = 0$ all scores vanish, so $\hat p_0(j \mid i) = 1/N$; moreover $\sum_j u_j = U \mathbf{1}$, $\sum_i u_i v_i^\top = UV^\top$ and $\sum_i v_i^\top = (V\mathbf{1})^\top$, whence
\begin{equation}
  - \nabla_W \mathcal{L}(0) = \frac{1}{N} \Big( UV^\top - \frac{1}{N} (U\mathbf{1})(V\mathbf{1})^\top \Big)
  = \frac{1}{N} U \Big( I_N - \frac{\mathbf{1}\mathbf{1}^\top}{N} \Big) V^\top .
\end{equation}
\emph{One step.} With step size $\eta > 0$, GD gives $-\eta \nabla_W \mathcal{L}(0) = \tfrac{\eta}{N} W^{\mathrm{GD}}$ and Muon gives $\eta \operatorname{Orth}(-\nabla_W \mathcal{L}(0)) = \eta \operatorname{Orth}(W^{\mathrm{GD}})$, using $\operatorname{Orth}(cX) = \operatorname{Orth}(X)$ for $c > 0$. \emph{Normalization.} Replacing $W$ by $cW$ with $c > 0$ multiplies every score $u_j^\top W v_i$ by $c$, hence $h_i \mapsto ch_i$ and $\kappa_i \mapsto c\kappa_i$, so $\bar h \mapsto c \bar h$, $\bar\kappa \mapsto c\bar\kappa$ and $\rho$ is unchanged.
\end{proof}

\begin{lemma}[Full rank and orthogonality]
\label{lem:fullrank}
Almost surely $\operatorname{rank} W^{\mathrm{GD}} = d$. Consequently $W^{\mathrm{Muon}}$ is an orthogonal matrix: $\lVert W^{\mathrm{Muon}} x \rVert = \lVert x \rVert$, $\lVert W^{\mathrm{Muon}} \rVert_{\mathrm{op}} = 1$ and $\lVert W^{\mathrm{Muon}} \rVert_F^2 = d$.
\end{lemma}

\begin{proof}
$\det W^{\mathrm{GD}}$ is a polynomial in the entries of $(U, V)$. Taking $U = [\, I_d \ \ 0 \,]$ and $V = [\, I_d \ \ -\sum_{k \leq d} e_k \ \ 0 \,]$, which is possible because $N \geq d + 1$, gives $V\mathbf{1} = 0$ and $W^{\mathrm{GD}} = UV^\top = I_d$, so the polynomial is not identically zero; since $(U, V)$ has a density, $\Pr[\det W^{\mathrm{GD}} = 0] = 0$. At full rank the factors $P, Q$ of the compact SVD are $d \times d$ orthogonal matrices, so $W^{\mathrm{Muon}} = PQ^\top$ is orthogonal and $\lVert W^{\mathrm{Muon}} \rVert_F^2 = \operatorname{rank} W^{\mathrm{GD}} = d$.
\end{proof}

\subsection{Concentration estimates for the data}

\begin{lemma}[Concentration for $U$, $V$]
\label{lem:conc-uv}
As $d \to \infty$ along the sequence above:
\begin{enumerate}[label=(\roman*), leftmargin=*, itemsep=1pt, topsep=2pt]
  \item $\operatorname{mean}_k \lVert u_k \rVert^2 = d \big( 1 + O_p((Nd)^{-1/2}) \big)$, and likewise for $V$.
  \item $\operatorname{mean}_k \big( \lVert u_k \rVert \lVert v_k \rVert - d \big)^2 = O_p(d)$.
  \item $\lVert U\mathbf{1} \rVert^2 / (Nd) \to 1$ in probability; in particular $\lVert U\mathbf{1} \rVert^2 = O_p(Nd)$. Likewise for $V$.
  \item $\lVert U \rVert_{\mathrm{op}} / \sqrt{N} \to 1 + \gamma^{-1/2}$ almost surely, hence $\lVert U \rVert_{\mathrm{op}}^2 \leq 4N(1 + o_p(1))$; and for every $m \geq 1$ there is $C_m$ with $\mathbb{E} \lVert U \rVert_{\mathrm{op}}^{2m} \leq C_m N^m$. Likewise for $V$.
  \item $\lVert UU^\top - N I_d \rVert_{\mathrm{op}} = O_p(\sqrt{Nd})$. If $\gamma \geq 16$, then with probability tending to one both $S = \tfrac{1}{N} VV^\top$ and $S = \tfrac{1}{N}(VV^\top - \tfrac{1}{N}(V\mathbf{1})(V\mathbf{1})^\top)$ satisfy $\lVert S - I_d \rVert_{\mathrm{op}} \leq 3\gamma^{-1/2}$ and $\lambda_{\min}(S) \geq \tfrac{1}{8}$.
  \item $\lVert VV^\top \rVert_F^2 = O_p(N^2 d)$, and $\mathbb{E} \lVert VV^\top \rVert_F^{2m} \leq C_{2m} (N^2 d)^m$. Likewise for $UU^\top$.
  \item $\lVert UV^\top \rVert_F^2 = Nd^2 (1 + O_p(d^{-1}))$ and $\lVert UV^\top \rVert_F^2 - d \sum_k \lVert v_k \rVert^2 = O_p(\sqrt{N} d^{3/2} + Nd)$.
  \item $UV^\top \overset{d}{=} (UU^\top)^{1/2} \Xi$ with $\Xi \in \mathbb{R}^{d \times d}$ having i.i.d.\ $\mathcal{N}(0,1)$ entries independent of $U$; with probability tending to one, $\lVert UV^\top \rVert_{\mathrm{op}} \leq 5\sqrt{Nd}$.
\end{enumerate}
\end{lemma}

\begin{proof}
(i) $\sum_k \lVert u_k \rVert^2 \sim \chi^2_{Nd}$ has mean $Nd$ and variance $2Nd$, so Chebyshev gives $\sum_k \lVert u_k \rVert^2 = Nd + O_p(\sqrt{Nd})$; divide by $N$.

(ii) For $x \geq 0$ one has $\sqrt{x} \geq 1 + \tfrac{x-1}{2} - \tfrac{(x-1)^2}{2}$: the difference $g$ of the two sides satisfies $g(0) = g(1) = g'(1) = 0$ and $g''(x) = 1 - \tfrac{1}{4} x^{-3/2}$ changes sign once, at $x_0 = 4^{-2/3}$, so $g$ is concave on $[0, x_0]$ and convex on $[x_0, \infty)$, and the two boundary values force $g \geq 0$. With $u \sim \mathcal{N}(0, I_d)$, $\mathbb{E} \lVert u \rVert^2 / d = 1$ and $\operatorname{var}( \lVert u \rVert^2 / d) = 2/d$, so $\mathbb{E} \lVert u \rVert \geq \sqrt{d}(1 - \tfrac{1}{d})$ and $(\mathbb{E} \lVert u \rVert)^2 \geq d - 2$. Hence for independent $u, v$,
\begin{equation}
  \mathbb{E} \big( \lVert u \rVert \lVert v \rVert - d \big)^2
  = \mathbb{E} \lVert u \rVert^2 \, \mathbb{E} \lVert v \rVert^2 - 2d \, \mathbb{E} \lVert u \rVert \, \mathbb{E} \lVert v \rVert + d^2
  \leq 2d^2 - 2d(d-2) = 4d ,
\end{equation}
and Markov's inequality gives the claim.

(iii) $U\mathbf{1} = \sum_k u_k \sim \mathcal{N}(0, N I_d)$, so $\lVert U\mathbf{1} \rVert^2 / N \sim \chi^2_d$ has mean $d$ and variance $2d$; Chebyshev gives $\lVert U\mathbf{1} \rVert^2/(Nd) = 1 + O_p(d^{-1/2})$.

(iv) $\lVert U \rVert_{\mathrm{op}}^2 = \lambda_{\max}(UU^\top)$, and by the Bai--Yin law $\lambda_{\max}(UU^\top/N) \to (1 + \gamma^{-1/2})^2 \leq 4$ almost surely, since $\gamma \geq 1$. For the moments, $U \mapsto \lVert U \rVert_{\mathrm{op}}$ is $1$-Lipschitz for $\lVert \cdot \rVert_F$ and $\mathbb{E} \lVert U \rVert_{\mathrm{op}} \leq \sqrt{N} + \sqrt{d} \leq 2\sqrt{N}$ by Gordon's theorem, so Gaussian concentration gives $\Pr[\lVert U \rVert_{\mathrm{op}} \geq 2\sqrt{N} + t] \leq e^{-t^2/2}$; integrating the tail bounds all moments.

(v) By Bai--Yin, $\lambda_{\max}(UU^\top/N) \to (1 + \gamma^{-1/2})^2$ and $\lambda_{\min}(UU^\top/N) \to (1 - \gamma^{-1/2})^2$ almost surely, so $\lVert UU^\top/N - I_d \rVert_{\mathrm{op}} \to \max\{(1+\gamma^{-1/2})^2 - 1, 1 - (1-\gamma^{-1/2})^2\} = 2\gamma^{-1/2} + \gamma^{-1}$, which is at most $3\gamma^{-1/2} < 4\gamma^{-1/2}$; multiplying by $N$ gives the first claim. For $\gamma \geq 16$ we have $\gamma^{-1/2} \leq \tfrac14$, so in probability $\lVert VV^\top/N - I_d \rVert_{\mathrm{op}} \to 2\gamma^{-1/2} + \gamma^{-1} \leq \tfrac94 \gamma^{-1/2} < 3\gamma^{-1/2}$ and $\lambda_{\min}(VV^\top/N) \to (1 - \gamma^{-1/2})^2 \geq \tfrac{9}{16} > \tfrac18$, and the inequalities being strict they hold with probability tending to one. For the centred version, write $\tfrac1N \big( VV^\top - \tfrac1N (V\mathbf{1})(V\mathbf{1})^\top \big) = \tfrac{VV^\top}{N} - \tfrac{1}{N^2} (V\mathbf{1})(V\mathbf{1})^\top$, where $\tfrac{1}{N^2} (V\mathbf{1})(V\mathbf{1})^\top \succeq 0$ has operator norm $\lVert V\mathbf{1} \rVert^2/N^2 = \tfrac{\lVert V \mathbf{1}\rVert^2}{Nd} \cdot \tfrac{d}{N} \to \gamma^{-1}$ in probability by (iii). By Weyl's inequality,
\begin{equation}
\begin{aligned}
  \lambda_{\min}\Big( \frac1N \big( VV^\top - \frac1N (V\mathbf{1})(V\mathbf{1})^\top \big) \Big)
  &\geq \lambda_{\min}\Big( \frac{VV^\top}{N} \Big) - \frac{\lVert V\mathbf{1} \rVert^2}{N^2} , \\
  \Big\lVert \frac1N \big( VV^\top - \frac1N (V\mathbf{1})(V\mathbf{1})^\top \big) - I_d \Big\rVert_{\mathrm{op}}
  &\leq \Big\lVert \frac{VV^\top}{N} - I_d \Big\rVert_{\mathrm{op}} + \frac{\lVert V\mathbf{1} \rVert^2}{N^2} ,
\end{aligned}
\end{equation}
and in probability the two right-hand sides converge to
\begin{equation}
  (1 - \gamma^{-1/2})^2 - \gamma^{-1} \geq \frac{9}{16} - \frac{1}{16} = \frac12 > \frac18 ,
  \qquad
  2\gamma^{-1/2} + 2\gamma^{-1} \leq \frac52 \gamma^{-1/2} < 3\gamma^{-1/2} ,
\end{equation}
using $\gamma^{-1/2} \leq \tfrac14$; the inequalities being strict, they hold with probability tending to one.

(vi) $VV^\top$ has rank at most $d$, so $\lVert VV^\top \rVert_F \leq \sqrt{d} \lVert VV^\top \rVert_{\mathrm{op}} = \sqrt{d} \lVert V \rVert_{\mathrm{op}}^2$ by \cref{eq:norms}; now use (iv).

(vii) Expanding, $\lVert UV^\top \rVert_F^2 = \sum_{k,l} (u_k^\top u_l)(v_k^\top v_l) = \sum_k \lVert u_k \rVert^2 \lVert v_k \rVert^2 + \sum_{k \neq l} (u_k^\top u_l)(v_k^\top v_l)$.

\emph{Mean.} $\mathbb{E} \lVert u_k \rVert^2 \lVert v_k \rVert^2 = d^2$ and, for $k \neq l$, $\mathbb{E} (u_k^\top u_l)(v_k^\top v_l) = 0$, so $\mathbb{E} \lVert UV^\top \rVert_F^2 = Nd^2$.

\emph{Variance.} The variables $\lVert u_k \rVert^2 \lVert v_k \rVert^2$, $k \in [N]$, are independent with $\operatorname{var}(\lVert u_k \rVert^2 \lVert v_k \rVert^2) = (d^2 + 2d)^2 - d^4 \leq 8d^3$. Next, $\sum_{k \neq l} (u_k^\top u_l)(v_k^\top v_l) = 2 \sum_{k < l} (u_k^\top u_l)(v_k^\top v_l)$; for unordered pairs $\{k,l\} \neq \{k',l'\}$ one of the indices $k, l, k', l'$ occurs exactly once, and integrating over the $u$ with that index gives $\mathbb{E}\big[ (u_k^\top u_l)(v_k^\top v_l)(u_{k'}^\top u_{l'})(v_{k'}^\top v_{l'}) \big] = 0$; since $\mathbb{E}(u_k^\top u_l)^2 (v_k^\top v_l)^2 = d^2$, this gives $\operatorname{var} \sum_{k \neq l} (u_k^\top u_l)(v_k^\top v_l) = 4 \cdot \tfrac{N(N-1)}{2} \cdot d^2 \leq 2N^2 d^2$. For $k' \in [N]$ and $k \neq l$, one of $k, l$ differs from $k'$, and integrating over the $u$ with that index gives $\mathbb{E}\big[ (\lVert u_{k'} \rVert^2 \lVert v_{k'} \rVert^2 - d^2)(u_k^\top u_l)(v_k^\top v_l) \big] = 0$, so the two sums are uncorrelated. Therefore
\begin{equation}
  \operatorname{var} \lVert UV^\top \rVert_F^2 \leq 8Nd^3 + 2N^2 d^2 ,
  \qquad
  \frac{\big( \operatorname{var} \lVert UV^\top \rVert_F^2 \big)^{1/2}}{\mathbb{E} \lVert UV^\top \rVert_F^2}
  \leq \sqrt{\frac{8}{Nd} + \frac{2}{d^2}} \leq \frac{4}{d} ,
\end{equation}
and Chebyshev gives the first claim.

\emph{Centred version.} $\lVert UV^\top \rVert_F^2 - d \sum_k \lVert v_k \rVert^2 = \sum_k (\lVert u_k \rVert^2 - d) \lVert v_k \rVert^2 + \sum_{k \neq l} (u_k^\top u_l)(v_k^\top v_l)$ has mean zero. The variables $(\lVert u_k \rVert^2 - d) \lVert v_k \rVert^2$ are independent with variance $\mathbb{E}(\lVert u \rVert^2 - d)^2 \, \mathbb{E} \lVert v \rVert^4 = 2d(d^2 + 2d) \leq 6d^3$; the second sum has variance at most $2N^2 d^2$ as above; and for $k' \in [N]$ and $k \neq l$, integrating over the $u$ whose index among $k, l$ differs from $k'$ gives $\mathbb{E}\big[ (\lVert u_{k'} \rVert^2 - d) \lVert v_{k'} \rVert^2 (u_k^\top u_l)(v_k^\top v_l) \big] = 0$. Hence the variance is at most $6Nd^3 + 2N^2 d^2$, its square root is at most $\sqrt6 \sqrt N d^{3/2} + \sqrt2 Nd$, and Chebyshev gives the second claim.

(viii) Conditionally on $U$, the columns of $UV^\top$ are i.i.d.\ $\mathcal{N}(0, UU^\top)$, i.e.\ of the form $(UU^\top)^{1/2} g$ with $g \sim \mathcal{N}(0, I_d)$; stacking the $d$ vectors $g$ as the columns of $\Xi$ gives the stated equality in distribution, and the conditional law of $\Xi$ does not depend on $U$, so $\Xi$ is independent of $U$. For the norm, $\lVert (UU^\top)^{1/2} \Xi \rVert_{\mathrm{op}} \leq \lVert U \rVert_{\mathrm{op}} \lVert \Xi \rVert_{\mathrm{op}}$, and almost surely $\lVert U \rVert_{\mathrm{op}}/\sqrt{N} \to 1 + \gamma^{-1/2} \leq 2$ by (iv) while $\lVert \Xi \rVert_{\mathrm{op}}/\sqrt{d} \to 2$ by Bai--Yin with aspect ratio $1$.
\end{proof}

\begin{lmmbox}{Norms of $W^{\mathrm{GD}}$}{lem:norms-wgd}
\begin{enumerate}[label=(\roman*), leftmargin=*, itemsep=1pt, topsep=2pt]
  \item $\lVert W^{\mathrm{GD}} \rVert_F^2 = (N-1) d^2 (1 + O_p(d^{-1}))$, and $\lVert W^{\mathrm{GD}} \rVert_F^2 - d \operatorname{tr}\big( VV^\top - \tfrac{1}{N}(V\mathbf{1})(V\mathbf{1})^\top \big) = O_p(\sqrt{N} d^{3/2} + Nd)$.
  \item $\lVert W^{\mathrm{GD}} \rVert_{\mathrm{op}} \leq \lVert U \rVert_{\mathrm{op}} \lVert V \rVert_{\mathrm{op}}$, and with probability tending to one $\lVert W^{\mathrm{GD}} \rVert_{\mathrm{op}} \leq 7\sqrt{Nd}$.
  \item With probability tending to one, $\tfrac{1}{8} d\sqrt{Nd} \leq \lVert W^{\mathrm{GD}} \rVert_* \leq 7 d \sqrt{Nd}$.
  \item In probability, $\lVert W^{\mathrm{GD}} \rVert_* / (Nd) \to \Phi(\gamma)/\sqrt{\gamma}$, where
  \begin{equation}
    \Phi(\gamma) := \frac{1}{\pi} \int_0^1 \bigg[ \sqrt{1 - \frac{p}{\gamma}} \, \frac{1+p}{\sqrt{p}}
    + \frac{\sqrt{p}\,(1-p)}{\gamma \sqrt{1 - p/\gamma}} \bigg] \mathrm{d}p ,
    \label{eq:phi-def}
  \end{equation}
  and for $\gamma \geq 16$
  \begin{equation}
    \Phi(\gamma) = \frac{8}{3\pi} \Big( 1 - \frac{1}{10\gamma} + r(\gamma) \Big) ,
    \qquad \lvert r(\gamma) \rvert \leq \gamma^{-2} .
    \label{eq:phi-exp}
  \end{equation}
\end{enumerate}
\end{lmmbox}
Part (iv) is where the constant $3\pi/8$ enters: Muon replaces $W^{\mathrm{GD}}$ by its polar factor, which turns the diagonal score into $\lVert W^{\mathrm{GD}} \rVert_*$, and $8/(3\pi)$ is the limiting Mar\v{c}enko--Pastur integral at $\gamma = \infty$.

\begin{proof}
(i) By \cref{eq:one-step} and $\lVert X - Y \rVert_F^2 = \lVert X \rVert_F^2 - 2 \langle X, Y \rangle + \lVert Y \rVert_F^2$,
\begin{equation}
  \lVert W^{\mathrm{GD}} \rVert_F^2 = \lVert UV^\top \rVert_F^2
  - \frac{2}{N} (U\mathbf{1})^\top UV^\top (V\mathbf{1})
  + \frac{1}{N^2} \lVert U\mathbf{1} \rVert^2 \lVert V\mathbf{1} \rVert^2 ,
\end{equation}
where the cross term is $O_p(N^2 d)/N = O_p(Nd)$ and the last term is $O_p(d^2)$ by \cref{lem:conc-uv}(iii),(iv),(viii). Substituting $\lVert UV^\top \rVert_F^2 = Nd^2(1 + O_p(d^{-1}))$ from \cref{lem:conc-uv}(vii) and using $d^2 \leq Nd$ gives the first claim. For the second, $\operatorname{tr}(VV^\top) = \sum_k \lVert v_k \rVert^2$ and $\operatorname{tr}((V\mathbf{1})(V\mathbf{1})^\top) = \lVert V\mathbf{1} \rVert^2$, so the same expansion gives
\begin{equation}
\begin{aligned}
  \lVert W^{\mathrm{GD}} \rVert_F^2 - d \operatorname{tr}\Big( VV^\top - \frac1N (V\mathbf{1})(V\mathbf{1})^\top \Big)
  = {}&\Big( \lVert UV^\top \rVert_F^2 - d \sum_k \lVert v_k \rVert^2 \Big) + \frac{d \lVert V\mathbf{1} \rVert^2}{N} \\
  &\quad {}- \frac{2}{N} (U\mathbf{1})^\top UV^\top (V\mathbf{1}) + \frac{\lVert U\mathbf{1} \rVert^2 \lVert V\mathbf{1} \rVert^2}{N^2} ,
\end{aligned}
\end{equation}
whose four terms are $O_p(\sqrt N d^{3/2} + Nd)$ by the second half of \cref{lem:conc-uv}(vii), $O_p(d^2)$, $O_p(Nd)$ and $O_p(d^2)$ respectively; since $d^2 \leq Nd$, the claim follows.

(ii) $\lVert I_N - \tfrac{\mathbf{1}\mathbf{1}^\top}{N} \rVert_{\mathrm{op}} = 1$ gives the first inequality. For the second, the triangle inequality and the operator norm of a rank-one matrix give $\lVert W^{\mathrm{GD}} \rVert_{\mathrm{op}} \leq \lVert UV^\top \rVert_{\mathrm{op}} + \lVert U\mathbf{1} \rVert \lVert V\mathbf{1} \rVert / N \leq 5\sqrt{Nd} + \tfrac32 d \leq 7 \sqrt{Nd}$ with probability tending to one, by \cref{lem:conc-uv}(iii),(viii) and $d \leq \sqrt{Nd}$.

(iii) For any $X$, $\lVert X \rVert_F^2 = \sum_a s_a^2 \leq s_1 \sum_a s_a = \lVert X \rVert_{\mathrm{op}} \lVert X \rVert_*$, and $\lVert X \rVert_* \leq \operatorname{rank}(X) \lVert X \rVert_{\mathrm{op}}$. Combining with (i) and (ii),
\begin{equation}
\begin{aligned}
  \lVert W^{\mathrm{GD}} \rVert_* &\geq \frac{\lVert W^{\mathrm{GD}} \rVert_F^2}{\lVert W^{\mathrm{GD}} \rVert_{\mathrm{op}}}
  \geq \frac{(N-1)d^2 (1 - o_p(1))}{7\sqrt{Nd}} \geq \frac18 d \sqrt{Nd} , \\
  \lVert W^{\mathrm{GD}} \rVert_* &\leq d \lVert W^{\mathrm{GD}} \rVert_{\mathrm{op}} \leq 7 d\sqrt{Nd} .
\end{aligned}
\end{equation}

(iv) \emph{Removing the centring.} By the triangle inequality for the nuclear norm and the nuclear norm of a rank-one matrix, $\lvert \lVert UV^\top \rVert_* - \lVert W^{\mathrm{GD}} \rVert_* \rvert \leq \lVert U\mathbf{1} \rVert \lVert V\mathbf{1} \rVert / N = O_p(d)$, which is $o(Nd)$; so it suffices to prove the limit for $\lVert UV^\top \rVert_*$.

\emph{Limit.} By \cref{lem:conc-uv}(viii), $\{ s_a(UV^\top)^2/(Nd) \}_{a \leq d}$ has the same law as the eigenvalues of $\tfrac1d (\tfrac{UU^\top}{N})^{1/2} \Xi \Xi^\top (\tfrac{UU^\top}{N})^{1/2}$ with $\Xi$ independent of $U$: conditionally on $U$ this is a sample covariance matrix with population covariance $\tfrac1N UU^\top$ and aspect ratio $1$, and the spectral distribution of $\tfrac1N UU^\top$ converges almost surely to the Mar\v{c}enko--Pastur law $H$ with ratio $1/\gamma$ and mean $1$. By the Mar\v{c}enko--Pastur--Silverstein theorem the empirical distribution of $\{ s_a(UV^\top)^2/(Nd) \}$ converges weakly in probability to a law $\nu$ whose Stieltjes transform $m$ satisfies
\begin{equation}
  z = -\frac{1}{m} + \int \frac{t}{1 + tm} \, \mathrm{d}H(t) .
  \label{eq:silverstein}
\end{equation}
By \cref{lem:conc-uv}(viii) all $s_a^2/(Nd) \leq 25$ with probability tending to one, so both the empirical law and $\nu$ are supported in $[0, 25]$; taking the limit for the bounded continuous function $\lambda \mapsto \min(\sqrt{\lambda}, 5)$ gives $\tfrac1d \sum_a s_a/\sqrt{Nd} \to \int \sqrt{\lambda} \, \mathrm{d}\nu$ in probability, that is
\begin{equation}
  \frac{\lVert UV^\top \rVert_*}{Nd} = \sqrt{\frac{d}{N}} \cdot \frac{1}{d} \sum_a \frac{s_a}{\sqrt{Nd}}
  \longrightarrow \frac{1}{\sqrt{\gamma}} \int \sqrt{\lambda} \, \mathrm{d}\nu(\lambda) .
\end{equation}

\emph{Closed form.} Let $\psi(x) := \int \tfrac{\lambda x}{1 - \lambda x} \mathrm{d}\nu$. From $\tfrac{\lambda x}{1-\lambda x} = -1 + \tfrac{1/x}{1/x - \lambda}$ we get $\psi(x) = -1 - \tfrac1x m(1/x)$, i.e.\ $m(1/x) = -x(1 + \psi(x))$. Taking $z = 1/x$ in \cref{eq:silverstein} and multiplying by $x(1 + \psi(x))$,
\begin{equation}
  \psi(x) = \int \frac{x(1 + \psi(x)) t}{1 - x(1 + \psi(x)) t} \, \mathrm{d}H(t) .
  \label{eq:psi}
\end{equation}
On the other hand the Stieltjes transform of the Mar\v{c}enko--Pastur law with ratio $1/\gamma$ and mean $1$ satisfies $\tfrac1\gamma z m_H(z)^2 - (1 - \tfrac1\gamma - z) m_H(z) + 1 = 0$, and $\tfrac{ut}{1-ut} = -1 + \tfrac{1/u}{1/u - t}$ gives $m_H(1/u) = -u \big( 1 + \int \tfrac{ut}{1-ut} \mathrm{d}H(t) \big)$; substituting $z = 1/u$ into the quadratic yields, for all $u$,
\begin{equation}
  \frac{u}{\gamma} \Big( 1 + \int \frac{ut}{1-ut} \mathrm{d}H(t) \Big)^2
  + \Big( 1 - \frac1\gamma \Big) u \Big( 1 + \int \frac{ut}{1-ut} \mathrm{d}H(t) \Big)
  - \int \frac{ut}{1-ut} \mathrm{d}H(t) = 0 .
\end{equation}
Take $u = x(1 + \psi(x))$. By \cref{eq:psi} the integral then equals $\psi(x)$, so
\begin{equation}
  \frac1\gamma x (1+\psi)^3 + \Big( 1 - \frac1\gamma \Big) x (1+\psi)^2 - \psi = 0 ,
  \qquad\text{i.e.}\qquad
  x = \frac{\psi(x)}{(1 + \psi(x))^2 \, (1 + \psi(x)/\gamma)} .
\end{equation}
Using $\pi \sqrt{\lambda} = \int_0^\infty \tfrac{\lambda}{\lambda + w} \tfrac{\mathrm{d}w}{\sqrt{w}}$ and Tonelli's theorem, and then substituting $p = -\psi(-1/w) \in (0,1)$,
\begin{equation}
\begin{aligned}
  \int \sqrt{\lambda} \, \mathrm{d}\nu &= \frac1\pi \int_0^\infty \frac{-\psi(-1/w)}{\sqrt w} \, \mathrm{d}w , \\
  w &= \frac{(1-p)^2 (1 - p/\gamma)}{p} ,
  \qquad - \frac{\mathrm{d}w}{\mathrm{d}p} = w \Big[ \frac{2}{1-p} + \frac{1/\gamma}{1 - p/\gamma} + \frac1p \Big] , \\
  \int \sqrt{\lambda} \, \mathrm{d}\nu &= \frac1\pi \int_0^1 \sqrt{1 - \frac{p}{\gamma}} \Big[ 2\sqrt p
  + \frac{\sqrt p (1-p)/\gamma}{1 - p/\gamma} + \frac{1-p}{\sqrt p} \Big] \mathrm{d}p ;
\end{aligned}
\end{equation}
then $2\sqrt p + \tfrac{1-p}{\sqrt p} = \tfrac{1+p}{\sqrt p}$ and $\sqrt{1 - p/\gamma} \cdot \tfrac{\sqrt p (1-p)/\gamma}{1 - p/\gamma} = \tfrac{\sqrt p (1-p)}{\gamma \sqrt{1 - p/\gamma}}$ give \cref{eq:phi-def}.

\emph{Expansion.} For $0 \leq s \leq \tfrac{1}{16}$ one has
\begin{equation}
  \Big\lvert \sqrt{1-s} - 1 + \frac{s}{2} \Big\rvert \leq \frac{s^2}{5} ,
  \qquad
  0 \leq (1-s)^{-1/2} - 1 \leq \frac{2s}{3} .
\end{equation}
For the first, $g(s) := 1 - \tfrac s2 - \sqrt{1-s}$ satisfies $g(0) = 0$, $g'(s) = \tfrac12((1-s)^{-1/2} - 1) \geq 0$ and $g''(s) = \tfrac14 (1-s)^{-3/2} \leq \tfrac14 (1-\tfrac1{16})^{-3/2} < \tfrac25$ on $[0,\tfrac1{16}]$, so $g(s) \leq \tfrac{s^2}{2} \max g'' < \tfrac{s^2}{5}$. For the second, $f(s) := (1-s)^{-1/2} - 1 - \tfrac{2s}{3}$ satisfies $f(0) = 0$ and $f'(s) = \tfrac12 (1-s)^{-3/2} - \tfrac23 < 0$ for $s < 1 - (\tfrac34)^{2/3} \approx 0.175$, hence on $[0, \tfrac1{16}]$, so $f \leq 0$ there. Take $s = p/\gamma$ with $\gamma \geq 16$. From
\begin{equation}
  \frac{1}{\pi} \int_0^1 \frac{1+p}{\sqrt{p}} \, \mathrm{d}p = \frac{8}{3\pi} ,
  \qquad
  - \frac{1}{2\gamma\pi} \int_0^1 \sqrt{p} (1+p) \, \mathrm{d}p + \frac{1}{\gamma\pi} \int_0^1 \sqrt{p}(1-p) \, \mathrm{d}p
  = \frac{8}{3\pi} \Big( - \frac{1}{10\gamma} \Big) ,
\end{equation}
and \cref{eq:phi-def}, the remainder is bounded by
\begin{equation}
\begin{aligned}
  \Big\lvert \Phi(\gamma) - \frac{8}{3\pi} \Big( 1 - \frac{1}{10\gamma} \Big) \Big\rvert
  &\leq \frac{1}{\pi \gamma^2} \Big[ \frac15 \int_0^1 p^{3/2}(1+p) \, \mathrm{d}p + \frac23 \int_0^1 p^{3/2}(1-p) \, \mathrm{d}p \Big] \\
  &= \frac{1}{\pi\gamma^2} \Big[ \frac{24}{175} + \frac{8}{105} \Big] < \frac{8}{3\pi} \gamma^{-2} ,
\end{aligned}
\end{equation}
which is \cref{eq:phi-exp}.
\end{proof}

\subsection{The GD step}

By \cref{eq:general-hk} with $W = W^{\mathrm{GD}}$,
\begin{equation}
  \bar h^{\mathrm{GD}} = \underbrace{\operatorname*{mean}_i u_i^\top W^{\mathrm{GD}} v_i}_{\text{diagonal}}
  - \underbrace{\operatorname*{mean}_i \operatorname*{mean}_{j \neq i} u_j^\top W^{\mathrm{GD}} v_i}_{\text{off-diagonal}} .
  \label{eq:hgd-split}
\end{equation}

\begin{lemma}[Diagonal-sum identity]
\label{lem:diag-gd}
The diagonal scores of $W^{\mathrm{GD}}$ sum to
\[
  \sum_{i=1}^N u_i^\top W^{\mathrm{GD}} v_i = \lVert W^{\mathrm{GD}} \rVert_F^2
  + \frac{1}{N} (U\mathbf{1})^\top W^{\mathrm{GD}} (V\mathbf{1}) .
\]
\end{lemma}

\begin{proof}
A scalar equals its own trace and $\sum_i u_i v_i^\top = UV^\top$, so $\sum_i u_i^\top W^{\mathrm{GD}} v_i = \operatorname{tr}\big( (W^{\mathrm{GD}})^\top \sum_i u_i v_i^\top \big) = \langle W^{\mathrm{GD}}, UV^\top \rangle$. Rearranging \cref{eq:one-step} gives $UV^\top = W^{\mathrm{GD}} + \tfrac1N (U\mathbf{1})(V\mathbf{1})^\top$, and bilinearity together with $\langle Z, ab^\top \rangle = a^\top Z b$ gives the claim.
\end{proof}

\begin{lemma}[Rank-one cross term]
\label{lem:cross-gd}
$\tfrac{1}{N^2} \lvert (U\mathbf{1})^\top W^{\mathrm{GD}} (V\mathbf{1}) \rvert = O_p(d)$.
\end{lemma}

\begin{proof}
By \cref{lem:norms-wgd}(ii) and \cref{lem:conc-uv}(iii),(iv),
\begin{equation}
  \frac{\lvert (U\mathbf{1})^\top W^{\mathrm{GD}} (V\mathbf{1}) \rvert}{N^2}
  \leq \frac{\lVert U\mathbf{1} \rVert \, \lVert W^{\mathrm{GD}} \rVert_{\mathrm{op}} \, \lVert V\mathbf{1} \rVert}{N^2}
  = \frac{O_p(\sqrt{Nd}) \cdot O_p(N) \cdot O_p(\sqrt{Nd})}{N^2} = O_p(d) . \qedhere
\end{equation}
\end{proof}

\begin{lemma}
\label{lem:diagmean-gd}
$\operatorname{mean}_i u_i^\top W^{\mathrm{GD}} v_i = d^2 (1 + O_p(d^{-1}))$.
\end{lemma}

\begin{proof}
Divide \cref{lem:diag-gd} by $N$. By \cref{lem:norms-wgd}(i), $\lVert W^{\mathrm{GD}} \rVert_F^2 / N = \tfrac{N-1}{N} d^2 (1 + O_p(d^{-1})) = d^2 (1 + O_p(d^{-1}))$, using $\tfrac1N \leq \tfrac1d$; by \cref{lem:cross-gd} the second term is $O_p(d) = d^2 \cdot O_p(d^{-1})$.
\end{proof}

\begin{lemma}[Column-sum identity]
\label{lem:colsum}
For every $W \in \mathbb{R}^{d \times d}$ and every $i$, $(N-1) \operatorname{mean}_{j \neq i} u_j^\top W v_i = (U\mathbf{1})^\top W v_i - u_i^\top W v_i$.
\end{lemma}

\begin{proof}
$(N-1)\operatorname{mean}_{j \neq i} u_j^\top W v_i = \sum_j u_j^\top W v_i - u_i^\top W v_i$ and $\sum_j u_j = U\mathbf{1}$.
\end{proof}

\begin{lemma}[Two sums of squares]
\label{lem:twosums-gd}
$\sum_i \big( (U\mathbf{1})^\top W^{\mathrm{GD}} v_i \big)^2 = O_p(N^4 d)$ and $\sum_i \big( u_i^\top W^{\mathrm{GD}} v_i \big)^2 = O_p(N^4 d)$.
\end{lemma}

\begin{proof}
The numbers $(U\mathbf{1})^\top W^{\mathrm{GD}} v_i$ are the entries of $V^\top (W^{\mathrm{GD}})^\top U\mathbf{1}$, so by \cref{lem:norms-wgd}(ii) and \cref{lem:conc-uv}(iii),(iv),
\begin{equation}
  \big\lVert V^\top (W^{\mathrm{GD}})^\top U\mathbf{1} \big\rVert^2
  \leq \lVert V \rVert_{\mathrm{op}}^2 \lVert W^{\mathrm{GD}} \rVert_{\mathrm{op}}^2 \lVert U\mathbf{1} \rVert^2
  = O_p(N \cdot N^2 \cdot Nd) = O_p(N^4 d) .
\end{equation}
For the second sum, $u_i^\top W^{\mathrm{GD}} v_i = [U^\top W^{\mathrm{GD}} V]_{ii}$ and $\sum_i X_{ii}^2 \leq \lVert X \rVert_F^2$; by \cref{eq:one-step}, $U^\top W^{\mathrm{GD}} V = (U^\top U)(I_N - \tfrac{\mathbf{1}\mathbf{1}^\top}{N})(V^\top V)$, and since the nonzero eigenvalues of $V^\top V$ and $VV^\top$ coincide, $\lVert V^\top V \rVert_F = \lVert VV^\top \rVert_F$, so by \cref{lem:conc-uv}(iv),(vi),
\begin{equation}
  \sum_i \big( u_i^\top W^{\mathrm{GD}} v_i \big)^2 \leq \lVert U^\top W^{\mathrm{GD}} V \rVert_F^2
  \leq \lVert U \rVert_{\mathrm{op}}^4 \lVert VV^\top \rVert_F^2 = O_p(N^2 \cdot N^2 d) = O_p(N^4 d) . \qedhere
\end{equation}
\end{proof}

\begin{lemma}
\label{lem:offdiag-gd}
$\operatorname{mean}_i \big( \operatorname{mean}_{j \neq i} u_j^\top W^{\mathrm{GD}} v_i \big)^2 = O_p(\gamma^2 d^2)$, hence $\lvert \operatorname{mean}_i \operatorname{mean}_{j \neq i} u_j^\top W^{\mathrm{GD}} v_i \rvert = O_p(\gamma d)$.
\end{lemma}

\begin{proof}
Square \cref{lem:colsum} with $W = W^{\mathrm{GD}}$, use $(x-y)^2 \leq 2x^2 + 2y^2$, sum over $i$ and apply \cref{lem:twosums-gd}:
\begin{equation}
  \sum_i \Big( (N-1) \operatorname*{mean}_{j \neq i} u_j^\top W^{\mathrm{GD}} v_i \Big)^2 = O_p(N^4 d) \leq O_p(N^5) ,
\end{equation}
using $d \leq N$. Dividing by $N(N-1)^2 \geq \tfrac14 N^3$ and substituting $N = \gamma d$ gives the first claim; the second follows from Cauchy--Schwarz, $\lvert \operatorname{mean}_i x_i \rvert \leq (\operatorname{mean}_i x_i^2)^{1/2}$.
\end{proof}

\begin{lemma}[$\bar h$ for GD]
\label{lem:hbar-gd}
$\bar h^{\mathrm{GD}} = d^2 \big( 1 + O_p(\gamma/d) \big)$.
\end{lemma}

\begin{proof}
By \cref{eq:hgd-split}, \cref{lem:diagmean-gd} and \cref{lem:offdiag-gd}, $\bar h^{\mathrm{GD}} = d^2(1 + O_p(d^{-1})) - O_p(\gamma d) = d^2 (1 + O_p(d^{-1}) + O_p(\gamma/d))$, and $\gamma \geq 1$ absorbs the first error term into the second.
\end{proof}

We turn to $\bar\kappa^{\mathrm{GD}}$. By \cref{eq:general-hk},
\begin{equation}
  (\bar\kappa^{\mathrm{GD}})^2 = \underbrace{\frac{1}{N(N-1)} \sum_{i=1}^N \sum_{j \neq i} \big( u_j^\top W^{\mathrm{GD}} v_i \big)^2}_{\text{second moment}}
  - \underbrace{\operatorname*{mean}_i \Big( \operatorname*{mean}_{j \neq i} u_j^\top W^{\mathrm{GD}} v_i \Big)^2}_{\text{squared mean}} .
  \label{eq:kgd-split}
\end{equation}

\begin{lemma}[Expectation of the second moment]
\label{lem:exp-gd}
\begin{equation}
\begin{aligned}
  \mathbb{E} \sum_{i=1}^N \sum_{j \neq i} \big( u_j^\top W^{\mathrm{GD}} v_i \big)^2
  = {}&N(N-1) d^2 (N + 2d + 2) \\
  &\quad {}- d^2 \Big[ \Big( 1 + \frac1N \Big) (N+d+1)^2 - 2(d+2)^2 - 2(N-1) \Big] \\
  = {}&N(N-1) d^2 (N + 2d + 2) \big( 1 + O(N^{-1}) \big) .
\end{aligned}
  \label{eq:exp-gd}
\end{equation}
\end{lemma}

\begin{proof}
Write $u_j^\top W^{\mathrm{GD}} v_i = [U^\top W^{\mathrm{GD}} V]_{ji}$; by \cref{eq:one-step}, $U^\top W^{\mathrm{GD}} V = (U^\top U)(I_N - \tfrac{\mathbf{1}\mathbf{1}^\top}{N})(V^\top V)$, i.e.\ $u_j^\top W^{\mathrm{GD}} v_i = \sum_{k,l} (u_j^\top u_k)(\delta_{kl} - \tfrac1N)(v_l^\top v_i)$, with $U$ and $V$ independent. Two second moments, identical for $u$ and $v$, are used:
\begin{equation}
  \mathbb{E}\big[ (u_k^\top u_l)(u_l^\top u_m) \big] =
  \begin{cases} d^2 + 2d, &k = l = m \\ d, &k = m \neq l \\ 0, &\text{otherwise,} \end{cases}
  \qquad
  \mathbb{E}\big[ (u_i^\top u_k)(u_i^\top u_{k'}) \big] = \delta_{kk'} d \big[ 1 + (d+1)\delta_{ki} \big] .
\end{equation}
The first of these gives $\big[ \mathbb{E} (U^\top U)^2 \big]_{km} = \sum_l \mathbb{E} \big[ (u_k^\top u_l)(u_l^\top u_m) \big] = \delta_{km} \big[ (d^2 + 2d) + (N-1) d \big] = \delta_{km} d (d + N + 1)$, i.e.\ $\mathbb{E}(U^\top U)^2 = d(d+N+1) I_N$, and likewise for $V$. \emph{All pairs $(i,j)$.} Since $\lVert X \rVert_F^2 = \operatorname{tr}(X^\top X)$ and by cyclicity,
\begin{equation}
  \sum_{i,j} \big( u_j^\top W^{\mathrm{GD}} v_i \big)^2 = \lVert U^\top W^{\mathrm{GD}} V \rVert_F^2
  = \operatorname{tr}\Big[ (U^\top U)^2 \Big( I_N - \frac{\mathbf{1}\mathbf{1}^\top}{N} \Big) (V^\top V)^2 \Big( I_N - \frac{\mathbf{1}\mathbf{1}^\top}{N} \Big) \Big] ,
\end{equation}
and by independence and $\operatorname{tr}[(I_N - \tfrac{\mathbf{1}\mathbf{1}^\top}{N})^2] = N-1$,
\begin{equation}
  \mathbb{E} \sum_{i,j} \big( u_j^\top W^{\mathrm{GD}} v_i \big)^2 = \big[ d(d + N + 1) \big]^2 (N-1) .
\end{equation}
\emph{Diagonal $j = i$.} Using the independence of $U$ and $V$, the second display above for both factors, and then $(\delta_{kl} - \tfrac1N)^2 = \delta_{kl}(1 - \tfrac2N) + \tfrac{1}{N^2}$ together with $\sum_k [1 + (d+1)\delta_{ki}] = N + d + 1$ and $\sum_k [1 + (d+1)\delta_{ki}]^2 = N - 1 + (d+2)^2$,
\begin{equation}
\begin{aligned}
  \mathbb{E}\big( u_i^\top W^{\mathrm{GD}} v_i \big)^2
  &= \sum_{k,l,k',l'} \mathbb{E}\big[ (u_i^\top u_k)(u_i^\top u_{k'}) \big] \,
     \mathbb{E}\big[ (v_l^\top v_i)(v_{l'}^\top v_i) \big] \Big( \delta_{kl} - \frac1N \Big) \Big( \delta_{k'l'} - \frac1N \Big) \\
  &= d^2 \sum_{k,l} \big[ 1 + (d+1)\delta_{ki} \big] \big[ 1 + (d+1)\delta_{li} \big] \Big( \delta_{kl} - \frac1N \Big)^2 \\
  &= d^2 \Big[ \Big( 1 - \frac2N \Big) \big( N - 1 + (d+2)^2 \big) + \frac{(N+d+1)^2}{N^2} \Big] .
\end{aligned}
\end{equation}
\emph{Subtracting.} $\mathbb{E} \sum_i \sum_{j \neq i} = \mathbb{E} \sum_{i,j} - N \mathbb{E}(u_i^\top W^{\mathrm{GD}} v_i)^2$; rearranging with $(N+d+1)^2 - (d+2)^2 - (N-1) = (N-1)(N + 2d + 2)$ gives the first equality. Since $d \leq N$, the discarded terms are bounded by $38 N^2 d^2 \leq \tfrac{38}{N-1} \cdot N(N-1)d^2(N + 2d + 2)$, which is the stated relative error.
\end{proof}

\begin{lemma}[Variance of the second moment]
\label{lem:var-gd}
There is an absolute constant $C$ with $\operatorname{var}\big[ \sum_{i} \sum_{j \neq i} (u_j^\top W^{\mathrm{GD}} v_i)^2 \big] \leq C N^7 d$.
\end{lemma}

\begin{proof}
By \cref{eq:one-step}, $\sum_i \sum_{j \neq i} (u_j^\top W^{\mathrm{GD}} v_i)^2 = \lVert U^\top W^{\mathrm{GD}} V \rVert_F^2 - \sum_i [U^\top W^{\mathrm{GD}} V]_{ii}^2$, a smooth function of the $2Nd$ independent standard Gaussian entries of $(U,V)$. By the Gaussian Poincar\'e inequality the variance is at most $\mathbb{E} \lVert \nabla_U \cdot \rVert_F^2 + \mathbb{E} \lVert \nabla_V \cdot \rVert_F^2$.

The gradient of $M \mapsto \lVert M \rVert_F^2 - \sum_i M_{ii}^2$ is $2M - 2 \operatorname{diag}(M)$, of Frobenius norm at most $2 \lVert M \rVert_F$; the directional derivative of $U \mapsto (U^\top U)(I_N - \tfrac{\mathbf{1}\mathbf{1}^\top}{N})(V^\top V)$ along $H$ has Frobenius norm at most $2 \lVert H \rVert_F \lVert U \rVert_{\mathrm{op}} \lVert V \rVert_{\mathrm{op}}^2$. With $\lVert U^\top W^{\mathrm{GD}} V \rVert_F \leq \lVert U \rVert_{\mathrm{op}}^2 \lVert VV^\top \rVert_F$, the chain rule gives
\begin{equation}
  \Big\lVert \nabla_U \sum_i \sum_{j \neq i} (u_j^\top W^{\mathrm{GD}} v_i)^2 \Big\rVert_F
  \leq 4 \lVert U \rVert_{\mathrm{op}}^3 \lVert V \rVert_{\mathrm{op}}^2 \lVert VV^\top \rVert_F .
\end{equation}
Squaring, taking expectations and applying H\"older together with \cref{lem:conc-uv}(iv),(vi) gives $\mathbb{E} \lVert \nabla_U \cdot \rVert_F^2 \leq 16 (\mathbb{E} \lVert U \rVert_{\mathrm{op}}^{18})^{1/3} (\mathbb{E} \lVert V \rVert_{\mathrm{op}}^{12})^{1/3} (\mathbb{E} \lVert VV^\top \rVert_F^6)^{1/3} \leq 16 (C_9 C_6^2)^{1/3} N^7 d$. The expression is invariant under transposition combined with swapping $U$ and $V$, so the same bound holds for $\nabla_V$; adding the two gives the claim with $C = 32 (C_9 C_6^2)^{1/3}$.
\end{proof}

\begin{lemma}[Second moment]
\label{lem:secondmoment-gd}
The second moment of the off-diagonal GD scores satisfies
\[
  \frac{1}{N(N-1)} \sum_{i=1}^N \sum_{j \neq i} \big( u_j^\top W^{\mathrm{GD}} v_i \big)^2
  = d^3 (2 + \gamma) \big( 1 + O_p(\sqrt{\gamma}/d) \big) .
\]
\end{lemma}

\begin{proof}
By \cref{eq:exp-gd} and $N(N-1)d^2(N + 2d + 2) \geq N^2(N-1)d^2 \geq \tfrac12 N^3 d^2$ for $N \geq 2$, the expectation is at least $\tfrac14 N^3 d^2$ for $N$ large. Combined with \cref{lem:var-gd}, the relative standard deviation is at most $\sqrt{C N^7 d} / (\tfrac14 N^3 d^2) = 4\sqrt{C} \sqrt{N/d^3} = 4\sqrt{C} \sqrt{\gamma}/d$, so Chebyshev gives
\begin{equation}
  \sum_i \sum_{j \neq i} \big( u_j^\top W^{\mathrm{GD}} v_i \big)^2
  = \Big( \mathbb{E} \sum_i \sum_{j \neq i} \big( u_j^\top W^{\mathrm{GD}} v_i \big)^2 \Big) \big( 1 + O_p(\sqrt{\gamma}/d) \big) .
\end{equation}
Since $N^{-1} = \gamma^{-1} d^{-1} \leq \sqrt{\gamma}/d$, the relative error $O(N^{-1})$ of \cref{lem:exp-gd} is absorbed; dividing by $N(N-1)$ leaves $d^2(N + 2d + 2) = d^3(\gamma + 2 + \tfrac2d) = d^3(2+\gamma)(1 + \tfrac{2}{d(2+\gamma)})$ and $\tfrac{2}{d(2+\gamma)} \leq \tfrac1d \leq \sqrt{\gamma}/d$.
\end{proof}

\begin{lemma}[$\bar\kappa$ for GD]
\label{lem:kbar-gd}
$(\bar\kappa^{\mathrm{GD}})^2 = d^3 (2 + \gamma) \big( 1 + O_p(\gamma/d) \big)$.
\end{lemma}

\begin{proof}
By \cref{lem:offdiag-gd} the squared mean in \cref{eq:kgd-split} is $O_p(\gamma^2 d^2) = d^3(2+\gamma) \cdot O_p\big( \tfrac{\gamma^2}{d(2+\gamma)} \big) = d^3(2+\gamma) \cdot O_p(\gamma/d)$. Subtracting it from \cref{lem:secondmoment-gd} and using $\sqrt\gamma \leq \gamma$ gives the claim.
\end{proof}

\begin{lmmbox}{$\rho$ for GD}{lem:rho-gd}
The margin-to-spread ratio of the GD iterate is
\[
  \rho^{\mathrm{GD}} = \sqrt{\frac{d}{2+\gamma}} \big( 1 + O_p(\gamma/d) \big)
  = \sqrt{\frac{d}{\gamma}} \Big( 1 - \frac{1}{\gamma} + O(\gamma^{-2}) + o_p(1) \Big) .
\]
\end{lmmbox}
Both $\bar h^{\mathrm{GD}}$ and $\bar\kappa^{\mathrm{GD}}$ are a power of $d$ times a function of $\gamma$ alone, so $\rho^{\mathrm{GD}}$ depends on $d$ only through the prefactor $\sqrt d$.

\begin{proof}
By \cref{lem:hbar-gd}, \cref{lem:kbar-gd} and $\sqrt{1 + O_p(\gamma/d)} = 1 + O_p(\gamma/d)$,
\begin{equation}
  \rho^{\mathrm{GD}} = \frac{\bar h^{\mathrm{GD}}}{\bar\kappa^{\mathrm{GD}}}
  = \frac{d^2 \big( 1 + O_p(\gamma/d) \big)}{d^{3/2} \sqrt{2+\gamma} \big( 1 + O_p(\gamma/d) \big)}
  = \sqrt{\frac{d}{2+\gamma}} \big( 1 + O_p(\gamma/d) \big) .
\end{equation}
With $\gamma$ fixed and $d \to \infty$, $O_p(\gamma/d) = o_p(1)$; moreover $\sqrt{\gamma/(2+\gamma)} = (1 + 2/\gamma)^{-1/2} = 1 - \gamma^{-1} + O(\gamma^{-2})$.
\end{proof}

\subsection{The Muon step}

Throughout this subsection $\gamma \geq 16$. By \cref{eq:general-hk} with $W = W^{\mathrm{Muon}}$ the same two decompositions hold as in \cref{eq:hgd-split} and \cref{eq:kgd-split}, with $W^{\mathrm{GD}}$ replaced by $W^{\mathrm{Muon}}$. We use throughout that $W^{\mathrm{Muon}}$ is orthogonal (\cref{lem:fullrank}).

\begin{lemma}[Diagonal-sum identity]
\label{lem:diag-muon}
The diagonal scores of $W^{\mathrm{Muon}}$ sum to
\[
  \sum_{i=1}^N u_i^\top W^{\mathrm{Muon}} v_i = \lVert W^{\mathrm{GD}} \rVert_*
  + \frac{1}{N}(U\mathbf{1})^\top W^{\mathrm{Muon}} (V\mathbf{1}) .
\]
\end{lemma}

\begin{proof}
As in \cref{lem:diag-gd}, $\sum_i u_i^\top W^{\mathrm{Muon}} v_i = \langle W^{\mathrm{Muon}}, UV^\top \rangle = \langle W^{\mathrm{Muon}}, W^{\mathrm{GD}} \rangle + \tfrac1N (U\mathbf{1})^\top W^{\mathrm{Muon}}(V\mathbf{1})$. Writing the compact SVD $W^{\mathrm{GD}} = P\Sigma Q^\top$, so that $W^{\mathrm{Muon}} = PQ^\top$, and using $P^\top P = Q^\top Q = I$,
\begin{equation}
  \langle W^{\mathrm{Muon}}, W^{\mathrm{GD}} \rangle = \operatorname{tr}\big( QP^\top P \Sigma Q^\top \big)
  = \operatorname{tr}\big( \Sigma Q^\top Q \big) = \operatorname{tr} \Sigma = \lVert W^{\mathrm{GD}} \rVert_* . \qedhere
\end{equation}
\end{proof}

\begin{lemma}[Rank-one cross term]
\label{lem:cross-muon}
$\tfrac{1}{N^2} \lvert (U\mathbf{1})^\top W^{\mathrm{Muon}} (V\mathbf{1}) \rvert = O_p(d/N)$.
\end{lemma}

\begin{proof}
By $\lVert W^{\mathrm{Muon}} \rVert_{\mathrm{op}} \leq 1$ and \cref{lem:conc-uv}(iii), $\lvert (U\mathbf{1})^\top W^{\mathrm{Muon}}(V\mathbf{1}) \rvert / N^2 \leq \lVert U\mathbf{1} \rVert \lVert V\mathbf{1} \rVert / N^2 = O_p(Nd)/N^2 = O_p(d/N)$.
\end{proof}

\begin{lemma}
\label{lem:diagmean-muon}
$\operatorname{mean}_i u_i^\top W^{\mathrm{Muon}} v_i = \tfrac{\lVert W^{\mathrm{GD}} \rVert_*}{N} \big( 1 + O_p((Nd)^{-1/2}) \big)$.
\end{lemma}

\begin{proof}
Divide \cref{lem:diag-muon} by $N$ and use \cref{lem:cross-muon}. By \cref{lem:norms-wgd}(iii), with probability tending to one $\lVert W^{\mathrm{GD}} \rVert_* / N \geq d\sqrt{Nd}/(8N) = d^{3/2}/(8\sqrt{N})$, so the error term satisfies $O_p(d/N) = \tfrac{\lVert W^{\mathrm{GD}} \rVert_*}{N} O_p\big( \tfrac{d}{N} \tfrac{8\sqrt N}{d^{3/2}} \big) = \tfrac{\lVert W^{\mathrm{GD}} \rVert_*}{N} O_p((Nd)^{-1/2})$.
\end{proof}

\begin{lemma}[Two sums of squares]
\label{lem:twosums-muon}
$\sum_i \big( (U\mathbf{1})^\top W^{\mathrm{Muon}} v_i \big)^2 = O_p(N^2 d)$ and $\sum_i \big( u_i^\top W^{\mathrm{Muon}} v_i \big)^2 = O_p(Nd^2)$.
\end{lemma}

\begin{proof}
For the first, $W^{\mathrm{Muon}}$ is orthogonal, so $\lVert (W^{\mathrm{Muon}})^\top U\mathbf{1} \rVert = \lVert U\mathbf{1} \rVert$, and $\sum_i v_i v_i^\top = VV^\top$, whence by \cref{lem:conc-uv}(iii),(iv)
\begin{equation}
  \sum_i \big( (U\mathbf{1})^\top W^{\mathrm{Muon}} v_i \big)^2
  = (U\mathbf{1})^\top W^{\mathrm{Muon}} VV^\top (W^{\mathrm{Muon}})^\top (U\mathbf{1})
  \leq \lVert V \rVert_{\mathrm{op}}^2 \lVert U \mathbf{1} \rVert^2 = O_p(N \cdot Nd) .
\end{equation}
For the second, Cauchy--Schwarz and $\lVert W^{\mathrm{Muon}} v_i \rVert = \lVert v_i \rVert$ give $(u_i^\top W^{\mathrm{Muon}} v_i)^2 \leq \lVert u_i \rVert^2 \lVert v_i \rVert^2$, and $x^2 \leq 2(x-d)^2 + 2d^2$ with \cref{lem:conc-uv}(ii) gives $\sum_i \lVert u_i \rVert^2 \lVert v_i \rVert^2 \leq 2 \sum_i (\lVert u_i \rVert \lVert v_i \rVert - d)^2 + 2Nd^2 = N \cdot O_p(d) + 2Nd^2 = O_p(Nd^2)$.
\end{proof}

\begin{lemma}
\label{lem:offdiag-muon}
$\operatorname{mean}_i \big( \operatorname{mean}_{j \neq i} u_j^\top W^{\mathrm{Muon}} v_i \big)^2 = O_p(d/N)$, hence $\lvert \operatorname{mean}_i \operatorname{mean}_{j \neq i} u_j^\top W^{\mathrm{Muon}} v_i \rvert = O_p(\sqrt{d/N})$.
\end{lemma}

\begin{proof}
Square \cref{lem:colsum} with $W = W^{\mathrm{Muon}}$, use $(x-y)^2 \leq 2x^2 + 2y^2$, sum over $i$ and apply \cref{lem:twosums-muon}:
\begin{equation}
  \sum_i \Big( (N-1) \operatorname*{mean}_{j \neq i} u_j^\top W^{\mathrm{Muon}} v_i \Big)^2
  \leq 2 O_p(N^2 d) + 2 O_p(Nd^2) = O_p(N^2 d) ,
\end{equation}
using $d \leq N$. Dividing by $N(N-1)^2 \geq \tfrac14 N^3$ gives the first claim, and Cauchy--Schwarz the second.
\end{proof}

\begin{lemma}[$\bar h$ for Muon]
\label{lem:hbar-muon}
The mean margin of the Muon iterate is
\[
  \bar h^{\mathrm{Muon}} = \frac{\lVert W^{\mathrm{GD}} \rVert_*}{N} \big( 1 + O_p(d^{-1}) \big)
  = \frac{d \, \Phi(\gamma)}{\sqrt{\gamma}} \big( 1 + o_p(1) \big) .
\]
\end{lemma}

\begin{proof}
By \cref{lem:diagmean-muon} and \cref{lem:offdiag-muon}, $\bar h^{\mathrm{Muon}} = \tfrac{\lVert W^{\mathrm{GD}} \rVert_*}{N}(1 + O_p((Nd)^{-1/2})) - O_p(\sqrt{d/N})$. By \cref{lem:norms-wgd}(iii), with probability tending to one $\tfrac{\lVert W^{\mathrm{GD}} \rVert_*}{N} \geq \tfrac{d^{3/2}}{8\sqrt N}$, so
\begin{equation}
  O_p\big( \sqrt{d/N} \big) = \frac{\lVert W^{\mathrm{GD}} \rVert_*}{N} \cdot O_p\Big( \sqrt{\frac{d}{N}} \cdot \frac{8\sqrt N}{d^{3/2}} \Big)
  = \frac{\lVert W^{\mathrm{GD}} \rVert_*}{N} \cdot O_p(d^{-1}) ,
\end{equation}
and $(Nd)^{-1/2} \leq d^{-1}$, which gives the first equality. For the second, \cref{lem:norms-wgd}(iv) gives $\tfrac{\lVert W^{\mathrm{GD}} \rVert_*}{Nd} = \tfrac{\Phi(\gamma)}{\sqrt\gamma} + o_p(1)$, and $\tfrac{\Phi(\gamma)}{\sqrt\gamma} \geq \tfrac{1}{8\sqrt\gamma} > 0$ by \cref{lem:norms-wgd}(iii), so the additive $o_p(1)$ may be written multiplicatively.
\end{proof}

We now turn to $\bar\kappa^{\mathrm{Muon}}$. The orthogonality of $W^{\mathrm{Muon}}$ turns the second moment into three terms.

\begin{lemma}[Three-term identity]
\label{lem:threeterm}
Almost surely
\begin{equation}
\begin{aligned}
  \frac{1}{N(N-1)} \sum_{i=1}^N \sum_{j \neq i} \big( u_j^\top W^{\mathrm{Muon}} v_i \big)^2
  = {}&\frac{N}{N-1} \operatorname*{mean}_i \lVert v_i \rVert^2
  - \frac{\operatorname{mean}_i \big( u_i^\top W^{\mathrm{Muon}} v_i \big)^2}{N-1} \\
  &\quad {}+ \frac{\operatorname{tr}\big[ (UU^\top - N I_d) W^{\mathrm{Muon}} VV^\top (W^{\mathrm{Muon}})^\top \big]}{N(N-1)} .
\end{aligned}
\end{equation}
\end{lemma}

\begin{proof}
Since $(U^\top x)_j = u_j^\top x$ and $\lVert W^{\mathrm{Muon}} v_i \rVert = \lVert v_i \rVert$,
\begin{equation}
\begin{aligned}
  \sum_{j \neq i} \big( u_j^\top W^{\mathrm{Muon}} v_i \big)^2 &= \lVert U^\top W^{\mathrm{Muon}} v_i \rVert^2 - \big( u_i^\top W^{\mathrm{Muon}} v_i \big)^2 , \\
  \lVert U^\top W^{\mathrm{Muon}} v_i \rVert^2 &= N \lVert v_i \rVert^2 + \big( W^{\mathrm{Muon}} v_i \big)^\top \big( UU^\top - N I_d \big) \big( W^{\mathrm{Muon}} v_i \big) .
\end{aligned}
\end{equation}
Summing over $i$ and using $x^\top Z x = \operatorname{tr}(Zxx^\top)$ with $\sum_i v_i v_i^\top = VV^\top$ turns the middle term into the stated trace; dividing by $N(N-1)$ and using $\sum_i \lVert v_i \rVert^2 = N \operatorname{mean}_i \lVert v_i \rVert^2$ gives the claim.
\end{proof}

\begin{lemma}[Norm term]
\label{lem:normterm}
$\tfrac{N}{N-1} \operatorname{mean}_i \lVert v_i \rVert^2 = d \big( 1 + O(N^{-1}) + O_p((Nd)^{-1/2}) \big)$.
\end{lemma}

\begin{proof}
\cref{lem:conc-uv}(i) gives $\operatorname{mean}_i \lVert v_i \rVert^2 = d(1 + O_p((Nd)^{-1/2}))$, and $\tfrac{N}{N-1} = 1 + \tfrac{1}{N-1} = 1 + O(N^{-1})$.
\end{proof}

The trace term requires two auxiliary facts.

\begin{lemma}[Decomposition of $UU^\top$]
\label{lem:decomp-uu}
Almost surely $VV^\top - \tfrac1N (V\mathbf{1})(V\mathbf{1})^\top$ is invertible and there is $\Gamma \in \mathbb{R}^{(N-1-d) \times d}$ with
\begin{equation}
  UU^\top - \frac1N (U\mathbf{1})(U\mathbf{1})^\top
  = W^{\mathrm{GD}} \Big( VV^\top - \frac1N (V\mathbf{1})(V\mathbf{1})^\top \Big)^{-1} (W^{\mathrm{GD}})^\top + \Gamma^\top \Gamma ,
\end{equation}
where, conditionally on $V$, the entries of $\Gamma$ are i.i.d.\ $\mathcal{N}(0,1)$ and independent of $W^{\mathrm{GD}}$. Consequently, for every symmetric $Y \in \mathbb{R}^{d \times d}$ that is a function of $(V, W^{\mathrm{GD}})$,
\begin{equation}
\begin{aligned}
  \mathbb{E}\big[ \operatorname{tr}(\Gamma^\top \Gamma Y) \mid V, W^{\mathrm{GD}} \big] &= (N-1-d) \operatorname{tr} Y , \\
  \operatorname{var}\big[ \operatorname{tr}(\Gamma^\top \Gamma Y) \mid V, W^{\mathrm{GD}} \big] &= 2(N-1-d) \lVert Y \rVert_F^2 .
\end{aligned}
\end{equation}
\end{lemma}

\begin{proof}
\emph{Invertibility.} $VV^\top - \tfrac1N (V\mathbf{1})(V\mathbf{1})^\top = V(I_N - \tfrac{\mathbf{1}\mathbf{1}^\top}{N})V^\top \succeq 0$ and its determinant is a polynomial in $V$ which is nonzero at $V = [\, I_d \ \ -\sum_{k \leq d} e_k \ \ 0 \,]$, where it equals $\det(I_d + (\sum_k e_k)(\sum_k e_k)^\top) = 1 + d \neq 0$; so it is nonzero almost surely.

\emph{Basis.} If $c_0 \mathbf{1}^\top + c^\top V = 0$ for some $c_0 \in \mathbb{R}$ and $c \in \mathbb{R}^d$, then multiplying on the right by $(I_N - \tfrac{\mathbf{1}\mathbf{1}^\top}{N}) V^\top c$, and using $\mathbf{1}^\top (I_N - \tfrac{\mathbf{1}\mathbf{1}^\top}{N}) = 0$, gives $c^\top (VV^\top - \tfrac1N (V\mathbf{1})(V\mathbf{1})^\top) c = 0$, so $c = 0$ and then $c_0 = 0$. Hence $\{x \in \mathbb{R}^N : \mathbf{1}^\top x = 0, Vx = 0\}$ has dimension $N-1-d$ almost surely; stack an orthonormal basis of it as the columns of $R \in \mathbb{R}^{N \times (N-1-d)}$, so that $R^\top R = I$, $\mathbf{1}^\top R = 0$ and $VR = 0$, and set $\Gamma := R^\top U^\top$.

\emph{Projection identity.} Because $\mathbb{R}^N = \operatorname{span}\mathbf{1} \oplus \operatorname{col}\big( (I_N - \tfrac{\mathbf{1}\mathbf{1}^\top}{N})V^\top \big) \oplus \operatorname{col} R$ with the three subspaces pairwise orthogonal and of dimensions $1 + d + (N-1-d) = N$,
\begin{equation}
  I_N - \frac{\mathbf{1}\mathbf{1}^\top}{N}
  = \Big( I_N - \frac{\mathbf{1}\mathbf{1}^\top}{N} \Big) V^\top \Big( VV^\top - \frac1N (V\mathbf{1})(V\mathbf{1})^\top \Big)^{-1} V \Big( I_N - \frac{\mathbf{1}\mathbf{1}^\top}{N} \Big) + RR^\top .
  \label{eq:projid}
\end{equation}
Both sides are symmetric, so it suffices to check them on the three subspaces. Write $P := I_N - \tfrac{\mathbf{1}\mathbf{1}^\top}{N}$ and $G := VV^\top - \tfrac1N (V\mathbf{1})(V\mathbf{1})^\top$, and use $P^2 = P$, $VPV^\top = G$ and $PR = R - \tfrac1N \mathbf{1}(\mathbf{1}^\top R) = R$. For $x = \mathbf{1}$: $Px = 0$, $VPx = 0$ and $R^\top x = 0$, so both sides vanish. For $x = P V^\top y$: $Px = x$, while $V P x = V P V^\top y = Gy$, so the first term of the right-hand side is $P V^\top G^{-1} G y = x$, and $R^\top x = (VR)^\top y = 0$, so the second vanishes. For $x = Rz$: $Px = Rz = x$, while $VPx = VRz = 0$ kills the first term, and $RR^\top Rz = Rz = x$. In all three cases both sides agree. Multiplying \cref{eq:projid} on the left by $U$ and on the right by $U^\top$, and using $U(I_N - \tfrac{\mathbf{1}\mathbf{1}^\top}{N})U^\top = UU^\top - \tfrac1N (U\mathbf{1})(U\mathbf{1})^\top$ and $U(I_N - \tfrac{\mathbf{1}\mathbf{1}^\top}{N})V^\top = W^{\mathrm{GD}}$, gives the decomposition with $URR^\top U^\top = \Gamma^\top \Gamma$.

\emph{Distribution.} Conditionally on $V$, the rows of $U$ are i.i.d.\ $\mathcal{N}(0, I_N)$; the $a$-th column of $\Gamma$ is $R^\top U^\top e_a \sim \mathcal{N}(0, R^\top R) = \mathcal{N}(0, I_{N-1-d})$ with independent columns, so the entries of $\Gamma$ are i.i.d.\ standard Gaussian. The $a$-th row of $W^{\mathrm{GD}}$ is $\big( V(I_N - \tfrac{\mathbf{1}\mathbf{1}^\top}{N}) U^\top e_a \big)^\top$ and $\operatorname{Cov}\big( V(I_N - \tfrac{\mathbf{1}\mathbf{1}^\top}{N})U^\top e_a, R^\top U^\top e_a \big) = V(I_N - \tfrac{\mathbf{1}\mathbf{1}^\top}{N})R = 0$; being jointly Gaussian conditionally on $V$, $\Gamma$ and $W^{\mathrm{GD}}$ are conditionally independent.

\emph{Moments.} The rows of $\Gamma$ are i.i.d.\ $\mathcal{N}(0, I_d)$ and $\operatorname{tr}(\Gamma^\top \Gamma Y) = \sum_m (\Gamma^\top e_m)^\top Y (\Gamma^\top e_m)$; for $g \sim \mathcal{N}(0, I_d)$ and fixed symmetric $Y$, $\mathbb{E}[g^\top Y g] = \operatorname{tr} Y$ and $\operatorname{var}[g^\top Y g] = 2 \lVert Y \rVert_F^2$; summing over the $N-1-d$ rows gives the claim.
\end{proof}

\begin{lemma}[Trace of a commutator]
\label{lem:commutator}
Almost surely, with $S := \tfrac1N \big( VV^\top - \tfrac1N (V\mathbf{1})(V\mathbf{1})^\top \big)$,
\begin{equation}
  \Big\lvert \operatorname{tr}\Big[ W^{\mathrm{GD}} (NS)^{-1} (W^{\mathrm{GD}})^\top W^{\mathrm{Muon}} (NS) (W^{\mathrm{Muon}})^\top \Big] - \lVert W^{\mathrm{GD}} \rVert_F^2 \Big\rvert
  \leq \frac{2 \lVert S - I_d \rVert_{\mathrm{op}}^2}{\lambda_{\min}(S)} \lVert W^{\mathrm{GD}} \rVert_F^2 .
\end{equation}
\end{lemma}

\begin{proof}
\emph{A matrix inequality.} For symmetric $S \succ 0$ and symmetric $T$,
\begin{equation}
  \big\lvert \operatorname{tr}[S^{-1} T S T] - \lVert T \rVert_F^2 \big\rvert \leq \frac{2 \lVert S - I_d \rVert_{\mathrm{op}}^2 \lVert T \rVert_F^2}{\lambda_{\min}(S)} .
  \label{eq:matineq}
\end{equation}
Indeed, cyclicity gives $\operatorname{tr}[S^{-1}STT] = \lVert T \rVert_F^2$ and $\operatorname{tr}[(TS - ST)T] = 0$, so with $S^{-1} = I_d - S^{-1}(S - I_d)$,
\begin{equation}
  \operatorname{tr}[S^{-1}TST] - \lVert T \rVert_F^2 = \operatorname{tr}\big[ S^{-1}(TS - ST)T \big] = -\operatorname{tr}\big[ S^{-1}(S - I_d)(TS - ST)T \big] ,
\end{equation}
and $\lvert \operatorname{tr}(ABC) \rvert \leq \lVert A \rVert_{\mathrm{op}} \lVert B \rVert_F \lVert C \rVert_F$, $\lVert S^{-1}(S - I_d) \rVert_{\mathrm{op}} \leq \lVert S - I_d \rVert_{\mathrm{op}} / \lambda_{\min}(S)$ and $\lVert TS - ST \rVert_F \leq 2 \lVert S - I_d \rVert_{\mathrm{op}} \lVert T \rVert_F$ give \cref{eq:matineq}.

\emph{Polar decomposition.} With $W^{\mathrm{GD}} = P \Sigma Q^\top$ and $W^{\mathrm{Muon}} = PQ^\top$ we have $(W^{\mathrm{GD}})^\top W^{\mathrm{Muon}} = Q \Sigma Q^\top = (W^{\mathrm{Muon}})^\top W^{\mathrm{GD}}$ and $\lVert Q\Sigma Q^\top \rVert_F = \lVert \Sigma \rVert_F = \lVert W^{\mathrm{GD}} \rVert_F$.

\emph{Substitution.} Moving the leading $W^{\mathrm{GD}}$ to the end by cyclicity,
\begin{equation}
  \operatorname{tr}\Big[ W^{\mathrm{GD}} (NS)^{-1} (W^{\mathrm{GD}})^\top W^{\mathrm{Muon}} (NS) (W^{\mathrm{Muon}})^\top \Big]
  = \operatorname{tr}\Big[ (NS)^{-1} \big( Q\Sigma Q^\top \big) (NS) \big( Q \Sigma Q^\top \big) \Big] ,
\end{equation}
and applying \cref{eq:matineq} with $T = Q\Sigma Q^\top$ (the factors $N$ cancel) gives the claim.
\end{proof}

\begin{lmmbox}{Trace term}{lem:traceterm}
If $\gamma \geq 16$, then with probability tending to one
\begin{equation}
  \frac{1}{N(N-1)d} \Big\lvert \operatorname{tr}\Big[ \big( UU^\top - N I_d \big) W^{\mathrm{Muon}} VV^\top (W^{\mathrm{Muon}})^\top \Big] \Big\rvert \leq 145 \gamma^{-2} .
\end{equation}
\end{lmmbox}
The assumption $\gamma \geq 16$ is tailored to this estimate: through \cref{lem:conc-uv}(v) it controls how far $\tfrac1N VV^\top$ is from the identity, so that the fluctuation of $UU^\top$ around $N I_d$, seen through the Muon iterate, contributes only $O(\gamma^{-2})$ to the second moment.

\begin{proof}
Write $B := W^{\mathrm{Muon}} \big( VV^\top - \tfrac1N (V\mathbf{1})(V\mathbf{1})^\top \big) (W^{\mathrm{Muon}})^\top$. By \cref{lem:fullrank} orthogonal conjugation preserves $\lVert \cdot \rVert_F$, $\lVert \cdot \rVert_{\mathrm{op}}$ and the trace; moreover $VV^\top - \tfrac1N (V\mathbf{1})(V\mathbf{1})^\top = V(I_N - \tfrac{\mathbf{1}\mathbf{1}^\top}{N}) V^\top$ satisfies $0 \preceq VV^\top - \tfrac1N (V\mathbf{1})(V\mathbf{1})^\top \preceq VV^\top$. Hence, by \cref{lem:conc-uv}(iii),(iv),(vi)
\begin{equation}
\begin{aligned}
  \lVert B \rVert_F &\leq \lVert VV^\top \rVert_F + \frac{\lVert V\mathbf{1} \rVert^2}{N} = O_p(N\sqrt d) , \\
  \lVert B \rVert_{\mathrm{op}} &\leq \lVert V \rVert_{\mathrm{op}}^2 = O_p(N) , \\
  \operatorname{tr} B &= \sum_k \lVert v_k \rVert^2 - \frac{\lVert V\mathbf{1} \rVert^2}{N} = O_p(Nd) .
\end{aligned}
  \label{eq:Bbounds}
\end{equation}
\emph{Rank-one replacement.} $VV^\top - (VV^\top - \tfrac1N (V\mathbf{1})(V\mathbf{1})^\top) = \tfrac1N (V\mathbf{1})(V\mathbf{1})^\top$, so by \cref{lem:conc-uv}(iii),(v),
\begin{equation}
\begin{aligned}
  \Big\lvert \operatorname{tr}\big[ (UU^\top &- NI_d) W^{\mathrm{Muon}} VV^\top (W^{\mathrm{Muon}})^\top \big] - \operatorname{tr}\big[ (UU^\top - NI_d) B \big] \Big\rvert \\
  &\leq \frac{1}{N} \lVert UU^\top - NI_d \rVert_{\mathrm{op}} \lVert V\mathbf{1} \rVert^2 = N(N-1)d \cdot O_p\big( \sqrt{d} N^{-3/2} \big) .
\end{aligned}
\end{equation}
\emph{Splitting.} By \cref{lem:decomp-uu} and $d + (N-1-d) + 1 = N$,
\begin{equation}
  UU^\top - NI_d = \Big[ W^{\mathrm{GD}} \big( NS \big)^{-1} (W^{\mathrm{GD}})^\top - d I_d \Big]
  + \Big[ \Gamma^\top \Gamma - (N-1-d) I_d \Big] + \Big[ \frac1N (U\mathbf{1})(U\mathbf{1})^\top - I_d \Big] ,
\end{equation}
and we bound the trace of each bracket against $B$ separately.

\emph{First bracket.} Since $\operatorname{tr} B$ appears, write
\begin{equation}
\begin{aligned}
  \operatorname{tr}\big[ (W^{\mathrm{GD}}(NS)^{-1}(W^{\mathrm{GD}})^\top - dI_d) B \big]
  = {}& \big( \operatorname{tr}[W^{\mathrm{GD}}(NS)^{-1}(W^{\mathrm{GD}})^\top B] - \lVert W^{\mathrm{GD}} \rVert_F^2 \big) \\
  &\quad {}+ \big( \lVert W^{\mathrm{GD}} \rVert_F^2 - d \operatorname{tr} B \big) .
\end{aligned}
\end{equation}
By \cref{lem:commutator}, \cref{lem:conc-uv}(v) ($\gamma \geq 16$) and \cref{lem:norms-wgd}(i), with probability tending to one the first difference is at most
\begin{equation}
  \frac{2(3\gamma^{-1/2})^2}{1/8} \lVert W^{\mathrm{GD}} \rVert_F^2
  = 144 \gamma^{-1} (N-1) d^2 \big( 1 + O_p(d^{-1}) \big)
  = N(N-1)d \cdot 144\gamma^{-2} \big( 1 + O_p(d^{-1}) \big) ,
\end{equation}
while the second is $O_p(\sqrt N d^{3/2} + Nd) = N(N-1)d \cdot O_p(\sqrt d N^{-3/2} + N^{-1})$ by \cref{lem:norms-wgd}(i).

\emph{Second bracket.} $B$ is symmetric and a function of $(V, W^{\mathrm{GD}})$, so \cref{lem:decomp-uu} gives $\mathbb{E}\big[ \operatorname{tr}[(\Gamma^\top \Gamma - (N-1-d)I_d) B] \mid V, W^{\mathrm{GD}} \big] = 0$ and, by \cref{eq:Bbounds}, conditional variance $2(N-1-d) \lVert B \rVert_F^2 = O(N) \cdot O_p(N^2 d) = O_p(N^3 d)$. The conditional Chebyshev inequality then gives $\operatorname{tr}[(\Gamma^\top \Gamma - (N-1-d)I_d)B] = O_p(N^{3/2} d^{1/2}) = N(N-1)d \cdot O_p((Nd)^{-1/2})$.

\emph{Third bracket.} By $\operatorname{tr}(aa^\top Y) = a^\top Y a \leq \lVert a \rVert^2 \lVert Y \rVert_{\mathrm{op}}$, \cref{eq:Bbounds} and \cref{lem:conc-uv}(iii),
\begin{equation}
\begin{aligned}
  \Big\lvert \operatorname{tr}\Big[ \big( \frac1N (U\mathbf{1})(U\mathbf{1})^\top - I_d \big) B \Big] \Big\rvert
  &\leq \frac{\lVert U\mathbf{1} \rVert^2}{N} \lVert B \rVert_{\mathrm{op}} + \lvert \operatorname{tr} B \rvert \\
  &= O_p(d) \cdot O_p(N) + O_p(Nd) = N(N-1)d \cdot O_p(N^{-1}) .
\end{aligned}
\end{equation}
\emph{Total.} Adding the four contributions and dividing by $N(N-1)d$, with probability tending to one the left-hand side is at most $144\gamma^{-2}(1 + O_p(d^{-1})) + o_p(1) = 144\gamma^{-2} + o_p(1)$, which is at most $145 \gamma^{-2}$ since $\gamma$ is fixed.
\end{proof}

The remaining term of \cref{lem:threeterm}, the diagonal one, needs the following sandwich.

\begin{lemma}[Nuclear-norm sandwich]
\label{lem:sandwich}
For all $a, b \in \mathbb{R}^d$,
\begin{equation}
  a^\top \operatorname{Orth}\big( W^{\mathrm{GD}} - ab^\top \big) b
  \;\leq\; \lVert W^{\mathrm{GD}} \rVert_* - \lVert W^{\mathrm{GD}} - ab^\top \rVert_*
  \;\leq\; a^\top W^{\mathrm{Muon}} b .
\end{equation}
\end{lemma}

\begin{proof}
For all $X, Y$, $\langle X, \operatorname{Orth}(Y) \rangle \leq \lVert X \rVert_* \lVert \operatorname{Orth}(Y) \rVert_{\mathrm{op}} \leq \lVert X \rVert_*$ by \cref{eq:norms}, and writing the compact SVD $Y = P\Sigma Q^\top$ gives $\langle Y, \operatorname{Orth}(Y) \rangle = \operatorname{tr}(Q \Sigma P^\top P Q^\top) = \lVert Y \rVert_*$; also $\langle ab^\top, Z \rangle = a^\top Z b$. Hence
\[
\begin{aligned}
  a^\top \operatorname{Orth}\big( W^{\mathrm{GD}} - ab^\top \big) b
  &= \big\langle W^{\mathrm{GD}}, \operatorname{Orth}(W^{\mathrm{GD}} - ab^\top) \big\rangle - \big\langle W^{\mathrm{GD}} - ab^\top, \operatorname{Orth}(W^{\mathrm{GD}} - ab^\top) \big\rangle \\
  &\leq \lVert W^{\mathrm{GD}} \rVert_* - \lVert W^{\mathrm{GD}} - ab^\top \rVert_* ,
\end{aligned}
\]
and symmetrically
\[
\begin{aligned}
  \lVert W^{\mathrm{GD}} \rVert_* - \lVert W^{\mathrm{GD}} - ab^\top \rVert_*
  &= \langle W^{\mathrm{GD}}, W^{\mathrm{Muon}} \rangle - \lVert W^{\mathrm{GD}} - ab^\top \rVert_* \\
  &\leq \langle W^{\mathrm{GD}}, W^{\mathrm{Muon}} \rangle - \langle W^{\mathrm{GD}} - ab^\top, W^{\mathrm{Muon}} \rangle
  = a^\top W^{\mathrm{Muon}} b .
\end{aligned}
\qedhere
\]
\end{proof}

\begin{lmmbox}{Self term}{lem:selfterm}
If $\gamma \geq 16$, then with probability tending to one
\begin{equation}
  0 \leq \frac{1}{(N-1)d} \operatorname*{mean}_i \big( u_i^\top W^{\mathrm{Muon}} v_i \big)^2 \leq 15 \gamma^{-3/2} .
\end{equation}
\end{lmmbox}
This bounds the diagonal term of \cref{lem:threeterm}; it is what forces the $\gamma^{-3/2}$ error in $\bar\kappa^{\mathrm{Muon}}$, and the only place where \cref{lem:sandwich} is used.

\begin{proof}
Nonnegativity is clear.

\emph{Step 1 (reduction to a first absolute moment).} By $\lVert W^{\mathrm{Muon}} \rVert_{\mathrm{op}} \leq 1$, $\lvert u_i^\top W^{\mathrm{Muon}} v_i \rvert \leq \lVert u_i \rVert \lVert v_i \rVert$, so
\begin{equation}
\begin{aligned}
  \operatorname*{mean}_i \big( u_i^\top W^{\mathrm{Muon}} v_i \big)^2
  \leq {}&d \operatorname*{mean}_i \big\lvert u_i^\top W^{\mathrm{Muon}} v_i \big\rvert \\
  &\quad {}+ \Big( \operatorname*{mean}_i \big( u_i^\top W^{\mathrm{Muon}} v_i \big)^2 \Big)^{1/2} \Big( \operatorname*{mean}_i \big( \lVert u_i \rVert \lVert v_i \rVert - d \big)^2 \Big)^{1/2} .
\end{aligned}
\end{equation}
For $x, c_0, c_1 \geq 0$, $x^2 \leq c_0 + c_1 x$ forces $x \leq \tfrac{c_1 + \sqrt{c_1^2 + 4c_0}}{2} \leq c_1 + \sqrt{c_0}$ and hence $x^2 \leq 2c_1^2 + 2c_0$; taking $x = (\operatorname{mean}_i (u_i^\top W^{\mathrm{Muon}} v_i)^2)^{1/2}$, $c_0 = d \operatorname{mean}_i \lvert u_i^\top W^{\mathrm{Muon}} v_i \rvert$ and $c_1 = (\operatorname{mean}_i (\lVert u_i \rVert \lVert v_i \rVert - d)^2)^{1/2}$, and using \cref{lem:conc-uv}(ii),
\begin{equation}
  \operatorname*{mean}_i \big( u_i^\top W^{\mathrm{Muon}} v_i \big)^2 \leq O_p(d) + 2d \operatorname*{mean}_i \big\lvert u_i^\top W^{\mathrm{Muon}} v_i \big\rvert .
  \label{eq:self-step1}
\end{equation}

\emph{Step 2 (a signed mean).} By \cref{lem:diag-muon} and \cref{lem:cross-muon}, $\operatorname{mean}_i u_i^\top W^{\mathrm{Muon}} v_i = \tfrac{\lVert W^{\mathrm{GD}} \rVert_*}{N} + O_p(d/N)$. For real $x \geq \ell$ one has
\begin{equation}
  \lvert x \rvert \leq x + 2 \lvert \ell \rvert :
  \label{eq:abs-trick}
\end{equation}
this is clear for $x \geq 0$, and for $x < 0$ it follows from $\lvert x \rvert = -x \leq -\ell \leq \lvert \ell \rvert \leq x + 2\lvert \ell \rvert$, the last step because $x + 2\lvert \ell \rvert \geq \ell + 2 \lvert \ell \rvert \geq \lvert \ell \rvert$.

\emph{Step 3 (conditioning).} Fix $i$ and set
\[
  a := \frac{u_i - \frac1N U\mathbf{1}}{\sqrt{1 - 1/N}} ,
  \qquad
  b := \frac{v_i - \frac1N V\mathbf{1}}{\sqrt{1 - 1/N}} .
\]
Since $\mathbf{1}^\top (e_i - \tfrac1N \mathbf{1}) = 0$ and $\lVert e_i - \tfrac1N \mathbf{1} \rVert^2 = 1 - \tfrac1N$, \cref{eq:one-step} gives
\begin{equation}
  W^{\mathrm{GD}} - ab^\top = U \Big[ \Big( I_N - \frac{\mathbf{1}\mathbf{1}^\top}{N} \Big) - \frac{(e_i - \frac1N \mathbf{1})(e_i - \frac1N\mathbf{1})^\top}{1 - 1/N} \Big] V^\top ,
\end{equation}
where the bracket, call it $M$, is the orthogonal projection onto $\{x \in \mathbb{R}^N : \mathbf{1}^\top x = 0, \ (e_i - \tfrac1N \mathbf{1})^\top x = 0\}$. Indeed $M = P - \Pi$ with $P := I_N - \tfrac{\mathbf{1}\mathbf{1}^\top}{N}$ and $\Pi := (e_i - \tfrac1N \mathbf{1})(e_i - \tfrac1N \mathbf{1})^\top / \lVert e_i - \tfrac1N \mathbf{1} \rVert^2$ the orthogonal projections onto $\mathbf{1}^\perp$ and onto $\operatorname{span}(e_i - \tfrac1N \mathbf{1})$; since $\mathbf{1}^\top(e_i - \tfrac1N \mathbf{1}) = 0$ the range of $\Pi$ lies in that of $P$, so $P\Pi = \Pi P = \Pi$ and hence $M^\top = M$, $M^2 = M$ and $M (e_i - \tfrac1N \mathbf{1}) = 0$. Consequently $a$ and $b$ are standard Gaussian and independent of $W^{\mathrm{GD}} - ab^\top$. For $a$: its $k$-th coordinate is $(e_i - \tfrac1N \mathbf{1})^\top U^\top e_k / \sqrt{1 - 1/N}$ and the $k$-th row of $UM$ is $M U^\top e_k$; conditionally on $V$ the vectors $U^\top e_k$ are i.i.d.\ $\mathcal{N}(0, I_N)$, and
\begin{equation}
  \operatorname{Cov}\Big( (e_i - \frac1N \mathbf{1})^\top U^\top e_k , \; M U^\top e_k \Big)
  = M (e_i - \frac1N \mathbf{1}) = 0 ,
\end{equation}
so, being jointly Gaussian, $a$ is independent of $UM$, and $\lVert (e_i - \tfrac1N \mathbf{1}) / \sqrt{1 - 1/N} \rVert = 1$ gives $a \sim \mathcal{N}(0, I_d)$. The same argument applied to the rows of $V$ gives $b \sim \mathcal{N}(0, I_d)$ independent of $VM$, and $U \perp V$ makes $a$, $b$ independent of each other; since $W^{\mathrm{GD}} - ab^\top = (UM)(VM)^\top$ by idempotence of $M$, the pair $(a,b)$ is independent of $W^{\mathrm{GD}} - ab^\top$.

\emph{Step 4 (two mean-square bounds).} For independent $a, b \sim \mathcal{N}(0,I_d)$ and $Z$ independent of them, $\mathbb{E}[(a^\top Z b)^2 \mid Z] = \lVert Z \rVert_F^2$; taking $Z = \operatorname{Orth}(W^{\mathrm{GD}} - ab^\top)$, whose squared Frobenius norm is its rank, at most $d$, and averaging over $i$,
\begin{equation}
  \Big( \operatorname*{mean}_i \big( a^\top \operatorname{Orth}(W^{\mathrm{GD}} - ab^\top) b \big)^2 \Big)^{1/2} = O_p(\sqrt d) .
  \label{eq:self-27}
\end{equation}
Substituting $u_i = \sqrt{1 - \tfrac1N} \, a + \tfrac1N U\mathbf{1}$ and $v_i = \sqrt{1 - \tfrac1N} \, b + \tfrac1N V\mathbf{1}$,
\begin{equation}
\begin{aligned}
  u_i^\top W^{\mathrm{Muon}} v_i - \Big( 1 - \frac1N \Big) a^\top W^{\mathrm{Muon}} b
  = {}&\sqrt{1 - \frac1N} \, a^\top W^{\mathrm{Muon}} \frac{V\mathbf{1}}{N}
  + \sqrt{1 - \frac1N} \, \frac{(U\mathbf{1})^\top}{N} W^{\mathrm{Muon}} b \\
  &\quad {}+ \frac{(U\mathbf{1})^\top}{N} W^{\mathrm{Muon}} \frac{V\mathbf{1}}{N} ,
\end{aligned}
\end{equation}
so by $(x + y + z)^2 \leq 3(x^2 + y^2 + z^2)$, $\lVert W^{\mathrm{Muon}} \rVert_{\mathrm{op}} \leq 1$, $\mathbb{E} \lVert a \rVert^2 = \mathbb{E} \lVert b \rVert^2 = d$ (so that $\operatorname{mean}_i \lVert a \rVert^2 = O_p(d)$ by Markov) and \cref{lem:conc-uv}(iii),
\begin{equation}
\begin{aligned}
  \operatorname*{mean}_i \Big( u_i^\top W^{\mathrm{Muon}} v_i &- \big( 1 - \frac1N \big) a^\top W^{\mathrm{Muon}} b \Big)^2 \\
  &\leq 3 \Big[ \operatorname*{mean}_i \lVert a \rVert^2 \frac{\lVert V\mathbf{1} \rVert^2}{N^2}
  + \frac{\lVert U\mathbf{1} \rVert^2}{N^2} \operatorname*{mean}_i \lVert b \rVert^2
  + \frac{\lVert U\mathbf{1} \rVert^2 \lVert V\mathbf{1} \rVert^2}{N^4} \Big] = O_p\Big( \frac{d^2}{N} \Big) ,
\end{aligned}
\end{equation}
that is
\begin{equation}
  \Big( \operatorname*{mean}_i \Big( u_i^\top W^{\mathrm{Muon}} v_i - \big( 1 - \frac1N \big) a^\top W^{\mathrm{Muon}} b \Big)^2 \Big)^{1/2} = O_p\big( d/\sqrt N \big) .
  \label{eq:self-28}
\end{equation}

\emph{Step 5 (combining).} By \cref{lem:sandwich} and $1 - \tfrac1N > 0$, $u_i^\top W^{\mathrm{Muon}} v_i \geq (1 - \tfrac1N) a^\top \operatorname{Orth}(W^{\mathrm{GD}} - ab^\top) b - \lvert u_i^\top W^{\mathrm{Muon}} v_i - (1 - \tfrac1N) a^\top W^{\mathrm{Muon}} b \rvert$. Using \cref{eq:abs-trick} of Step 2 with this lower bound, averaging over $i$, and applying \cref{eq:self-27}, \cref{eq:self-28} and $\lVert W^{\mathrm{GD}} \rVert_* \leq 7 d \sqrt{Nd}$ from \cref{lem:norms-wgd}(iii),
\begin{equation}
  \operatorname*{mean}_i \big\lvert u_i^\top W^{\mathrm{Muon}} v_i \big\rvert
  \leq \frac{\lVert W^{\mathrm{GD}} \rVert_*}{N} + O_p(d/N) + O_p(\sqrt d) + O_p(d/\sqrt N)
  \leq \frac{7 d^{3/2}}{\sqrt N} + O_p(\sqrt d) .
\end{equation}
Substituting into \cref{eq:self-step1} and dividing by $(N-1)d$,
\begin{equation}
\begin{aligned}
  \frac{1}{(N-1)d} \operatorname*{mean}_i \big( u_i^\top W^{\mathrm{Muon}} v_i \big)^2
  &\leq \frac{O_p(d)}{(N-1)d} + \frac{2}{N-1} \Big( \frac{7d^{3/2}}{\sqrt N} + O_p(\sqrt d) \Big) \\
  &= 14 \gamma^{-3/2} \Big( 1 + \frac{1}{N-1} \Big) + o_p(1) ,
\end{aligned}
\end{equation}
which is at most $15\gamma^{-3/2}$ with probability tending to one.
\end{proof}

\begin{lemma}[Second moment for Muon]
\label{lem:secondmoment-muon}
If $\gamma \geq 16$, then $\tfrac{1}{N(N-1)d} \sum_{i=1}^N \sum_{j \neq i} \big( u_j^\top W^{\mathrm{Muon}} v_i \big)^2 = 1 + O(\gamma^{-3/2}) + o_p(1)$, where with probability tending to one the $O(\gamma^{-3/2})$ term has absolute value at most $52 \gamma^{-3/2}$.
\end{lemma}

\begin{proof}
Divide \cref{lem:threeterm} by $d$ and subtract $1$. By \cref{lem:normterm} the norm term contributes $O(N^{-1}) + O_p((Nd)^{-1/2}) = o_p(1)$; by \cref{lem:traceterm} the trace term is at most $145\gamma^{-2} \leq \tfrac{145}{4} \gamma^{-3/2}$ in absolute value for $\gamma \geq 16$; by \cref{lem:selfterm} the diagonal term lies in $[-15\gamma^{-3/2}, 0]$. Adding, $\tfrac{145}{4} + 15 \leq 52$.
\end{proof}

\begin{lemma}[$\bar\kappa$ for Muon]
\label{lem:kbar-muon}
If $\gamma \geq 16$, then $(\bar\kappa^{\mathrm{Muon}})^2 = d \big( 1 + O(\gamma^{-3/2}) + o_p(1) \big)$ with the same constant as in \cref{lem:secondmoment-muon}.
\end{lemma}

\begin{proof}
By \cref{eq:kgd-split} with $W = W^{\mathrm{Muon}}$, \cref{lem:secondmoment-muon} and \cref{lem:offdiag-muon}, $(\bar\kappa^{\mathrm{Muon}})^2 = d(1 + O(\gamma^{-3/2}) + o_p(1)) - O_p(d/N) = d(1 + O(\gamma^{-3/2}) + o_p(1))$.
\end{proof}

\begin{lmmbox}{$\rho$ for Muon}{lem:rho-muon}
If $\gamma \geq 16$, then
\begin{equation}
  \rho^{\mathrm{Muon}} = \sqrt{\frac{d}{\gamma}} \Big[ \frac{8}{3\pi} \Big( 1 - \frac{1}{10\gamma} \Big) + O(\gamma^{-3/2}) + o_p(1) \Big] .
\end{equation}
\end{lmmbox}
Unlike for GD, the data budget enters $\rho^{\mathrm{Muon}}$ a second time, through $\Phi(\gamma)$.

\begin{proof}
By \cref{lem:hbar-muon}, $\bar h^{\mathrm{Muon}} / \sqrt d = \sqrt{d/\gamma} \, \Phi(\gamma) (1 + o_p(1))$. Set $y := (\bar\kappa^{\mathrm{Muon}})^2/d$. By \cref{lem:kbar-muon} and $\gamma \geq 16$, with probability tending to one
\[
  \lvert y - 1 \rvert \leq 52\gamma^{-3/2} + o_p(1) \leq \frac{52}{64} + o_p(1) \leq \frac{9}{10} ,
\]
so $y \geq \tfrac{1}{10}$. For $y \geq \tfrac{1}{10}$,
\[
  \lvert y^{-1/2} - 1 \rvert = \frac{\lvert 1 - y \rvert}{\sqrt y \, (1 + \sqrt y)} \leq \sqrt{10} \, \lvert y - 1 \rvert ,
\]
which gives $\sqrt d / \bar\kappa^{\mathrm{Muon}} = 1 + O(\gamma^{-3/2}) + o_p(1)$. Multiplying the two,
\begin{equation}
  \rho^{\mathrm{Muon}} = \frac{\bar h^{\mathrm{Muon}}}{\sqrt d} \cdot \frac{\sqrt d}{\bar\kappa^{\mathrm{Muon}}}
  = \sqrt{\frac{d}{\gamma}} \, \Phi(\gamma) \big( 1 + O(\gamma^{-3/2}) + o_p(1) \big) ,
\end{equation}
and \cref{lem:norms-wgd}(iv) gives $\Phi(\gamma) = \tfrac{8}{3\pi}(1 - \tfrac{1}{10\gamma} + r(\gamma))$ with $\lvert r(\gamma) \rvert \leq \gamma^{-2} \leq \tfrac14 \gamma^{-3/2}$, so $0 < \Phi(\gamma) \leq \tfrac{8}{3\pi}(1 + \gamma^{-2}) < 1$ and $\Phi(\gamma)(1 + O(\gamma^{-3/2}) + o_p(1)) = \tfrac{8}{3\pi}(1 - \tfrac{1}{10\gamma}) + O(\gamma^{-3/2}) + o_p(1)$.

Two by-products of this computation are used in the proof of \cref{thm:ratio}. First, the constant implied by the $O(\gamma^{-3/2})$ in the displayed multiplicative form is absolute, so there is an absolute $\gamma_1 \geq 16$ such that for $\gamma \geq \gamma_1$ that factor lies in $[\tfrac12, \tfrac32]$ with probability tending to one. Second, for every $\gamma \geq 16$ the same display bounds $\rho^{\mathrm{Muon}}$ away from zero without any expansion: $y \leq \tfrac{19}{10}$ gives $\sqrt d / \bar\kappa^{\mathrm{Muon}} \geq (19/10)^{-1/2} > \tfrac{7}{10}$, and $\Phi(\gamma) \geq \tfrac18$ by \cref{lem:norms-wgd}(iii),(iv), so with probability tending to one
\begin{equation}
  \rho^{\mathrm{Muon}} \;\geq\; \frac{7}{80} \sqrt{d/\gamma} \, (1 - o_p(1)) \;\geq\; \frac{1}{16} \sqrt{d/\gamma} .
  \label{eq:rho-muon-lower}
\end{equation}
\end{proof}

\subsection{Combining the GD and Muon estimates}

\begin{proof}[Proof of \cref{eq:ratio}]
\emph{Reduction to large $\gamma$.} By \cref{lem:rho-gd} and $2 + \gamma \geq \gamma$, $\rho^{\mathrm{GD}} \leq \sqrt{d/\gamma}\,(1 + o_p(1))$, so \cref{eq:rho-muon-lower} gives $\mathcal{R} \leq 17$ with probability tending to one, for every $\gamma \geq 16$; together with $0 \leq \tfrac{3\pi}{8}(1 - \tfrac{9}{10\gamma}) \leq \tfrac{3\pi}{8}$ this makes the left-hand side of \cref{eq:ratio} at most $19$ there. On a bounded range $16 \leq \gamma \leq \gamma_1$ we have $19 \leq 19\gamma_1^{3/2} \gamma^{-3/2}$, so \cref{eq:ratio} holds on that range as soon as $C_1 \geq 19 \gamma_1^{3/2}$. It therefore suffices to prove it for $\gamma \geq \gamma_1$, with $\gamma_1$ the absolute constant of \cref{lem:rho-muon}; on that range the error factor of \cref{lem:rho-muon} lies in $[\tfrac12, \tfrac32]$ with probability tending to one, so its reciprocal is again $1 + O(\gamma^{-3/2}) + o_p(1)$.

\emph{Expansion.} By \cref{lem:rho-gd} and \cref{lem:rho-muon}, and since $\Phi(\gamma) > 0$,
\begin{equation}
\begin{aligned}
  \mathcal{R} = \frac{\rho^{\mathrm{GD}}}{\rho^{\mathrm{Muon}}}
  &= \sqrt{\frac{d}{2+\gamma}} \big( 1 + o_p(1) \big) \cdot \sqrt{\frac{\gamma}{d}} \, \frac{1}{\Phi(\gamma) \big( 1 + O(\gamma^{-3/2}) + o_p(1) \big)} \\
  &= \frac{\sqrt{\gamma/(2+\gamma)}}{\Phi(\gamma)} \big( 1 + O(\gamma^{-3/2}) + o_p(1) \big) .
\end{aligned}
\end{equation}
The deterministic prefactor is expanded in $1/\gamma$: $\sqrt{\gamma/(2+\gamma)} = (1 + 2/\gamma)^{-1/2} = 1 - \gamma^{-1} + O(\gamma^{-2})$, while \cref{lem:norms-wgd}(iv) gives $\Phi(\gamma)^{-1} = \tfrac{3\pi}{8}\big( 1 + \tfrac{1}{10\gamma} + O(\gamma^{-2}) \big)$, so
\begin{equation}
  \frac{\sqrt{\gamma/(2+\gamma)}}{\Phi(\gamma)}
  = \frac{3\pi}{8} \Big( 1 - \frac{1}{\gamma} + O(\gamma^{-2}) \Big) \Big( 1 + \frac{1}{10\gamma} + O(\gamma^{-2}) \Big)
  = \frac{3\pi}{8} \Big( 1 - \frac{9}{10\gamma} \Big) + O(\gamma^{-2}) .
\end{equation}
For $\gamma \geq 16$ the prefactor is bounded, $0 < \sqrt{\gamma/(2+\gamma)}/\Phi(\gamma) \leq \tfrac{3\pi}{8} (1 - \tfrac{1}{160} - \tfrac{1}{256})^{-1}$, so multiplying by $1 + O(\gamma^{-3/2}) + o_p(1)$ leaves the error of the same order; and $\gamma^{-2} \leq \tfrac14 \gamma^{-3/2}$ in that range absorbs the $O(\gamma^{-2})$ term. Hence
\begin{equation}
  \mathcal{R} = \frac{3\pi}{8} \Big( 1 - \frac{9}{10\gamma} \Big) + O(\gamma^{-3/2}) + o_p(1) ,
\end{equation}
where, with probability tending to one, the $O(\gamma^{-3/2})$ term has absolute value at most $C_1 \gamma^{-3/2}$ with $C_1$ independent of $\gamma, d, N$. This is \cref{eq:ratio}.
\end{proof}

\subsection{Monotonicity in the data budget}

The expansion \cref{eq:ratio} has two consequences for how $\mathcal{R}$ changes with $\gamma$. The first compares a fixed pair of data budgets; the second concerns the limit $d \to \infty$.

\begin{lemma}[Fixed ratio]
\label{lem:fixed-ratio}
Let $\lambda > 0$ and
\[
  \gamma > \gamma_0(\lambda) := \max\Big\{ 16, \Big( \frac{160 C_1 (1+\lambda)}{27 \pi \lambda} \Big)^2 \Big\} .
\]
For the same $d$, let $N_1 = \gamma d$ and $N_2 = (1+\lambda)\gamma d$, and let $\mathcal{R}_{N_1, d}$ and $\mathcal{R}_{N_2, d}$ be the corresponding values of $\mathcal{R}$. Then, with probability tending to one, $\mathcal{R}_{N_2, d} > \mathcal{R}_{N_1, d}$.
\end{lemma}

\begin{proof}
Apply \cref{eq:ratio} once at $\gamma$ and once at $(1+\lambda)\gamma$, both of which are at least $16$: for every $\varepsilon > 0$, with probability tending to one,
\begin{equation}
\begin{aligned}
  \mathcal{R}_{N_2, d} - \mathcal{R}_{N_1, d}
  &\geq \frac{3\pi}{8} \Big( 1 - \frac{9}{10(1+\lambda)\gamma} \Big) - \frac{3\pi}{8} \Big( 1 - \frac{9}{10\gamma} \Big) \\
  &\qquad - C_1 \big( (1+\lambda)\gamma \big)^{-3/2} - C_1 \gamma^{-3/2} - 2\varepsilon \\
  &\geq \frac{27\pi}{80} \cdot \frac{\lambda}{(1+\lambda)\gamma} - 2 C_1 \gamma^{-3/2} - 2\varepsilon .
\end{aligned}
\end{equation}
Moreover $\tfrac{27\pi}{80} \cdot \tfrac{\lambda}{(1+\lambda)\gamma} > 2C_1 \gamma^{-3/2}$ is equivalent to $\gamma^{1/2} > \tfrac{160 C_1 (1+\lambda)}{27\pi\lambda}$, which holds by assumption; taking $\varepsilon = \tfrac14 \big( \tfrac{27\pi}{80} \cdot \tfrac{\lambda}{(1+\lambda)\gamma} - 2C_1 \gamma^{-3/2} \big)$ gives the claim.
\end{proof}

\noindent\textbf{Deduction of the informal theorem.} Write $A=3\pi/8$ and $B=27\pi/80$. By \cref{eq:ratio}, $\mathcal{R}-A=-B/\gamma+O(\gamma^{-3/2})+o_p(1)$. For sufficiently large fixed $\gamma$, the $O(\gamma^{-3/2})$ term has absolute value at most $B/(4\gamma)$ with probability tending to one, and the $o_p(1)$ term has absolute value at most $B/(4\gamma)$ with probability tending to one as $d\to\infty$. Consequently,
\[
  -\frac{3B}{2\gamma}\leq\mathcal{R}-A\leq-\frac{B}{2\gamma}
\]
with probability tending to one. This gives the claimed two-sided $1/\gamma$ rate, and $\mathcal{R}>1$ once $\gamma$ is sufficiently large because $A>1$. Together with Lemma~\ref{lem:fixed-ratio}, this proves all three conclusions of \cref{thm:ratio}.

\begin{lemma}[Pointwise monotonicity]
\label{lem:pointwise-mono}
Suppose that for every sufficiently large $\gamma$ the limit in probability $\mathcal{R}_\infty(\gamma) := \lim_{d \to \infty} \mathcal{R}$ exists, and that $\mathcal{R}_\infty$ is analytic as a function of $1/\gamma$ in a neighbourhood of $0$, i.e.\ $\mathcal{R}_\infty(\gamma) = \sum_{k \geq 0} c_k \gamma^{-k}$ for all sufficiently large $\gamma$. Then there is $\gamma_2$ such that $\mathcal{R}_\infty'(\gamma) > 0$ for $\gamma > \gamma_2$.
\end{lemma}

\begin{proof}
By \cref{eq:ratio} and convergence in probability, $\lvert \mathcal{R}_\infty(\gamma) - \tfrac{3\pi}{8}(1 - \tfrac{9}{10\gamma}) \rvert \leq C_1 \gamma^{-3/2}$, that is
\begin{equation}
  \Big\lvert \sum_{k \geq 0} c_k \gamma^{-k} - \frac{3\pi}{8} + \frac{27\pi}{80} \gamma^{-1} \Big\rvert \leq C_1 \gamma^{-3/2} .
\end{equation}
Letting $\gamma \to \infty$ gives $c_0 = \tfrac{3\pi}{8}$; dividing both sides by $\gamma^{-1}$ and letting $\gamma \to \infty$ gives $c_1 = -\tfrac{27\pi}{80}$. Hence
\begin{equation}
  \mathcal{R}_\infty'(\gamma) = - \sum_{k \geq 1} k \, c_k \gamma^{-k-1}
  = \gamma^{-2} \Big( \frac{27\pi}{80} - \sum_{k \geq 2} k \, c_k \gamma^{-(k-1)} \Big) ,
\end{equation}
and the series in the parentheses tends to $0$ as $\gamma \to \infty$, so the parenthesized quantity is positive for all sufficiently large $\gamma$.
\end{proof}

\begin{remark}
\label{rem:analyticity}
Neither hypothesis of \cref{lem:pointwise-mono} is proved here. The estimates above pin $(\bar\kappa^{\mathrm{Muon}})^2/d$ down only to within $O(\gamma^{-3/2})$ (\cref{lem:kbar-muon}), so they do not by themselves give the limit in probability; $\rho^{\mathrm{GD}}/\sqrt d$ and $\bar h^{\mathrm{Muon}}/d$, by contrast, do converge, by \cref{lem:rho-gd} and \cref{lem:hbar-muon}. Analyticity does not follow from \cref{eq:ratio} alone either: with $x = 1/\gamma$, the function $1 - x + x^{3/2}\sin(1/x)$ admits an expansion of the same form and is not monotone near $x = 0$.
\end{remark}

\noindent\textbf{Numerical consequences.} Since $\tfrac{3\pi}{8} \approx 1.178 > 1$, \cref{eq:ratio} puts $\mathcal{R}$ above one for all large $\gamma$: past some data budget the GD iterate has the larger margin-to-spread ratio. The leading term $\tfrac{3\pi}{8}(1 - \tfrac{9}{10\gamma})$ is increasing in $\gamma$; \cref{lem:fixed-ratio} turns this into a statement about $\mathcal{R}$ itself, and \cref{lem:pointwise-mono} into one about its limit, under the hypotheses discussed in \cref{rem:analyticity}. The leading term crosses one at $\gamma \approx 5.95$, below the range $\gamma \geq 16$ in which \cref{lem:rho-muon} is proved, so the location of the crossing is established empirically rather than by \cref{thm:ratio}.

\end{document}